\documentclass{article}
\makeatletter
\def\input@path{{./}{styles/}}
\makeatother
\usepackage{arxiv}
\usepackage[round,authoryear]{natbib}
\setcitestyle{authoryear,round,citesep={;},aysep={,},yysep={;}}
\usepackage{amsmath,amssymb,bm,booktabs,array,graphicx,float,microtype,url,tabularx}
\usepackage{longtable}
\newcommand{\PDETableFont}{\fontsize{8}{10}\selectfont}
\newcommand{\PDEMeanSDFont}{\PDETableFont}
\newcommand{\PDEMeanSD}[2]{\mbox{\PDEMeanSDFont #1\textpm#2}}
\usepackage{hyperref}
\hypersetup{
    colorlinks=false,
    pdfborder={0 0 0.55},
    linkbordercolor={0.12 0.29 0.47},
    citebordercolor={0 1 0},
    urlbordercolor={0 1 1}
}
\ifdefined\pdfinfoomitdate\pdfinfoomitdate=1\fi
\ifdefined\pdfsuppressptexinfo\pdfsuppressptexinfo=15\fi
\ifdefined\pdftrailerid\pdftrailerid{}\fi
\usepackage[nameinlink,noabbrev]{cleveref}
\usepackage{amsmath,amssymb}
\usepackage{listings}
\usepackage{xcolor}
\usepackage{colortbl}
\DeclareRobustCommand{\addedcitep}[1]{{\color{blue}\citep{#1}}}

\definecolor{PDEInk}{HTML}{214A55}
\definecolor{PDEFocus}{HTML}{EAF2F3}
\definecolor{PDEGray}{HTML}{F3F4F5}
\newcolumntype{Y}{>{\raggedright\arraybackslash}X}
\newcommand{\dataset}{\textsc{PDE-OBS}}

\newcommand{\code}[1]{\texttt{#1}}

\usepackage{xspace}
\usepackage{enumitem}
\usepackage{titletoc}
\usepackage[most]{tcolorbox}
\definecolor{codegreen}{rgb}{0,0.6,0}
\definecolor{codegray}{rgb}{0.5,0.5,0.5}
\definecolor{codepurple}{rgb}{0.58,0,0.82}
\definecolor{backcolour}{rgb}{0.95,0.95,0.92}
\definecolor{indigo}{rgb}{0.0, 0.25, 0.42}
\definecolor{ceruleandark}{rgb}{0.03, 0.27, 0.49}
\definecolor{LightGray}{gray}{0.95}
\definecolor{lightblue}{rgb}{0.122, 0.435, 0.698}
\colorlet{lightgray}{gray!10!}
\tcbset{on line, boxsep=0pt, left=0pt,right=0pt,top=0pt,bottom=0pt, colframe=white,colback=lightgray, highlight math style={enhanced}}
\lstdefinestyle{al4pde}{
    backgroundcolor=\color{LightGray},
    commentstyle=\color{codegreen},
    keywordstyle=\color{magenta},
    numberstyle=\tiny\color{codegray},
    stringstyle=\color{codepurple},
    basicstyle=\ttfamily\footnotesize,
    frame=lines, framesep=2mm, columns=flexible,
    breakatwhitespace=false, breaklines=true, captionpos=b, keepspaces=true,
    numbers=left, numbersep=5pt, showspaces=false, showstringspaces=false, showtabs=false, tabsize=2
}
\newcommand{\appref}[1]{Appendix~\ref{#1}}
\newcommand{\weblink}[2]{\href{#1}{\textcolor{indigo}{#2}}}
\newcommand{\coderepo}{https://github.com/ru1ch3n/PDE-OBS}

\newcommand{\NewCDPositive}{49}
\newcommand{\NewCDMedian}{10.46}
\newcommand{\NewSubsetPositive}{13}
\newcommand{\NewSubsetCDMedian}{10.74}
\newcommand{\NewDestinationPositive}{3,281}
\newcommand{\NewStrictCount}{3,646}
\newcommand{\NewMinorCount}{314}
\newcommand{\NewLargeCount}{9}
\newcommand{\NewLayoutOneCount}{545/588}
\newcommand{\NewLayoutOnePercent}{92.69}
\newcommand{\NewLayoutTwoCount}{276/294}
\newcommand{\NewLayoutTwoPercent}{93.88}
\newcommand{\NewDensityFiftyCount}{13/49}
\newcommand{\NewDensityFiftyMedian}{1.58}
\newcommand{\NewDensitySixtyFiveCount}{37/49}
\newcommand{\NewDensitySixtyFiveMedian}{0.93}
\newcommand{\NewDensityEightyCount}{49/49}
\newcommand{\NewDensityEightyMedian}{0.29}
\newcommand{\NewSubsetDensityFiftyMedian}{9.93}

\author{Anonymous Authors}

\usepackage{xcolor}

\title{\dataset{}: Controlled Evaluation\\Across Observation Patterns}
\date{}
\graphicspath{{./}{figures/}}
\newcommand{\pdeobspublicauthors}{%
\begin{minipage}{\dimexpr\textwidth-2\tabcolsep\relax}
\centering\normalfont
{\large\bfseries
Ruichen Xu\textsuperscript{1,\ensuremath{\dagger}},\quad Siyao Wang\textsuperscript{2},\quad Fang Wan\textsuperscript{1},\quad Jiacheng Qiu\textsuperscript{1},\\[4pt]
Wenhan Gao\textsuperscript{1},\quad Jiaxing Zhang\textsuperscript{3},\quad Linsey Pang\textsuperscript{4,6},\\[4pt]
Ravid Shwartz-Ziv\textsuperscript{5},\quad Yann LeCun\textsuperscript{5},\quad Yuefan Deng\textsuperscript{1,\ensuremath{\dagger}}\par}
\vspace{8pt}
{\small
\textsuperscript{1}Stony Brook University\quad \textsuperscript{2}University of California, Davis\\[3pt]
\textsuperscript{3}Independent Research\quad \textsuperscript{4}PayPal\\[3pt]
\textsuperscript{5}New York University\quad \textsuperscript{6}Northeastern University\par}
\vspace{6pt}
{\small Code: \url{https://github.com/ru1ch3n/PDE-OBS}}
\end{minipage}%
}
\author{\pdeobspublicauthors}
\makeatletter
\g@addto@macro\@thanks{%
  \footnotetext[2]{Corresponding authors: Ruichen Xu
  (\href{mailto:ruichen.xu@stonybrook.edu}{\nolinkurl{ruichen.xu@stonybrook.edu}})
  and Yuefan Deng
  (\href{mailto:yuefan.deng@stonybrook.edu}{\nolinkurl{yuefan.deng@stonybrook.edu}}).}%
}
\makeatother
\hypersetup{
  pdftitle={PDE-OBS: Controlled Evaluation Across Observation Patterns},
  pdfauthor={Ruichen Xu, Siyao Wang, Fang Wan, Jiacheng Qiu, Wenhan Gao, Jiaxing Zhang, Linsey Pang, Ravid Shwartz-Ziv, Yann LeCun, Yuefan Deng},
  pdfsubject={Public preprint}
}

\renewcommand{\shorttitle}{PDE-OBS}
\renewcommand{\headeright}{Preprint}

\colorlet{RevisionColor}{black}
\colorlet{RevisionCodeColor}{black}

\DeclareRobustCommand{\revcode}[1]{#1}

\DeclareRobustCommand{\revnew}[1]{#1}
\newenvironment{newrevision}{\begingroup}{\endgroup}

\makeatletter

\@ifundefined{revisionblock}{%
    \newenvironment{revisionblock}
        {\begingroup}
        {\endgroup}
}{%
    
}

\@ifundefined{movedblock}{%
    \newenvironment{movedblock}[1]
        {\par\ignorespaces}
        {\par\ignorespacesafterend}
}{%
    \renewenvironment{movedblock}[1]
        {\par\ignorespaces}
        {\par\ignorespacesafterend}
}

\makeatother

\DeclareRobustCommand{\addedcitep}{\citep}

\begin{document}

\raggedbottom
\maketitle

\begin{abstract}
Physical-field reconstruction and forecasting depend on both measurement density and spatial layout, yet evaluation under a single observation pattern does not characterize performance when that pattern changes. We introduce \dataset{}, an integrated benchmarking platform spanning numerical data generation, model training, and inference and evaluation under varying observation conditions. It combines 560,000 fields and trajectories from seven partial differential equation families with configurable observation operators and seven adapted baseline methods for stationary reconstruction and short-horizon forecasting. Separating observation construction from physical records allows users to specify parameterized patterns and deterministic mixtures for training and testing while preserving prediction targets and data splits. The evaluation protocol uses references trained for each test pattern to compare models on identical test observations and targets, alongside equal-count groups for spatial-layout comparisons. On a 14,000-record subset, we evaluate 441 trained models under nine test patterns, yielding 3,969 evaluations. Mean cross-pattern error exceeds mean matched-pattern error in all 49 PDE--method pairs, and this finding persists in a configuration-matched subset of 117 models. Denser test observations do not consistently reduce error for a fixed model. \revnew{Mixed-pattern training on five completed pairs reduces large single-pattern transfer errors, although destination-trained references usually remain more accurate.} Together, the benchmark and findings support systematic evaluation of observation-pattern sensitivity and provide a reusable workflow for developing methods under changing measurement conditions. Code: \weblink{\coderepo}{Public GitHub repository}.
\end{abstract}

\section{Introduction}
\label{sec:motivation}

Reconstructing physical fields and predicting their evolution from partial observations \citep{evensen1994assimilation,carrassi2018assimilation,raissi2018numericalgp,brajard2020assimilation,williams2024shred} are central problems in scientific machine learning \citep{karniadakis2021physics,brunton2020fluid,willard2022integrating,kovachki2023operator,azizzadenesheli2024operators}. The information available to a model depends on both the number and the locations of its measurements \citep{manohar2018sensors,yildirim2009sensors,krause2008sensors}. Scattered, clustered, and line-based layouts can contain the same number of measurements yet reveal different information about the underlying field \citep{joshi2009sensors,clark2019greedy,chaturantabut2010deim,drmac2016selection,peherstorfer2020stability}. A model's accuracy under its training observation pattern therefore needs to be considered alongside its performance when measurement density or spatial layout changes. This distinction connects prediction from partial observations to the broader challenge of evaluation under distribution shift \citep{koh2021wilds,gulrajani2021domainbed} and motivates treating observation conditions as an explicit dimension of a scientific benchmark.

Shared benchmarks support systematic comparisons by providing datasets, baseline implementations, and common evaluation interfaces \citep{otness2021extensible,bonnet2022airfrans,luo2023cfdbench,herde2024poseidon}. PDEBench combines configurable numerical generation with baseline models \citep{takamoto2022pdebench}, while The Well unifies training and evaluation across diverse physical simulations \mbox{\citep{ohana2024well}}. Learning from partial observations is also an established setting \citep{raissi2018deephidden,raissi2019pinns,raissi2020hiddenfluid,jin2021nsfnets,sun2020bayesian}, including LANO and its POBench-PDE benchmark \citep{hou2026pobench}. Building on these efforts, we address a complementary evaluation question: \emph{how does prediction performance change when observation conditions at inference differ from those used in training?} Studying this question across physical systems and methods calls for a reusable workflow that connects numerical data generation, model training, and inference while allowing training and test observation patterns to be specified separately.

\begin{figure}[t]
    \centering
    \IfFileExists{figures/PDE_OBS_aligned_subtitles.pdf}{%
        \includegraphics[width=1.0\linewidth]{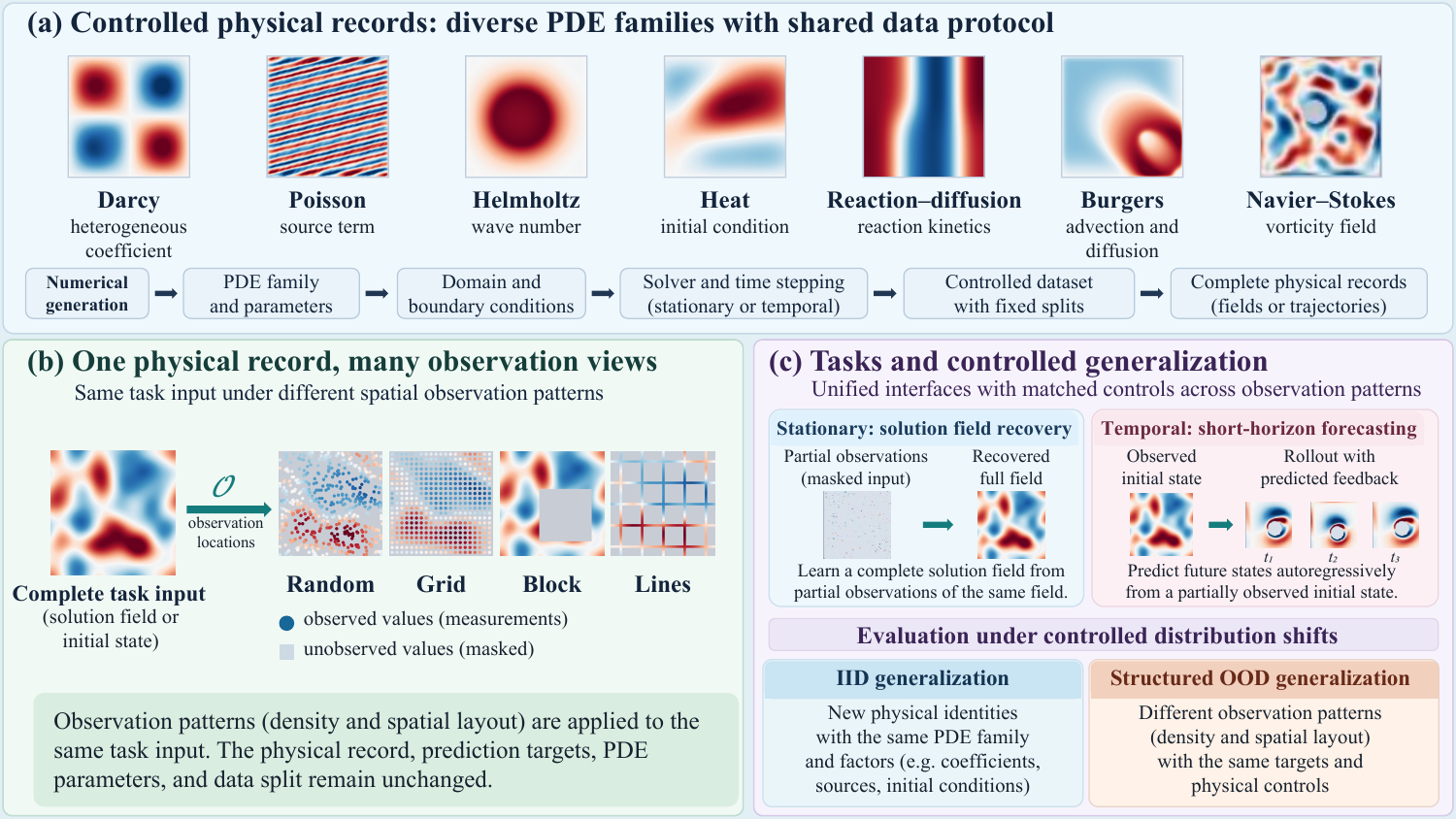}%
    }{%
        \fbox{%
            \parbox[c][4cm][c]{0.95\linewidth}{%
                \centering
                Insert the corrected PDE-OBS overview figure here.%
            }%
        }%
    }
    \caption{\textbf{Overview of \dataset{}.} The benchmark connects numerical data generation, observation construction, model training, and inference across observation conditions. \textbf{(a)} Numerical solvers generate complete fields and trajectories with fixed train/test splits. \textbf{(b)} Different observation patterns are applied to the same physical record while prediction targets remain unchanged. The observed field is the solution for reconstruction and the initial state for forecasting. \textbf{(c)} Models perform stationary reconstruction or autoregressive forecasting and are evaluated under matched and shifted observation patterns. For each test pattern, a reference model is trained on the training split using that pattern and evaluated on identical test observations and targets. Field maps are schematic; Section~\ref{sec:protocol} specifies the benchmark tasks and observation protocols.}
    \label{fig:overview}
\end{figure}

We introduce \dataset{}, an integrated benchmarking platform for stationary field reconstruction and short-horizon forecasting under partial observations (Figure~\ref{fig:overview}). The platform connects numerical data generation, observation construction, model training, checkpoint-based inference, and evaluation through shared definitions of physical records, model inputs, and prediction targets. Stationary reconstruction recovers a complete solution field from measurements of that same field; forecasting predicts future states from a partially observed initial state using autoregressive prediction. Both tasks use complete target fields for training supervision, while inference receives observed values, observation masks, and domain geometry. Users can generate or load physical records, train models under specified observation patterns or deterministic mixtures, and evaluate the resulting checkpoints under alternative observation conditions. Throughout these comparisons, the underlying physical records, prediction targets, and train/test splits remain fixed (Section~\ref{sec:protocol}).

Using a 14,000-record subset, we evaluate 441 trained models spanning seven PDE families, seven adapted baseline methods, and nine training observation patterns. Testing each model under all nine patterns produces 3,969 evaluations and 49 complete training-pattern-by-test-pattern matrices. Mean cross-pattern error exceeds mean matched-pattern error in every PDE--method pair, and the same finding holds in a configuration-matched subset of 117 models. Comparisons at equal measurement counts reveal performance differences across spatial layouts, while fixed-model density comparisons show that denser test observations do not consistently reduce error. \revnew{Additional mixed-pattern training on five completed pairs reduces large transfer errors, but does not uniformly outperform destination-trained references} (Section~\ref{sec:results}, \appref{app:mixed-control}). These findings motivate reporting observation-pattern sensitivity alongside matched accuracy and illustrate how the integrated workflow supports systematic investigation of training and inference conditions.

Our contributions are fourfold.
\begin{itemize}[leftmargin=*,itemsep=1pt,topsep=2pt]
    \item \textbf{An integrated benchmark workflow.} We connect numerical data generation, observation construction, model training, inference, and evaluation within a reusable platform. Seven adapted baseline methods share explicit task and input definitions, supporting method development and evaluation under configurable observation conditions.
    \item \textbf{A configurable physical data resource.} We provide 560,000 fields and trajectories across seven PDE families, four boundary protocols, and ten condition constructions. Numerical solvers and condition-field generators support selectable physical regimes, resolutions, and data tiers, with record-level metadata and reproducible data splits.
    \item \textbf{Systematic evaluation across observation patterns.} Nine reference patterns, including two equal-count layout groups, define matched and cross-pattern comparisons on shared physical records. For each test pattern, a destination-trained reference is evaluated on identical observations and targets. Parameterized operators and deterministic mixtures extend these comparisons to additional training and test conditions.
    \item \textbf{A comprehensive empirical result resource.} We provide complete cross-pattern results, full-precision per-record errors, and training configurations for the 441-model evaluation. Analyses of spatial layout, test density, a configuration-matched subset, and mixed-pattern training characterize complementary aspects of observation-pattern sensitivity (Section~\ref{sec:results}).
\end{itemize}

\section{Related Work}
\label{sec:related_work}

\paragraph{PDE benchmarks and autoregressive forecasting.}
Shared benchmarks support comparisons of physical prediction methods through common datasets and evaluation protocols \citep{rasp2020weatherbench,rasp2024weatherbench2,chung2023blastnet,johnson2023oceanbench}. PDEBench combines configurable numerical data generation with baseline implementations \citep{takamoto2022pdebench}, while The Well provides standardized model interfaces across diverse physical simulations \citep{ohana2024well}. APEBench focuses on autoregressive PDE emulation and examines how training procedures affect successive-step prediction accuracy \citep{koehler2024apebench}. PDEArena supports neural surrogate modeling \citep{gupta2022pdearena}, and AL4PDE studies data-efficient learning for neural PDE solvers \citep{musekamp2024al4pde}. Complementary resources address inverse problems through PDEInvBench \citep{goel2026pdeinvbench} and physical prediction from real measurements through RealPDEBench \citep{hu2026realpdebench}. \dataset{} contributes an integrated benchmark workflow spanning numerical data generation, model training, and inference and evaluation under varying observation conditions. By specifying training and test observation patterns separately on fixed physical records, targets, and data splits, it makes observation-pattern sensitivity a systematic dimension of benchmark evaluation.

\paragraph{Reconstruction from sparse sensors.}
Recovering physical fields from limited measurements connects classical sparse reconstruction and inverse problems \citep{donoho2006compressed,candes2006robust,stuart2010inverse,callaham2019robust} with data-driven field estimation. Gappy proper orthogonal decomposition reconstructs fields by fitting the coefficients of a low-rank basis to observed entries \citep{everson1995gappypod,willcox2006gappy,gunes2006gappy,rowley2017modelreduction}. Learned approaches map reduced measurements to complete fields \citep{erichson2020shallow,fukami2019superresolution,dubois2022reconstruction,fukami2023survey}. Senseiver supports reconstruction from sparse, irregular sensors \citep{santos2023senseiver}, and Voronoi-based reconstruction accommodates varying sensor numbers and locations \citep{fukami2021voronoi}. These approaches motivate examining reconstruction accuracy across measurement densities and spatial layouts. \dataset{} provides a common evaluation protocol for this purpose: each trained model is tested under multiple observation patterns on shared physical records, references trained for each test pattern are evaluated on identical observations and targets, and equal-count groups compare layouts at fixed measurement counts.

\paragraph{Learning from partial observations.}
Training supervision is another important distinction among partial-observation prediction settings. LANO introduces POBench-PDE for temporal prediction with point-wise and patch-wise masking, using partially observed inputs and outputs during training and complete target fields for evaluation \citep{hou2026pobench}. \dataset{} uses complete training targets for stationary reconstruction and short-horizon forecasting, while specifying the input observation patterns separately for training and inference. This setting supports comparisons of single-pattern and mixed-pattern training, together with matched and shifted test observations, under complete-target supervision.

DiffusionPDE and FunDPS use diffusion models to predict physical fields conditioned on partial observations \citep{huang2024diffusionpde,yao2025fundps}, while CORAL and DINo employ neural-field or implicit representations for operator learning and forecasting \citep{serrano2023coral,yin2023dino}. These methods offer complementary approaches to prediction and field representation. \dataset{} provides reusable data, observation operators, task definitions, and evaluation interfaces through which additional methods can be adapted and compared under shared measurement conditions and prediction targets.
\section{\dataset{}: A Benchmark for Learning from Partial Observations}
\label{sec:protocol}

Following the modular benchmark design of PDEBench \citep{takamoto2022pdebench}, \dataset{} integrates numerical data generation, observation construction, model training, and inference and evaluation under varying observation conditions. These stages share physical records, task definitions, and data splits, allowing training and test observation patterns to be specified separately. We first define the prediction tasks, then describe the data resource, observation protocols, evaluation metrics, baseline models, and end-to-end workflow. Detailed benchmark specifications and usage instructions are provided in \appref{app:have} and \appref{app:provide}.

\subsection{General Problem Definition}
\label{sec:problem_definition}

A physical record $i$ contains a complete condition field $c_i$, such as a coefficient, source, or initial state; a complete solution or trajectory; a geometry field $G_i$ marking walls or obstacles; and metadata. For a given task, let $x_i$ denote the input field and $y_i$ the prediction target. We use $v$ and $w$ to denote the training and test observation protocols, respectively. Each protocol specifies how a record-specific observation mask is constructed. Given a binary mask $M_{i,w}$ with ones at observed locations, the model input and prediction are
\begin{equation}
z_{i,w}=M_{i,w}\odot x_i,
\qquad
\widehat y_i^{\,v,w}=f_{\theta_v}\bigl(z_{i,w},M_{i,w},G_i\bigr),
\label{eq:input}
\end{equation}
where $\odot$ denotes elementwise multiplication and $\theta_v$ denotes parameters learned under protocol $v$. Unobserved entries are zero-filled, and the mask distinguishes missing entries from measured zeros. The geometry field describes the physical domain separately from observation availability. Changing $w$ changes the observations supplied to a fixed model while preserving the underlying record and prediction target.

In \emph{stationary field reconstruction}, $x_i=y_i=u_i$: the model reconstructs a complete solution field from measurements of that same field. Training supervises the complete solution, including both observed and hidden locations.

In \emph{short-horizon forecasting}, $x_i=u_{i,0}$ is the initial state and $y_i=(u_{i,1},u_{i,2},u_{i,3})$ comprises the next three stored states. The predictor $f_{\theta_v}$ applies a shared one-step model autoregressively:
\begin{equation}
\widehat u_{i,1}=F_{\theta_v}\bigl(M_{i,w}\odot u_{i,0},M_{i,w},G_i\bigr),
\qquad
\widehat u_{i,h+1}=F_{\theta_v}\bigl(\widehat u_{i,h},\mathbf 1,G_i\bigr),\quad h=1,2.
\label{eq:rollout}
\end{equation}
The all-one mask marks the fully specified predicted field supplied at subsequent steps. Training and evaluation follow the same recurrence: predictions provide the recurrent inputs, and complete future states provide supervision during training. Gradients propagate through the three-step rollout. At inference, the baselines receive only observed values, observation masks, and domain geometry. PINO additionally uses physical coefficients, sources, or parameters in its training residual. Detailed task definitions, information access, and the supported forward and inverse extensions are specified in \appref{app:have:tasks}.

\subsection{Overview of Datasets and Observation Protocols}
\label{sec:data_observations}

\paragraph{Physical systems and data generation.}
Configurable numerical solvers generate records for seven PDE families. The stationary systems are \textbf{Darcy}, \tcbhighmath{-\nabla\cdot(a\nabla u)=f}, with positive coefficient $a$; \textbf{Poisson}, \tcbhighmath{-\Delta u=f}; and \textbf{Helmholtz}, \tcbhighmath{(-\Delta-k^2)u=f}, in a coercive low-wavenumber setting. The temporal systems are \textbf{Heat}, \tcbhighmath{\partial_t u=\alpha\Delta u}; Allen--Cahn-type \textbf{reaction--diffusion}, \tcbhighmath{\partial_t u=\kappa\Delta u+\rho(u-u^3)}; scalar two-dimensional \textbf{Burgers}, \tcbhighmath{\partial_t u+u(\partial_{x_1}u+\partial_{x_2}u)=\nu\Delta u}; and \textbf{Navier--Stokes} vorticity, \tcbhighmath{\partial_t\omega+\boldsymbol v\cdot\nabla\omega=\nu\Delta\omega+f}. For Navier--Stokes, the state represented by $u$ in the task definitions is the scalar vorticity $\omega$.

The data resource contains 560,000 fields and trajectories across seven PDE families, four boundary protocols, and ten condition constructions. Each PDE--boundary--construction combination contains 2,000 records distributed across three family-specific physical regimes. Generation supports selectable spatial resolutions and nested data tiers. Generated records are assessed using family-specific discrete PDE residuals, boundary and initial-condition checks, and transition-consistency checks. Numerical methods, parameters, time grids, and validation results are documented in \appref{app:numerical-details}; data-generation options are described in \appref{app:provide:data}.

\paragraph{Evaluation subset and data splits.}
The reported experiments use a 14,000-record subset with 2,000 records per PDE family. This subset consists of scalar $128\times128$ fields with smooth random-field conditions, Dirichlet boundaries for stationary systems, and periodic boundaries for temporal systems. Stationary boundary values are zero on the outer stored cell-center layer. Temporal trajectories contain 15 equally spaced states, of which frame 0 supplies the partially observed input and frames 1--3 supply the forecasting targets.

A seeded ranking within each physical regime assigns 1,800 records to training and 200 to testing per PDE family. These identity lists are shared across all methods and observation patterns. Because each identity belongs to a physical record, changing its observation pattern preserves its data split. The data design, identity conventions, and exact partitioning rules are provided in \appref{app:have:data}.

\paragraph{Reference observation patterns.}
The benchmark constructs binary grid masks from the physical-record identity, observation protocol, and fixed observation seed. For a given record and protocol, the resulting mask is shared across methods and remains fixed across training epochs. Nine reference patterns define the main evaluation: uniformly random points at nominal densities of 50\%, 65\%, and 80\% (R50, R65, R80); horizontal rows (H); vertical columns (V); crossed horizontal and vertical lines (LI); observations outside a missing square (BL); a perimeter band (BD); and Gaussian-clustered points (CL).

Two groups support spatial-layout comparisons at fixed measurement counts. R50/H/V/CL contain exactly 8,192 observed locations, while BL/LI/BD contain exactly 8,284. Density comparisons use the separately sampled location sets of R50, R65, and R80, thereby changing both measurement counts and observed locations. All nine reference patterns use noiseless observations and cover at least half the grid. BL and CL use non-wrapped array coordinates, and BD specifies a perimeter sampling layout on both stationary and periodic domains. Exact constructions, observation counts, and seed definitions are given in \appref{app:have:obs}.

\paragraph{Configurable training and inference conditions.}
Beyond the reference patterns, users can specify observation densities and geometric parameters, combine patterns into deterministic mixtures, register custom operators, or supply stored masks. For example, \code{uniform 20} requests uniformly sampled observations at a target density of 20\%, while \code{line 30 + random 20} defines a mixture of two component patterns. A mixture assigns one component to each physical record and retains that assignment across epochs. The same specification can be used for training and as a test observation protocol, allowing a trained checkpoint to be evaluated under alternative measurement conditions. Stored masks also support explicitly constructed nested observation sets. Observation namespaces, parameter syntax, and reference-pattern replay are detailed in \appref{app:provide:obs}; mixed-pattern experiments are presented in Section~\ref{sec:results} and \appref{app:mixed-control}.

\subsection{Overview of Metrics}
\label{sec:metrics}

The primary metric is the mean per-record relative $L^2$ error. For each PDE--method pair, evaluating models trained under nine reference patterns against all nine test patterns produces a $9\times9$ matrix:
\begin{equation}
E_{v,w}=\frac{1}{200}\sum_{i\in\mathcal I_{\rm test}}
\frac{\|\widehat y_i^{\,v,w}-y_i\|_2}{\max(\|y_i\|_2,10^{-12})},\qquad
D=\frac{1}{9}\sum_v E_{v,v},\qquad C=\frac{1}{72}\sum_{v\ne w}E_{v,w}.
\label{eq:main_metrics}
\end{equation}
Rows correspond to training patterns and columns to test patterns. The matched-pattern mean $D$ summarizes the nine diagonal cells, while the cross-pattern mean $C$ summarizes the 72 off-diagonal cells. The norm covers the complete target field, including observed and hidden locations. For forecasting, it is computed jointly over all three target frames.

To examine individual training-to-test directions, we compare each model with a \emph{destination-trained reference}: a model of the same method trained on the common training split using test pattern $w$. The directional difference $\Delta_{v\to w}=E_{v,w}-E_{w,w}$ and ratio $R_{v\to w}=E_{v,w}/E_{w,w}$ compare predictions on identical test observations and targets. Positive differences and ratios above one indicate higher error than the reference. Averaging the 72 directional differences gives $C-D$, while individual directions and equal-count groups identify the patterns associated with larger gaps. Ratios use the positive, full-precision reference errors. Figure~\ref{fig:cross-observation} additionally reports absolute errors on a shared logarithmic scale.

Cell-level mean $\pm$ sample standard deviation summarizes the 200 per-record errors. The standard deviations reported with $D$ and $C$ describe variability across their nine and 72 cell means, respectively. Forecasting results additionally include errors at each prediction horizon. The evaluator also supports stationary observed/hidden error decomposition and data-consistency diagnostics, whose definitions are provided in \appref{app:have:metrics}.

Evaluation checks record identities, observation masks, array shapes, target-frame indices, and the finiteness of prediction and target entries before scoring. Each reported cell contains all 200 test records, with all finite errors included in the aggregation. Evaluation procedures, data and checkpoint bindings, and per-record result records are documented in \appref{app:validity}.

\subsection{Baseline Models and Training}
\label{sec:baselines}

Seven baseline adaptations share the observation interface in Equation~\eqref{eq:input}. \textbf{U-FNO} uses a two-dimensional architecture combining Fourier and U-Fourier blocks. \textbf{FNO} uses a compact mask-channel architecture, while the \textbf{CNO-inspired} encoder--decoder applies fixed low-pass filtering. \textbf{DeepONet} uses a convolutional branch to encode the masked field. \textbf{GNOT} uses a structured-grid adaptation, and \textbf{Transolver} uses structured-grid slice attention with a larger temporal preset. \textbf{PINO} augments an FNO backbone with a PDE-residual training term. The four temporal families use the autoregressive recurrence in Equation~\eqref{eq:rollout}. Architectural configurations and parameter counts are specified in \appref{app:have:models}, with source references, upstream revisions, licenses, and adaptation details in \appref{app:have:adaptations}.

\paragraph{Training settings.}
The shared data objective is the minibatch mean of per-record relative-$L^2$ losses over the complete target. For forecasting, the loss is computed jointly over three predicted frames, with gradients propagated through the autoregressive rollout. PINO additionally includes its physical residual. Each run follows its recorded optimizer \citep{kingma2015adam,loshchilov2019adamw}, learning-rate schedule \citep{smith2019superconvergence}, batch size, epoch budget, and training-loss stopping rule where enabled. We evaluate the final checkpoint of each run. Per-run configurations accompany the results, and Section~\ref{sec:results} examines observation-pattern sensitivity within a configuration-matched subset. Training recipes and model presets are provided in \appref{app:have:models}, and configurable training controls are described in \appref{app:provide:models}.

\subsection{Data Format, Access, and Extensibility}
\label{sec:software}

Complete physical records are stored as HDF5 shards with JSON metadata describing identities, PDE families, boundary protocols, condition constructions, physical regimes, and time coordinates. Observation masks are constructed when records are accessed, allowing the same physical data to support multiple training and inference conditions. The storage format and loading procedures are specified in \appref{app:have:data} and \appref{app:provide:data}.

The integrated workflow allows users to generate or load records, configure observations, train a model, reload its checkpoint, and perform inference under alternative observation patterns. Versioned input and prediction formats separate inference from scoring: the predictor receives observed values, masks, and geometry, while targets are supplied separately for evaluation. The same prediction interface accepts externally supplied observations. Input preparation and checkpoint loading are detailed in \appref{app:provide:infer}, with an end-to-end example in \appref{app:provide}.

Code, configurations, tests, and documentation are available in \weblink{\coderepo}{our public GitHub repository} under the MIT code license. The result release provides full-precision scores, per-record errors, and data and checkpoint bindings. Public registries support additional observation operators, condition constructions, prediction methods, PDE families, and metrics while preserving the nine reference-pattern definitions. Versioned formats, resolved configurations, and regression tests support maintenance and reuse. Access instructions and extension procedures are provided in \appref{app:provide:software}.

\section{Results Across Observation Patterns}
\label{sec:results}
\label{sec:evidence}
\paragraph{Evaluation setup.}
Using the benchmark workflow, we evaluate 441 trained models spanning seven PDE families, seven baseline methods, and nine training observation patterns. Each model is tested under all nine patterns on the same 200 held-out records for its PDE family, yielding 3,969 evaluations and 49 complete $9\times9$ matrices. Evaluation verifies record identities, observation masks, array shapes, finiteness, and physical-time mappings before aggregation (\appref{app:validity}). We first examine matched and cross-pattern performance, then assess the configuration-matched subset, equal-count layout changes, and fixed-model density responses. Additional mixed-pattern training experiments examine how including multiple observation patterns during training changes performance across the nine test conditions.

\begin{figure}[!t]
    \centering
    \begingroup
    \makeatletter
    \ifdefined\Gread@transgrouptrue
        \let\Gread@transgrouptrue\Gread@transgroupfalse
    \fi
    \makeatother
    \includegraphics[width=\linewidth]{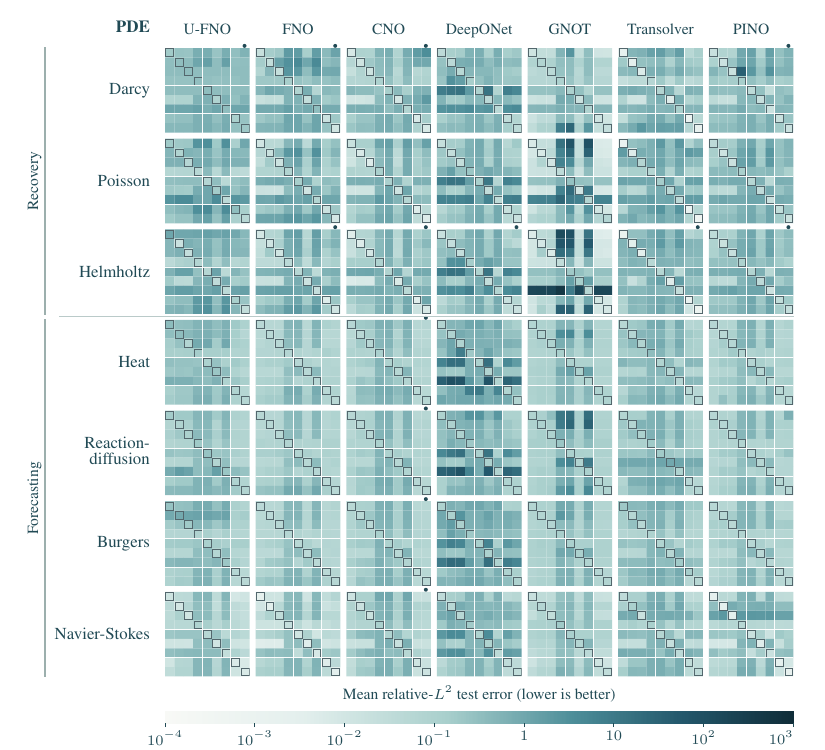}
    \endgroup
    \caption{\textbf{Performance across training and test observation patterns.} Each PDE--method panel contains a $9\times9$ matrix with training patterns in rows and test patterns in columns, both ordered R50, H, V, CL, BL, LI, BD, R65, R80. Color denotes mean relative-$L^2$ error over 200 test records on a shared absolute logarithmic scale; darker colors indicate larger errors. Outlined cells ({\color[HTML]{53676D}$\square$}) identify matched training and test patterns. Dots ({\color[HTML]{214A55}$\bullet$}) identify the 13 PDE--method pairs in the configuration-matched subset. Thin separators distinguish the two equal-count layout groups and the denser random patterns. The figure contains all 3,969 evaluations. Per-cell means, sample standard deviations, and training configurations are reported in \appref{app:distributions}. CNO denotes the CNO-inspired adaptation.}
    \label{fig:cross-observation}
\end{figure}

\paragraph{Does matched accuracy characterize cross-pattern performance?}
We first compare the matched-pattern mean $D$ with the cross-pattern mean $C$ within each PDE--method pair. In Figure~\ref{fig:cross-observation}, $D$ summarizes the outlined diagonal cells and $C$ summarizes the off-diagonal cells, with all evaluations using the same physical records and prediction targets. We find $C>D$ in \NewCDPositive{} of 49 pairs, with median $C/D=\NewCDMedian$ \revnew{(95\% paired test-identity bootstrap interval [10.07, 10.71]; \appref{app:bootstrap})}. At the directional level, \NewDestinationPositive{} of the 3,528 comparisons satisfy $E_{v,w}>E_{w,w}$, indicating higher error than the reference trained for the test pattern. These results reveal a widespread gap between matched and cross-pattern performance across the evaluated methods and physical systems. Figure~\ref{fig:cross-observation} shows the absolute errors underlying these comparisons, while Table~\ref{tab:main-summary} summarizes matched and cross-pattern errors across the seven methods for each PDE family.

\paragraph{Does the finding persist under matched training configurations?}
We next examine the 13-pair configuration-matched subset, comprising 117 models and marked by dots in Figure~\ref{fig:cross-observation}. Within each pair, all nine models complete 500 epochs with the same architecture, optimizer, learning-rate schedule, batch size, loss, and stopping settings. Configurations differ only in the observation pattern and run-name/data-root fields. Subset membership is determined from configuration metadata and was fixed before the current prediction evaluation. We again find $C>D$ in \NewSubsetPositive{} of 13 pairs, with median $C/D=\NewSubsetCDMedian$ \revnew{(95\% interval [10.44, 11.12], conditional on these trained models)}. Thus, the higher cross-pattern mean error persists within pairs whose recorded training configurations agree. The subset definition, configuration comparisons, and detailed results are provided in \appref{app:training-sensitivity}.

\paragraph{Does spatial layout matter at fixed measurement counts?}
To examine layout changes at a fixed number of measured locations, we restrict training and test patterns to either R50/H/V/CL ($K=8{,}192$) or BL/LI/BD ($K=8{,}284$). Cross-pattern error exceeds the destination-trained reference in $\NewLayoutOneCount$ (\NewLayoutOnePercent\%) and $\NewLayoutTwoCount$ (\NewLayoutTwoPercent\%) directed comparisons, respectively. Table~\ref{tab:equal-count} separates the results for stationary reconstruction and forecasting. Performance gaps therefore remain prevalent when training and test patterns contain the same number of measurements, supporting spatial layout as an evaluation dimension alongside density. For stationary Dirichlet tasks, the measurement count includes sampled prescribed boundary values; the observation constructions and boundary coverage are specified in \appref{app:have:obs}.

\paragraph{How does test observation density affect a fixed model?}
We now hold each checkpoint fixed and change its test observation pattern from R50 to R80. The ratio $E_{v,\mathrm{R80}}/E_{v,\mathrm{R50}}$ compares the resulting errors, with values below one indicating lower error under R80. Across the 49 PDE--method pairs, R80 reduces error in \NewDensityFiftyCount{} R50-trained models (median ratio $\NewDensityFiftyMedian$), \NewDensitySixtyFiveCount{} R65-trained models ($\NewDensitySixtyFiveMedian$), and \NewDensityEightyCount{} R80-trained models ($\NewDensityEightyMedian$). Table~\ref{tab:density-directions} additionally reports the responses to R65. The benefit of denser test observations therefore depends on the training pattern. R50, R65, and R80 use separately sampled location sets, so these comparisons evaluate changes in both measurement count and sampled locations under the benchmark's random-observation protocols (\appref{app:have:obs}).

\begin{figure}[!t]
    \centering
    \includegraphics[width=\linewidth]{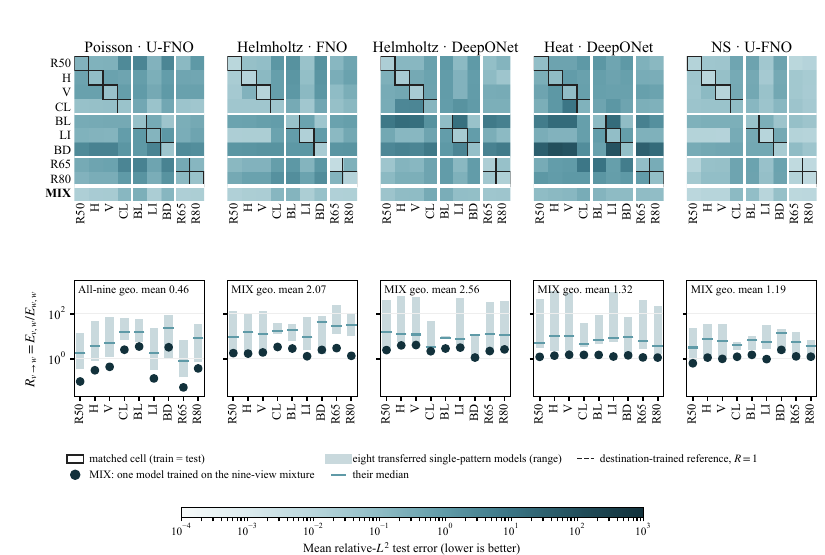}
    \caption{\revnew{\textbf{Single-pattern and mixed-pattern training: five completed pairs.} Top: nine single-pattern rows and one additional MIX row, with test patterns in columns and the shared error scale of Figure~\ref{fig:cross-observation}. Outlines mark matched cells. Bottom: MIX ratios to the destination-trained reference (dots) versus the range and median of eight transferred models (bars); the dashed line is $R=1$. Poisson's displayed geometric mean uses all nine destinations descriptively; its BL/BD/CL recipe-matched mean is 3.05. MIX rows supplement, rather than replace, the 441-model grid (\appref{app:mixed-control}).}}
    \label{fig:mixed-control}
\end{figure}

\paragraph{How does mixed-pattern training change performance across patterns?}
\begin{newrevision}
We examine five completed MIX models: Helmholtz/FNO, Helmholtz/DeepONet, Heat/DeepONet, Navier--Stokes/U-FNO, and Poisson/U-FNO. Each uses the same 1,800 training records as its single-pattern references, with one deterministically assigned reference pattern per record, fixed across epochs. The first four complete 500 epochs; Poisson completes 200 under the one-cycle recipe shared by its BL, BD, and CL references. All models are scored on the same 200 test records under each of nine patterns (Figure~\ref{fig:mixed-control}); actual update counts and comparison scopes are given in \appref{app:mixed-control}.

Against destination-trained references, $R_{\mathrm{MIX}\to w}=E_{\mathrm{MIX},w}/E_{w,w}$ ranges from 0.63 to 4.07 for the four 500-epoch pairs, with respective geometric means 2.07, 2.56, 1.32, and 1.19. The three recipe-matched Poisson destinations give 3.05 (range 2.51--3.49). Navier--Stokes MIX outperforms its destination-trained references on R50, LI, and V; thus a matched-pattern penalty is not universal. Across all 45 pair--test-pattern combinations, MIX is below the median of the eight transferred single-pattern models in 45/45 cases and below their best in 34/45; their transfer ratios reach 1,040.09. Mixed-pattern training reduces large transfer errors in these completed cases, but this descriptive comparison does not isolate the effects of training exposure from heterogeneous histories or establish a general architecture ranking.
\end{newrevision}

The complete results in \appref{app:distributions} include all 49 training-pattern-by-test-pattern matrices, per-cell mean $\pm$ sample standard deviation across 200 test records, pair-level $D$ and $C$, error distributions, three forecasting horizons, and training epochs. Full-precision scores and per-record errors accompany the benchmark, supporting further analysis of observation-pattern sensitivity.

\section{\revnew{Discussion and Conclusion}}
\label{sec:conclusion}
\begin{newrevision}
\dataset{} provides a common protocol for reconstruction and short-horizon forecasting under changing observations. All 49 PDE--method pairs have higher cross-pattern than matched mean error, including the configuration-matched subset; layout and density comparisons motivate reporting both. Five completed MIX models reduce large transfer errors, but do not uniformly beat destination-trained references. Stationary diagnostics locate error at observed points while preserving raw headline scores (\appref{app:controls}). The empirical scope is limited to selected smooth fields, boundary protocols, noiseless observations, coercive Helmholtz cases, and short horizons. Retrospective selection, one training seed, heterogeneous budgets, capacities and information access preclude causal attribution or a controlled architecture ranking. Bootstrap intervals describe test-identity uncertainty only. Formal simple baselines and forecasting ambiguity controls remain untested, and production residuals do not establish independent reference-solution accuracy.
\end{newrevision}

\clearpage

\section*{Ethics Statement}

This study uses synthetic data generated by numerical PDE simulations rather than human-subject records. The benchmark supports research on physical-field reconstruction and forecasting under partial observations. Applications to real-world sensing or control require application-specific validation and safety assessment. Data licensing and release permissions require separate verification.

\section*{AI Use Statement}

Generative AI tools assisted with implementation, suggestions for experimental protocols and parameter settings, analysis, figure and table preparation, literature search, and manuscript revision, including reviewer-style feedback. Reported numerical results are computed from numerical simulations, model predictions, and recorded evaluation outputs. Prediction and artifact verification procedures are described in \appref{app:validity}. The authors take full responsibility for the scientific claims, experimental results, citations, manuscript content, released artifacts, and release permissions.

\bibliographystyle{styles/iclr2027_conference}

\begingroup
    \raggedright
    \bibliography{references,references_added}

\begin{thebibliography}{84}
\providecommand{\natexlab}[1]{#1}
\providecommand{\url}[1]{\texttt{#1}}
\expandafter\ifx\csname urlstyle\endcsname\relax
  \providecommand{\doi}[1]{doi: #1}\else
  \providecommand{\doi}{doi: \begingroup \urlstyle{rm}\Url}\fi

\bibitem[Azizzadenesheli et~al.(2024)Azizzadenesheli, Kovachki, Li,
  Liu-Schiaffini, Kossaifi, and Anandkumar]{azizzadenesheli2024operators}
Kamyar Azizzadenesheli, Nikola Kovachki, Zongyi Li, Miguel Liu-Schiaffini, Jean
  Kossaifi, and Anima Anandkumar.
\newblock {Neural operators for accelerating scientific simulations and
  design}.
\newblock \emph{Nature Reviews Physics}, 6\penalty0 (5):\penalty0 320--328,
  2024.
\newblock \doi{10.1038/s42254-024-00712-5}.

\bibitem[Bonnet et~al.(2022)Bonnet, Mazari, Cinnella, and
  Gallinari]{bonnet2022airfrans}
Florent Bonnet, Jocelyn Mazari, Paola Cinnella, and Patrick Gallinari.
\newblock {AirfRANS: High Fidelity Computational Fluid Dynamics Dataset for
  Approximating Reynolds-Averaged Navier--Stokes Solutions}.
\newblock In \emph{Advances in Neural Information Processing Systems},
  volume~35, pp.\  23463--23478, 2022.
\newblock URL
  \url{https://proceedings.neurips.cc/paper_files/paper/2022/hash/94ab7b23a345f93333eac8748a66c763-Abstract-Datasets_and_Benchmarks.html}.

\bibitem[Boyd(2001)]{boyd2001spectral}
John~P. Boyd.
\newblock \emph{{Chebyshev and Fourier Spectral Methods}}.
\newblock Dover Publications, second edition, 2001.
\newblock ISBN 9780486411835.

\bibitem[Brajard et~al.(2020)Brajard, Carrassi, Bocquet, and
  Bertino]{brajard2020assimilation}
Julien Brajard, Alberto Carrassi, Marc Bocquet, and Laurent Bertino.
\newblock {Combining data assimilation and machine learning to emulate a
  dynamical model from sparse and noisy observations: A case study with the
  Lorenz 96 model}.
\newblock \emph{Journal of Computational Science}, 44:\penalty0 101171, 2020.
\newblock \doi{10.1016/j.jocs.2020.101171}.

\bibitem[Brunton et~al.(2020)Brunton, Noack, and
  Koumoutsakos]{brunton2020fluid}
Steven~L. Brunton, Bernd~R. Noack, and Petros Koumoutsakos.
\newblock {Machine Learning for Fluid Mechanics}.
\newblock \emph{Annual Review of Fluid Mechanics}, 52\penalty0 (1):\penalty0
  477--508, 2020.
\newblock \doi{10.1146/annurev-fluid-010719-060214}.

\bibitem[Callaham et~al.(2019)Callaham, Maeda, and Brunton]{callaham2019robust}
Jared~L. Callaham, Kazuki Maeda, and Steven~L. Brunton.
\newblock {Robust flow reconstruction from limited measurements via sparse
  representation}.
\newblock \emph{Physical Review Fluids}, 4\penalty0 (10):\penalty0 103907,
  2019.
\newblock \doi{10.1103/PhysRevFluids.4.103907}.

\bibitem[Candes et~al.(2006)Candes, Romberg, and Tao]{candes2006robust}
E.J. Candes, J.~Romberg, and T.~Tao.
\newblock {Robust uncertainty principles: exact signal reconstruction from
  highly incomplete frequency information}.
\newblock \emph{IEEE Transactions on Information Theory}, 52\penalty0
  (2):\penalty0 489--509, 2006.
\newblock \doi{10.1109/TIT.2005.862083}.

\bibitem[Carrassi et~al.(2018)Carrassi, Bocquet, Bertino, and
  Evensen]{carrassi2018assimilation}
Alberto Carrassi, Marc Bocquet, Laurent Bertino, and Geir Evensen.
\newblock {Data assimilation in the geosciences: An overview of methods,
  issues, and perspectives}.
\newblock \emph{WIREs Climate Change}, 9\penalty0 (5):\penalty0 e535, 2018.
\newblock \doi{10.1002/wcc.535}.

\bibitem[Chaturantabut \& Sorensen(2010)Chaturantabut and
  Sorensen]{chaturantabut2010deim}
Saifon Chaturantabut and Danny~C. Sorensen.
\newblock {Nonlinear Model Reduction via Discrete Empirical Interpolation}.
\newblock \emph{SIAM Journal on Scientific Computing}, 32\penalty0
  (5):\penalty0 2737--2764, 2010.
\newblock \doi{10.1137/090766498}.

\bibitem[Chung et~al.(2023)Chung, Akoush, Sharma, Tamkin, Jung, Chen, Guo,
  Brouzet, Talei, Savard, Poludnenko, and Ihme]{chung2023blastnet}
Wai~Tong Chung, Bassem Akoush, Pushan Sharma, Alex Tamkin, Ki~Sung Jung,
  Jacqueline Chen, Jack Guo, Davy Brouzet, Mohsen Talei, Bruno Savard, Alexei
  Poludnenko, and Matthias Ihme.
\newblock {Turbulence in Focus: Benchmarking Scaling Behavior of 3D Volumetric
  Super-Resolution with BLASTNet 2.0 Data}.
\newblock In \emph{Advances in Neural Information Processing Systems},
  volume~36, pp.\  77430--77484, 2023.
\newblock URL
  \url{https://proceedings.neurips.cc/paper_files/paper/2023/hash/f458af2455b1e12608c2a16c308d663d-Abstract-Datasets_and_Benchmarks.html}.

\bibitem[Clark et~al.(2019)Clark, Askham, Brunton, and Kutz]{clark2019greedy}
Emily Clark, Travis Askham, Steven~L. Brunton, and J.~Nathan Kutz.
\newblock {Greedy Sensor Placement With Cost Constraints}.
\newblock \emph{IEEE Sensors Journal}, 19\penalty0 (7):\penalty0 2642--2656,
  2019.
\newblock \doi{10.1109/JSEN.2018.2887044}.

\bibitem[Collette(2013)]{collette2013hdf5}
Andrew Collette.
\newblock \emph{{Python and HDF5: Unlocking Scientific Data}}.
\newblock O'Reilly Media, 2013.
\newblock ISBN 9781491944981.

\bibitem[Crank \& Nicolson(1947)Crank and Nicolson]{crank1947practical}
J.~Crank and P.~Nicolson.
\newblock {A practical method for numerical evaluation of solutions of partial
  differential equations of the heat-conduction type}.
\newblock \emph{Mathematical Proceedings of the Cambridge Philosophical
  Society}, 43\penalty0 (1):\penalty0 50--67, 1947.
\newblock \doi{10.1017/S0305004100023197}.

\bibitem[Donoho(2006)]{donoho2006compressed}
D.L. Donoho.
\newblock {Compressed sensing}.
\newblock \emph{IEEE Transactions on Information Theory}, 52\penalty0
  (4):\penalty0 1289--1306, 2006.
\newblock \doi{10.1109/TIT.2006.871582}.

\bibitem[Drma{\v{c}} \& Gugercin(2016)Drma{\v{c}} and
  Gugercin]{drmac2016selection}
Zlatko Drma{\v{c}} and Serkan Gugercin.
\newblock {A New Selection Operator for the Discrete Empirical Interpolation
  Method---Improved A Priori Error Bound and Extensions}.
\newblock \emph{SIAM Journal on Scientific Computing}, 38\penalty0
  (2):\penalty0 A631--A648, 2016.
\newblock \doi{10.1137/15M1019271}.

\bibitem[Dubois et~al.(2022)Dubois, Gomez, Planckaert, and
  Perret]{dubois2022reconstruction}
Pierre Dubois, Thomas Gomez, Laurent Planckaert, and Laurent Perret.
\newblock {Machine learning for fluid flow reconstruction from limited
  measurements}.
\newblock \emph{Journal of Computational Physics}, 448:\penalty0 110733, 2022.
\newblock \doi{10.1016/j.jcp.2021.110733}.

\bibitem[Erichson et~al.(2020)Erichson, Mathelin, Yao, Brunton, Mahoney, and
  Kutz]{erichson2020shallow}
N.~Benjamin Erichson, Lionel Mathelin, Zhewei Yao, Steven~L. Brunton,
  Michael~W. Mahoney, and J.~Nathan Kutz.
\newblock {Shallow neural networks for fluid flow reconstruction with limited
  sensors}.
\newblock \emph{Proceedings of the Royal Society A: Mathematical, Physical and
  Engineering Sciences}, 476\penalty0 (2238):\penalty0 20200097, 2020.
\newblock \doi{10.1098/rspa.2020.0097}.

\bibitem[Evensen(1994)]{evensen1994assimilation}
Geir Evensen.
\newblock {Sequential data assimilation with a nonlinear quasi-geostrophic
  model using Monte Carlo methods to forecast error statistics}.
\newblock \emph{Journal of Geophysical Research: Oceans}, 99\penalty0
  (C5):\penalty0 10143--10162, 1994.
\newblock \doi{10.1029/94JC00572}.

\bibitem[Everson \& Sirovich(1995)Everson and Sirovich]{everson1995gappypod}
Richard Everson and Lawrence Sirovich.
\newblock {Karhunen--Lo{\`e}ve} procedure for gappy data.
\newblock \emph{Journal of the Optical Society of America A}, 12\penalty0
  (8):\penalty0 1657--1664, 1995.
\newblock \doi{10.1364/JOSAA.12.001657}.
\newblock URL \url{https://doi.org/10.1364/JOSAA.12.001657}.

\bibitem[Fukami et~al.(2019)Fukami, Fukagata, and
  Taira]{fukami2019superresolution}
Kai Fukami, Koji Fukagata, and Kunihiko Taira.
\newblock {Super-resolution reconstruction of turbulent flows with machine
  learning}.
\newblock \emph{Journal of Fluid Mechanics}, 870:\penalty0 106--120, 2019.
\newblock \doi{10.1017/jfm.2019.238}.

\bibitem[Fukami et~al.(2021)Fukami, Maulik, Ramachandra, Fukagata, and
  Taira]{fukami2021voronoi}
Kai Fukami, Romit Maulik, Nesar Ramachandra, Koji Fukagata, and Kunihiko Taira.
\newblock {Global field reconstruction from sparse sensors with Voronoi
  tessellation-assisted deep learning}.
\newblock \emph{Nature Machine Intelligence}, 3\penalty0 (11):\penalty0
  945--951, 2021.
\newblock \doi{10.1038/s42256-021-00402-2}.

\bibitem[Fukami et~al.(2023)Fukami, Fukagata, and Taira]{fukami2023survey}
Kai Fukami, Koji Fukagata, and Kunihiko Taira.
\newblock {Super-resolution analysis via machine learning: a survey for fluid
  flows}.
\newblock \emph{Theoretical and Computational Fluid Dynamics}, 37\penalty0
  (4):\penalty0 421--444, 2023.
\newblock \doi{10.1007/s00162-023-00663-0}.

\bibitem[Goel et~al.(2026)Goel, Chalapathi, Raja, and
  Krishnapriyan]{goel2026pdeinvbench}
Divyam Goel, Nithin Chalapathi, Sanjeev Raja, and Aditi~S. Krishnapriyan.
\newblock {PDEInvBench}: A comprehensive dataset and design space exploration
  of neural networks for {PDE} inverse problems.
\newblock \emph{Transactions on Machine Learning Research}, 2026.
\newblock ISSN 2835-8856.
\newblock URL \url{https://openreview.net/forum?id=MSjhqRnNyZ}.

\bibitem[Gottlieb \& Shu(1998)Gottlieb and Shu]{gottlieb1998tvd}
Sigal Gottlieb and Chi-Wang Shu.
\newblock {Total variation diminishing Runge-Kutta schemes}.
\newblock \emph{Mathematics of Computation}, 67\penalty0 (221):\penalty0
  73--85, 1998.
\newblock \doi{10.1090/S0025-5718-98-00913-2}.

\bibitem[Gulrajani \& Lopez-Paz(2021)Gulrajani and
  Lopez-Paz]{gulrajani2021domainbed}
Ishaan Gulrajani and David Lopez-Paz.
\newblock {In Search of Lost Domain Generalization}.
\newblock In \emph{International Conference on Learning Representations}, 2021.
\newblock URL \url{https://openreview.net/forum?id=lQdXeXDoWtI}.

\bibitem[Gunes et~al.(2006)Gunes, Sirisup, and Karniadakis]{gunes2006gappy}
Hasan Gunes, Sirod Sirisup, and George~Em Karniadakis.
\newblock {Gappy data: To Krig or not to Krig?}
\newblock \emph{Journal of Computational Physics}, 212\penalty0 (1):\penalty0
  358--382, 2006.
\newblock \doi{10.1016/j.jcp.2005.06.023}.

\bibitem[Gupta \& Brandstetter(2023)Gupta and Brandstetter]{gupta2022pdearena}
Jayesh~K. Gupta and Johannes Brandstetter.
\newblock Towards multi-spatiotemporal-scale generalized {PDE} modeling.
\newblock \emph{Transactions on Machine Learning Research}, 2023.
\newblock ISSN 2835-8856.
\newblock URL \url{https://openreview.net/forum?id=dPSTDbGtBY}.

\bibitem[Hao et~al.(2023)Hao, Wang, Su, Ying, Dong, Liu, Cheng, Song, and
  Zhu]{hao2023gnot}
Zhongkai Hao, Zhengyi Wang, Hang Su, Chengyang Ying, Yinpeng Dong, Songming
  Liu, Ze~Cheng, Jian Song, and Jun Zhu.
\newblock {GNOT}: A general neural operator transformer for operator learning.
\newblock In \emph{Proceedings of the 40th International Conference on Machine
  Learning}, volume 202 of \emph{Proceedings of Machine Learning Research},
  pp.\  12556--12569. PMLR, 2023.
\newblock URL \url{https://proceedings.mlr.press/v202/hao23c.html}.

\bibitem[Harris et~al.(2020)Harris, Millman, van~der Walt, Gommers, Virtanen,
  Cournapeau, Wieser, Taylor, Berg, Smith, Kern, Picus, Hoyer, van Kerkwijk,
  Brett, Haldane, del R{\'{i}}o, Wiebe, Peterson, G{\'{e}}rard-Marchant,
  Sheppard, Reddy, Weckesser, Abbasi, Gohlke, and Oliphant]{harris2020numpy}
Charles~R. Harris, K.~Jarrod Millman, St{\'{e}}fan~J. van~der Walt, Ralf
  Gommers, Pauli Virtanen, David Cournapeau, Eric Wieser, Julian Taylor,
  Sebastian Berg, Nathaniel~J. Smith, Robert Kern, Matti Picus, Stephan Hoyer,
  Marten~H. van Kerkwijk, Matthew Brett, Allan Haldane, Jaime~Fern{\'{a}}ndez
  del R{\'{i}}o, Mark Wiebe, Pearu Peterson, Pierre G{\'{e}}rard-Marchant,
  Kevin Sheppard, Tyler Reddy, Warren Weckesser, Hameer Abbasi, Christoph
  Gohlke, and Travis~E. Oliphant.
\newblock {Array programming with NumPy}.
\newblock \emph{Nature}, 585\penalty0 (7825):\penalty0 357--362, 2020.
\newblock \doi{10.1038/s41586-020-2649-2}.

\bibitem[Herde et~al.(2024)Herde, Raoni{\'{c}}, Rohner, K{\"{a}}ppeli,
  Molinaro, de~B{\'{e}}zenac, and Mishra]{herde2024poseidon}
Maximilian Herde, Bogdan Raoni{\'{c}}, Tobias Rohner, Roger K{\"{a}}ppeli,
  Roberto Molinaro, Emmanuel de~B{\'{e}}zenac, and Siddhartha Mishra.
\newblock {Poseidon: Efficient Foundation Models for PDEs}.
\newblock In \emph{Advances in Neural Information Processing Systems},
  volume~37, pp.\  72525--72624, 2024.
\newblock URL
  \url{https://proceedings.neurips.cc/paper_files/paper/2024/hash/84e1b1ec17bb11c57234e96433022a9a-Abstract-Conference.html}.

\bibitem[Hestenes \& Stiefel(1952)Hestenes and Stiefel]{hestenes1952cg}
M.R. Hestenes and E.~Stiefel.
\newblock {Methods of conjugate gradients for solving linear systems}.
\newblock \emph{Journal of Research of the National Bureau of Standards},
  49\penalty0 (6):\penalty0 409--436, 1952.
\newblock \doi{10.6028/jres.049.044}.

\bibitem[Hou et~al.(2026)Hou, Wang, Xu, Gao, Liu, and Jing]{hou2026pobench}
Jingren Hou, Hong Wang, Pengyu Xu, Chang Gao, Huafeng Liu, and Liping Jing.
\newblock Learning neural operators from partial observations via latent
  autoregressive modeling.
\newblock \emph{Proceedings of the AAAI Conference on Artificial Intelligence},
  40\penalty0 (1):\penalty0 390--398, 2026.
\newblock \doi{10.1609/aaai.v40i1.37001}.
\newblock URL \url{https://ojs.aaai.org/index.php/AAAI/article/view/37001}.

\bibitem[Hu et~al.(2026)Hu, Feng, Liu, Yan, Deng, Gao, Zheng, Zheng, Yu, Wang,
  Li, Ma, Zhou, Lu, Fan, and Wu]{hu2026realpdebench}
Peiyan Hu, Haodong Feng, Hongyuan Liu, Tongtong Yan, Wenhao Deng, Tianrun Gao,
  Rong Zheng, Haoren Zheng, Chenglei Yu, Chuanrui Wang, Kaiwen Li, Zhi-Ming Ma,
  Dezhi Zhou, Xingcai Lu, Dixia Fan, and Tailin Wu.
\newblock {RealPDEBench}: A benchmark for complex physical systems with
  real-world data.
\newblock In \emph{The Fourteenth International Conference on Learning
  Representations}, 2026.
\newblock URL \url{https://openreview.net/forum?id=y3oHMcoItR}.
\newblock Oral Presentation.

\bibitem[Huang et~al.(2024)Huang, Yang, Wang, and Park]{huang2024diffusionpde}
Jiahe Huang, Guandao Yang, Zichen Wang, and Jeong~Joon Park.
\newblock {DiffusionPDE}: Generative {PDE}-solving under partial observation.
\newblock In \emph{Advances in Neural Information Processing Systems},
  volume~37, pp.\  130291--130323, 2024.
\newblock \doi{10.52202/079017-4140}.
\newblock URL
  \url{https://proceedings.neurips.cc/paper_files/paper/2024/hash/eb3878c1dcbfff9ee95d5d033e5f5942-Abstract-Conference.html}.

\bibitem[Jin et~al.(2021)Jin, Cai, Li, and Karniadakis]{jin2021nsfnets}
Xiaowei Jin, Shengze Cai, Hui Li, and George~Em Karniadakis.
\newblock {NSFnets (Navier-Stokes flow nets): Physics-informed neural networks
  for the incompressible Navier-Stokes equations}.
\newblock \emph{Journal of Computational Physics}, 426:\penalty0 109951, 2021.
\newblock \doi{10.1016/j.jcp.2020.109951}.

\bibitem[Johnson et~al.(2023)Johnson, Febvre, Gorbunova, Metref, Ballarotta,
  Le~Sommer, and Fablet]{johnson2023oceanbench}
J.~Emmanuel Johnson, Quentin Febvre, Anastasiia Gorbunova, Sam Metref, Maxime
  Ballarotta, Julien Le~Sommer, and Ronan Fablet.
\newblock {OceanBench: The Sea Surface Height Edition}.
\newblock In \emph{Advances in Neural Information Processing Systems},
  volume~36, pp.\  78275--78295, 2023.
\newblock URL
  \url{https://proceedings.neurips.cc/paper_files/paper/2023/hash/f6ccbf94fa57c2ae372ece91b537574d-Abstract-Datasets_and_Benchmarks.html}.

\bibitem[Joshi \& Boyd(2009)Joshi and Boyd]{joshi2009sensors}
S.~Joshi and S.~Boyd.
\newblock {Sensor Selection via Convex Optimization}.
\newblock \emph{IEEE Transactions on Signal Processing}, 57\penalty0
  (2):\penalty0 451--462, 2009.
\newblock \doi{10.1109/TSP.2008.2007095}.

\bibitem[Karniadakis et~al.(2021)Karniadakis, Kevrekidis, Lu, Perdikaris, Wang,
  and Yang]{karniadakis2021physics}
George~Em Karniadakis, Ioannis~G. Kevrekidis, Lu~Lu, Paris Perdikaris, Sifan
  Wang, and Liu Yang.
\newblock {Physics-informed machine learning}.
\newblock \emph{Nature Reviews Physics}, 3\penalty0 (6):\penalty0 422--440,
  2021.
\newblock \doi{10.1038/s42254-021-00314-5}.

\bibitem[Kingma \& Ba(2015)Kingma and Ba]{kingma2015adam}
Diederik~P. Kingma and Jimmy Ba.
\newblock {Adam: A Method for Stochastic Optimization}.
\newblock In \emph{International Conference on Learning Representations}, 2015.
\newblock URL \url{https://arxiv.org/abs/1412.6980}.

\bibitem[Koehler et~al.(2024)Koehler, Niedermayr, Westermann, and
  Thuerey]{koehler2024apebench}
Felix Koehler, Simon Niedermayr, R{\"u}diger Westermann, and Nils Thuerey.
\newblock {APEBench}: A benchmark for autoregressive neural emulators of
  {PDEs}.
\newblock In \emph{Advances in Neural Information Processing Systems},
  volume~37, pp.\  120252--120310, 2024.
\newblock \doi{10.52202/079017-3822}.
\newblock URL
  \url{https://proceedings.neurips.cc/paper_files/paper/2024/hash/d9875ebcf74bccdc5076acab0dbee62c-Abstract-Datasets_and_Benchmarks_Track.html}.

\bibitem[Koh et~al.(2021)Koh, Sagawa, Marklund, Xie, Zhang, Balsubramani, Hu,
  Yasunaga, Phillips, Gao, Lee, David, Stavness, Guo, Earnshaw, Haque, Beery,
  Leskovec, Kundaje, Pierson, Levine, Finn, and Liang]{koh2021wilds}
Pang~Wei Koh, Shiori Sagawa, Henrik Marklund, Sang~Michael Xie, Marvin Zhang,
  Akshay Balsubramani, Weihua Hu, Michihiro Yasunaga, Richard~Lanas Phillips,
  Irena Gao, Tony Lee, Etienne David, Ian Stavness, Wei Guo, Berton Earnshaw,
  Imran Haque, Sara~M Beery, Jure Leskovec, Anshul Kundaje, Emma Pierson,
  Sergey Levine, Chelsea Finn, and Percy Liang.
\newblock {WILDS: A Benchmark of in-the-Wild Distribution Shifts}.
\newblock In \emph{International Conference on Machine Learning}, volume 139 of
  \emph{Proceedings of Machine Learning Research}, pp.\  5637--5664. PMLR,
  2021.
\newblock URL \url{https://proceedings.mlr.press/v139/koh21a.html}.

\bibitem[Kovachki et~al.(2023)Kovachki, Li, Liu, Azizzadenesheli, Bhattacharya,
  Stuart, and Anandkumar]{kovachki2023operator}
Nikola Kovachki, Zongyi Li, Burigede Liu, Kamyar Azizzadenesheli, Kaushik
  Bhattacharya, Andrew Stuart, and Anima Anandkumar.
\newblock {Neural Operator: Learning Maps Between Function Spaces With
  Applications to PDEs}.
\newblock \emph{Journal of Machine Learning Research}, 24\penalty0
  (89):\penalty0 1--97, 2023.
\newblock URL \url{https://jmlr.org/papers/v24/21-1524.html}.

\bibitem[Krause et~al.(2008)Krause, Singh, and Guestrin]{krause2008sensors}
Andreas Krause, Ajit Singh, and Carlos Guestrin.
\newblock {Near-Optimal Sensor Placements in Gaussian Processes: Theory,
  Efficient Algorithms and Empirical Studies}.
\newblock \emph{Journal of Machine Learning Research}, 9\penalty0 (8):\penalty0
  235--284, 2008.
\newblock URL \url{https://jmlr.org/papers/v9/krause08a.html}.

\bibitem[LeVeque(2007)]{leveque2007finite}
Randall~J. LeVeque.
\newblock \emph{{Finite Difference Methods for Ordinary and Partial
  Differential Equations}}.
\newblock Society for Industrial and Applied Mathematics, 2007.
\newblock \doi{10.1137/1.9780898717839}.

\bibitem[Li et~al.(2021)Li, Kovachki, Azizzadenesheli, Liu, Bhattacharya,
  Stuart, and Anandkumar]{li2020fno}
Zongyi Li, Nikola Kovachki, Kamyar Azizzadenesheli, Burigede Liu, Kaushik
  Bhattacharya, Andrew Stuart, and Anima Anandkumar.
\newblock {Fourier} neural operator for parametric partial differential
  equations.
\newblock In \emph{International Conference on Learning Representations}, 2021.
\newblock URL \url{https://openreview.net/forum?id=c8P9NQVtmnO}.

\bibitem[Li et~al.(2024)Li, Zheng, Kovachki, Jin, Chen, Liu, Azizzadenesheli,
  and Anandkumar]{li2021pino}
Zongyi Li, Hongkai Zheng, Nikola Kovachki, David Jin, Haoxuan Chen, Burigede
  Liu, Kamyar Azizzadenesheli, and Anima Anandkumar.
\newblock Physics-informed neural operator for learning partial differential
  equations.
\newblock \emph{ACM / IMS Journal of Data Science}, 1\penalty0 (3):\penalty0
  1--27, 2024.
\newblock \doi{10.1145/3648506}.
\newblock URL \url{https://doi.org/10.1145/3648506}.

\bibitem[Loshchilov \& Hutter(2019)Loshchilov and Hutter]{loshchilov2019adamw}
Ilya Loshchilov and Frank Hutter.
\newblock {Decoupled Weight Decay Regularization}.
\newblock In \emph{International Conference on Learning Representations}, 2019.
\newblock URL \url{https://openreview.net/forum?id=Bkg6RiCqY7}.

\bibitem[Lu et~al.(2021)Lu, Jin, Pang, Zhang, and Karniadakis]{lu2021deeponet}
Lu~Lu, Pengzhan Jin, Guofei Pang, Zhongqiang Zhang, and George~Em Karniadakis.
\newblock Learning nonlinear operators via {DeepONet} based on the universal
  approximation theorem of operators.
\newblock \emph{Nature Machine Intelligence}, 3:\penalty0 218--229, 2021.
\newblock \doi{10.1038/s42256-021-00302-5}.
\newblock URL \url{https://doi.org/10.1038/s42256-021-00302-5}.

\bibitem[Luo et~al.(2023)Luo, Chen, and Zhang]{luo2023cfdbench}
Yining Luo, Yingfa Chen, and Zhen Zhang.
\newblock {CFDBench: A Large-Scale Benchmark for Machine Learning Methods in
  Fluid Dynamics}.
\newblock \emph{arXiv preprint arXiv:2310.05963}, 2023.
\newblock URL \url{https://arxiv.org/abs/2310.05963}.

\bibitem[Manohar et~al.(2018)Manohar, Brunton, Kutz, and
  Brunton]{manohar2018sensors}
Krithika Manohar, Bingni~W. Brunton, J.~Nathan Kutz, and Steven~L. Brunton.
\newblock {Data-Driven Sparse Sensor Placement for Reconstruction:
  Demonstrating the Benefits of Exploiting Known Patterns}.
\newblock \emph{IEEE Control Systems}, 38\penalty0 (3):\penalty0 63--86, 2018.
\newblock \doi{10.1109/MCS.2018.2810460}.

\bibitem[Musekamp et~al.(2025)Musekamp, Kalimuthu, Holzm{\"u}ller, Takamoto,
  and Niepert]{musekamp2024al4pde}
Daniel Musekamp, Marimuthu Kalimuthu, David Holzm{\"u}ller, Makoto Takamoto,
  and Mathias Niepert.
\newblock Active learning for neural {PDE} solvers.
\newblock In \emph{The Thirteenth International Conference on Learning
  Representations}. OpenReview.net, 2025.
\newblock URL \url{https://openreview.net/forum?id=x4ZmQaumRg}.

\bibitem[Ohana et~al.(2024)Ohana, McCabe, Meyer, Morel, Agocs, Beneitez,
  Berger, Burkhart, Dalziel, Fielding, Fortunato, Goldberg, Hirashima, Jiang,
  Kerswell, Maddu, Miller, Mukhopadhyay, Nixon, Shen, Watteaux,
  R{\'e}galdo-Saint~Blancard, Rozet, Parker, Cranmer, and Ho]{ohana2024well}
Ruben Ohana, Michael McCabe, Lucas Meyer, Rudy Morel, Fruzsina~J. Agocs, Miguel
  Beneitez, Marsha Berger, Blakesley Burkhart, Stuart~B. Dalziel, Drummond~B.
  Fielding, Daniel Fortunato, Jared~A. Goldberg, Keiya Hirashima, Yan-Fei
  Jiang, Rich~R. Kerswell, Suryanarayana Maddu, Jonah Miller, Payel
  Mukhopadhyay, Stefan~S. Nixon, Jeff Shen, Romain Watteaux, Bruno
  R{\'e}galdo-Saint~Blancard, Fran{\c c}ois Rozet, Liam~H. Parker, Miles
  Cranmer, and Shirley Ho.
\newblock {The Well}: a large-scale collection of diverse physics simulations
  for machine learning.
\newblock In \emph{Advances in Neural Information Processing Systems},
  volume~37, pp.\  44989--45037, 2024.
\newblock \doi{10.52202/079017-1430}.
\newblock URL
  \url{https://proceedings.neurips.cc/paper_files/paper/2024/hash/4f9a5acd91ac76569f2fe291b1f4772b-Abstract-Datasets_and_Benchmarks_Track.html}.

\bibitem[Otness et~al.(2021)Otness, Gjoka, Bruna, Panozzo, Peherstorfer,
  Schneider, and Zorin]{otness2021extensible}
Karl Otness, Arvi Gjoka, Joan Bruna, Daniele Panozzo, Benjamin Peherstorfer,
  Teseo Schneider, and Denis Zorin.
\newblock {An Extensible Benchmark Suite for Learning to Simulate Physical
  Systems}.
\newblock In \emph{Proceedings of the Neural Information Processing Systems
  Track on Datasets and Benchmarks}, volume~1, 2021.
\newblock URL
  \url{https://datasets-benchmarks-proceedings.neurips.cc/paper_files/paper/2021/hash/3def184ad8f4755ff269862ea77393dd-Abstract-round1.html}.

\bibitem[Paige \& Saunders(1975)Paige and Saunders]{paige1975minres}
C.~C. Paige and M.~A. Saunders.
\newblock {Solution of Sparse Indefinite Systems of Linear Equations}.
\newblock \emph{SIAM Journal on Numerical Analysis}, 12\penalty0 (4):\penalty0
  617--629, 1975.
\newblock \doi{10.1137/0712047}.

\bibitem[Paszke et~al.(2019)Paszke, Gross, Massa, Lerer, Bradbury, Chanan,
  Killeen, Lin, Gimelshein, Antiga, Desmaison, K{\"{o}}pf, Yang, DeVito,
  Raison, Tejani, Chilamkurthy, Steiner, Fang, Bai, and
  Chintala]{paszke2019pytorch}
Adam Paszke, Sam Gross, Francisco Massa, Adam Lerer, James Bradbury, Gregory
  Chanan, Trevor Killeen, Zeming Lin, Natalia Gimelshein, Luca Antiga, Alban
  Desmaison, Andreas K{\"{o}}pf, Edward Yang, Zach DeVito, Martin Raison,
  Alykhan Tejani, Sasank Chilamkurthy, Benoit Steiner, Lu~Fang, Junjie Bai, and
  Soumith Chintala.
\newblock {PyTorch: An Imperative Style, High-Performance Deep Learning
  Library}.
\newblock In \emph{Advances in Neural Information Processing Systems},
  volume~32, 2019.
\newblock URL \url{https://arxiv.org/abs/1912.01703}.

\bibitem[Patterson \& Orszag(1971)Patterson and Orszag]{patterson1971spectral}
G.~S. Patterson and Steven~A. Orszag.
\newblock {Spectral Calculations of Isotropic Turbulence: Efficient Removal of
  Aliasing Interactions}.
\newblock \emph{The Physics of Fluids}, 14\penalty0 (11):\penalty0 2538--2541,
  1971.
\newblock \doi{10.1063/1.1693365}.

\bibitem[Peherstorfer et~al.(2020)Peherstorfer, Drma{\v{c}}, and
  Gugercin]{peherstorfer2020stability}
Benjamin Peherstorfer, Zlatko Drma{\v{c}}, and Serkan Gugercin.
\newblock {Stability of Discrete Empirical Interpolation and Gappy Proper
  Orthogonal Decomposition with Randomized and Deterministic Sampling Points}.
\newblock \emph{SIAM Journal on Scientific Computing}, 42\penalty0
  (5):\penalty0 A2837--A2864, 2020.
\newblock \doi{10.1137/19M1307391}.

\bibitem[Raissi(2018)]{raissi2018deephidden}
Maziar Raissi.
\newblock {Deep Hidden Physics Models: Deep Learning of Nonlinear Partial
  Differential Equations}.
\newblock \emph{Journal of Machine Learning Research}, 19\penalty0
  (25):\penalty0 1--24, 2018.
\newblock URL \url{https://jmlr.org/papers/v19/18-046.html}.

\bibitem[Raissi et~al.(2018)Raissi, Perdikaris, and
  Karniadakis]{raissi2018numericalgp}
Maziar Raissi, Paris Perdikaris, and George~Em Karniadakis.
\newblock {Numerical Gaussian Processes for Time-Dependent and Nonlinear
  Partial Differential Equations}.
\newblock \emph{SIAM Journal on Scientific Computing}, 40\penalty0
  (1):\penalty0 A172--A198, 2018.
\newblock \doi{10.1137/17M1120762}.

\bibitem[Raissi et~al.(2019)Raissi, Perdikaris, and
  Karniadakis]{raissi2019pinns}
Maziar Raissi, Paris Perdikaris, and George~Em Karniadakis.
\newblock Physics-informed neural networks: A deep learning framework for
  solving forward and inverse problems involving nonlinear partial differential
  equations.
\newblock \emph{Journal of Computational Physics}, 378:\penalty0 686--707,
  2019.
\newblock \doi{10.1016/j.jcp.2018.10.045}.
\newblock URL \url{https://doi.org/10.1016/j.jcp.2018.10.045}.

\bibitem[Raissi et~al.(2020)Raissi, Yazdani, and
  Karniadakis]{raissi2020hiddenfluid}
Maziar Raissi, Alireza Yazdani, and George~Em Karniadakis.
\newblock {Hidden fluid mechanics: Learning velocity and pressure fields from
  flow visualizations}.
\newblock \emph{Science}, 367\penalty0 (6481):\penalty0 1026--1030, 2020.
\newblock \doi{10.1126/science.aaw4741}.

\bibitem[Raoni{\'c} et~al.(2023)Raoni{\'c}, Molinaro, De~Ryck, Rohner,
  Bartolucci, Alaifari, Mishra, and de~B{\'e}zenac]{raonic2023cno}
Bogdan Raoni{\'c}, Roberto Molinaro, Tim De~Ryck, Tobias Rohner, Francesca
  Bartolucci, Rima Alaifari, Siddhartha Mishra, and Emmanuel de~B{\'e}zenac.
\newblock Convolutional neural operators for robust and accurate learning of
  {PDEs}.
\newblock In \emph{Advances in Neural Information Processing Systems},
  volume~36, pp.\  77187--77200, 2023.
\newblock \doi{10.52202/075280-3376}.
\newblock URL
  \url{https://proceedings.neurips.cc/paper_files/paper/2023/hash/f3c1951b34f7f55ffaecada7fde6bd5a-Abstract-Conference.html}.

\bibitem[Rasp et~al.(2020)Rasp, Dueben, Scher, Weyn, Mouatadid, and
  Thuerey]{rasp2020weatherbench}
Stephan Rasp, Peter~D. Dueben, Sebastian Scher, Jonathan~A. Weyn, Soukayna
  Mouatadid, and Nils Thuerey.
\newblock {WeatherBench: A Benchmark Data Set for Data-Driven Weather
  Forecasting}.
\newblock \emph{Journal of Advances in Modeling Earth Systems}, 12\penalty0
  (11):\penalty0 e2020MS002203, 2020.
\newblock \doi{10.1029/2020MS002203}.

\bibitem[Rasp et~al.(2024)Rasp, Hoyer, Merose, Langmore, Battaglia, Russell,
  Sanchez-Gonzalez, Yang, Carver, Agrawal, Chantry, Ben~Bouallegue, Dueben,
  Bromberg, Sisk, Barrington, Bell, and Sha]{rasp2024weatherbench2}
Stephan Rasp, Stephan Hoyer, Alexander Merose, Ian Langmore, Peter Battaglia,
  Tyler Russell, Alvaro Sanchez-Gonzalez, Vivian Yang, Rob Carver, Shreya
  Agrawal, Matthew Chantry, Zied Ben~Bouallegue, Peter Dueben, Carla Bromberg,
  Jared Sisk, Luke Barrington, Aaron Bell, and Fei Sha.
\newblock {WeatherBench 2: A Benchmark for the Next Generation of Data-Driven
  Global Weather Models}.
\newblock \emph{Journal of Advances in Modeling Earth Systems}, 16\penalty0
  (6):\penalty0 e2023MS004019, 2024.
\newblock \doi{10.1029/2023MS004019}.

\bibitem[Ronneberger et~al.(2015)Ronneberger, Fischer, and
  Brox]{ronneberger2015unet}
Olaf Ronneberger, Philipp Fischer, and Thomas Brox.
\newblock {U-Net}: Convolutional networks for biomedical image segmentation.
\newblock In \emph{Medical Image Computing and Computer-Assisted Intervention},
  pp.\  234--241. Springer, 2015.
\newblock \doi{10.1007/978-3-319-24574-4_28}.
\newblock URL \url{https://doi.org/10.1007/978-3-319-24574-4_28}.

\bibitem[Rowley \& Dawson(2017)Rowley and Dawson]{rowley2017modelreduction}
Clarence~W. Rowley and Scott~T.M. Dawson.
\newblock {Model Reduction for Flow Analysis and Control}.
\newblock \emph{Annual Review of Fluid Mechanics}, 49\penalty0 (1):\penalty0
  387--417, 2017.
\newblock \doi{10.1146/annurev-fluid-010816-060042}.

\bibitem[Saad(2003)]{saad2003iterative}
Yousef Saad.
\newblock \emph{{Iterative Methods for Sparse Linear Systems}}.
\newblock Society for Industrial and Applied Mathematics, second edition, 2003.
\newblock \doi{10.1137/1.9780898718003}.

\bibitem[Santos et~al.(2023)Santos, Fox, Mohan, O'Malley, Viswanathan, and
  Lubbers]{santos2023senseiver}
Javier~E. Santos, Zachary~R. Fox, Arvind Mohan, Daniel O'Malley, Hari
  Viswanathan, and Nicholas Lubbers.
\newblock Development of the {Senseiver} for efficient field reconstruction
  from sparse observations.
\newblock \emph{Nature Machine Intelligence}, 5:\penalty0 1317--1325, 2023.
\newblock \doi{10.1038/s42256-023-00746-x}.
\newblock URL \url{https://www.nature.com/articles/s42256-023-00746-x}.

\bibitem[Serrano et~al.(2023)Serrano, Le~Boudec, Kassa{\"i}~Koupa{\"i}, Wang,
  Yin, Vittaut, and Gallinari]{serrano2023coral}
Louis Serrano, Lise Le~Boudec, Armand Kassa{\"i}~Koupa{\"i}, Thomas~X Wang,
  Yuan Yin, Jean-No{\"e}l Vittaut, and Patrick Gallinari.
\newblock Operator learning with neural fields: Tackling {PDEs} on general
  geometries.
\newblock In \emph{Advances in Neural Information Processing Systems},
  volume~36, 2023.
\newblock \doi{10.52202/075280-3093}.
\newblock URL
  \url{https://proceedings.neurips.cc/paper_files/paper/2023/hash/df54302388bbc145aacaa1a54a4a5933-Abstract-Conference.html}.

\bibitem[Smith \& Topin(2019)Smith and Topin]{smith2019superconvergence}
Leslie~N. Smith and Nicholay Topin.
\newblock {Super-convergence: very fast training of neural networks using large
  learning rates}.
\newblock In \emph{Artificial Intelligence and Machine Learning for
  Multi-Domain Operations Applications}. SPIE, 2019.
\newblock \doi{10.1117/12.2520589}.

\bibitem[Strang(1968)]{strang1968splitting}
Gilbert Strang.
\newblock {On the Construction and Comparison of Difference Schemes}.
\newblock \emph{SIAM Journal on Numerical Analysis}, 5\penalty0 (3):\penalty0
  506--517, 1968.
\newblock \doi{10.1137/0705041}.

\bibitem[Stuart(2010)]{stuart2010inverse}
A.~M. Stuart.
\newblock {Inverse problems: A Bayesian perspective}.
\newblock \emph{Acta Numerica}, 19:\penalty0 451--559, 2010.
\newblock \doi{10.1017/S0962492910000061}.

\bibitem[Sun \& Wang(2020)Sun and Wang]{sun2020bayesian}
Luning Sun and Jian-Xun Wang.
\newblock {Physics-constrained bayesian neural network for fluid flow
  reconstruction with sparse and noisy data}.
\newblock \emph{Theoretical and Applied Mechanics Letters}, 10\penalty0
  (3):\penalty0 161--169, 2020.
\newblock \doi{10.1016/j.taml.2020.01.031}.

\bibitem[Takamoto et~al.(2022)Takamoto, Praditia, Leiteritz, MacKinlay,
  Alesiani, Pfl{\"u}ger, and Niepert]{takamoto2022pdebench}
Makoto Takamoto, Timothy Praditia, Raphael Leiteritz, Daniel MacKinlay,
  Francesco Alesiani, Dirk Pfl{\"u}ger, and Mathias Niepert.
\newblock {PDEBench}: An extensive benchmark for scientific machine learning.
\newblock In \emph{Advances in Neural Information Processing Systems},
  volume~35, pp.\  1596--1611, 2022.
\newblock \doi{10.52202/068431-0117}.
\newblock URL
  \url{https://proceedings.neurips.cc/paper_files/paper/2022/hash/0a9747136d411fb83f0cf81820d44afb-Abstract-Datasets_and_Benchmarks.html}.

\bibitem[Trefethen(2000)]{trefethen2000spectral}
Lloyd~N. Trefethen.
\newblock \emph{{Spectral Methods in MATLAB}}.
\newblock Society for Industrial and Applied Mathematics, 2000.
\newblock \doi{10.1137/1.9780898719598}.

\bibitem[Virtanen et~al.(2020)Virtanen, Gommers, Oliphant, Haberland, Reddy,
  Cournapeau, Burovski, Peterson, Weckesser, Bright, van~der Walt, Brett,
  Wilson, Millman, Mayorov, Nelson, Jones, Kern, Larson, Carey, Polat, Feng,
  Moore, VanderPlas, Laxalde, Perktold, Cimrman, Henriksen, Quintero, Harris,
  Archibald, Ribeiro, Pedregosa, van Mulbregt, and {SciPy 1.0
  Contributors}]{virtanen2020scipy}
Pauli Virtanen, Ralf Gommers, Travis~E. Oliphant, Matt Haberland, Tyler Reddy,
  David Cournapeau, Evgeni Burovski, Pearu Peterson, Warren Weckesser, Jonathan
  Bright, St{\'{e}}fan~J. van~der Walt, Matthew Brett, Joshua Wilson, K.~Jarrod
  Millman, Nikolay Mayorov, Andrew R.~J. Nelson, Eric Jones, Robert Kern, Eric
  Larson, C~J Carey, {\.{I}}lhan Polat, Yu~Feng, Eric~W. Moore, Jake
  VanderPlas, Denis Laxalde, Josef Perktold, Robert Cimrman, Ian Henriksen,
  E.~A. Quintero, Charles~R. Harris, Anne~M. Archibald, Ant{\^{o}}nio~H.
  Ribeiro, Fabian Pedregosa, Paul van Mulbregt, and {SciPy 1.0 Contributors}.
\newblock {SciPy 1.0: fundamental algorithms for scientific computing in
  Python}.
\newblock \emph{Nature Methods}, 17\penalty0 (3):\penalty0 261--272, 2020.
\newblock \doi{10.1038/s41592-019-0686-2}.

\bibitem[Wen et~al.(2022)Wen, Li, Azizzadenesheli, Anandkumar, and
  Benson]{wen2022ufno}
Gege Wen, Zongyi Li, Kamyar Azizzadenesheli, Anima Anandkumar, and Sally~M.
  Benson.
\newblock {U-FNO}: An enhanced {Fourier} neural operator-based deep-learning
  model for multiphase flow.
\newblock \emph{Advances in Water Resources}, 163:\penalty0 104180, 2022.
\newblock \doi{10.1016/j.advwatres.2022.104180}.
\newblock URL \url{https://doi.org/10.1016/j.advwatres.2022.104180}.

\bibitem[Willard et~al.(2022)Willard, Jia, Xu, Steinbach, and
  Kumar]{willard2022integrating}
Jared Willard, Xiaowei Jia, Shaoming Xu, Michael Steinbach, and Vipin Kumar.
\newblock {Integrating Scientific Knowledge with Machine Learning for
  Engineering and Environmental Systems}.
\newblock \emph{ACM Computing Surveys}, 55\penalty0 (4):\penalty0 1--37, 2022.
\newblock \doi{10.1145/3514228}.

\bibitem[Willcox(2006)]{willcox2006gappy}
K.~Willcox.
\newblock {Unsteady flow sensing and estimation via the gappy proper orthogonal
  decomposition}.
\newblock \emph{Computers \& Fluids}, 35\penalty0 (2):\penalty0 208--226, 2006.
\newblock \doi{10.1016/j.compfluid.2004.11.006}.

\bibitem[Williams et~al.(2024)Williams, Zahn, and Kutz]{williams2024shred}
Jan~P. Williams, Olivia Zahn, and J.~Nathan Kutz.
\newblock {Sensing with shallow recurrent decoder networks}.
\newblock \emph{Proceedings of the Royal Society A: Mathematical, Physical and
  Engineering Sciences}, 480\penalty0 (2298):\penalty0 20240054, 2024.
\newblock \doi{10.1098/rspa.2024.0054}.

\bibitem[Wu et~al.(2024)Wu, Luo, Wang, Wang, and Long]{wu2024transolver}
Haixu Wu, Huakun Luo, Haowen Wang, Jianmin Wang, and Mingsheng Long.
\newblock Transolver: A fast transformer solver for {PDEs} on general
  geometries.
\newblock In \emph{Proceedings of the 41st International Conference on Machine
  Learning}, volume 235 of \emph{Proceedings of Machine Learning Research},
  pp.\  53681--53705. PMLR, 2024.
\newblock URL \url{https://proceedings.mlr.press/v235/wu24r.html}.

\bibitem[Yao et~al.(2025)Yao, Mammadov, Berner, Kerrigan, Ye, Azizzadenesheli,
  and Anandkumar]{yao2025fundps}
Jiachen Yao, Abbas Mammadov, Julius Berner, Gavin Kerrigan, Jong~Chul Ye,
  Kamyar Azizzadenesheli, and Anima Anandkumar.
\newblock Guided diffusion sampling on function spaces with applications to
  {PDEs}.
\newblock In \emph{Advances in Neural Information Processing Systems},
  volume~38, pp.\  140872--140909, 2025.
\newblock \doi{10.52202/085713-4233}.
\newblock URL
  \url{https://proceedings.neurips.cc/paper_files/paper/2025/hash/b8a28baaa25cebfa9345a8a2edf7f19d-Abstract-Conference.html}.

\bibitem[Yildirim et~al.(2009)Yildirim, Chryssostomidis, and
  Karniadakis]{yildirim2009sensors}
B.~Yildirim, C.~Chryssostomidis, and G.E. Karniadakis.
\newblock {Efficient sensor placement for ocean measurements using
  low-dimensional concepts}.
\newblock \emph{Ocean Modelling}, 27\penalty0 (3-4):\penalty0 160--173, 2009.
\newblock \doi{10.1016/j.ocemod.2009.01.001}.

\bibitem[Yin et~al.(2023)Yin, Kirchmeyer, Franceschi, Rakotomamonjy, and
  Gallinari]{yin2023dino}
Yuan Yin, Matthieu Kirchmeyer, Jean-Yves Franceschi, Alain Rakotomamonjy, and
  Patrick Gallinari.
\newblock Continuous {PDE} dynamics forecasting with implicit neural
  representations.
\newblock In \emph{International Conference on Learning Representations}, 2023.
\newblock URL \url{https://openreview.net/forum?id=B73niNjbPs}.

\end{thebibliography}
\endgroup

\clearpage
\appendix
\startcontents[sections]

\begin{center}
    \color{ceruleandark}
    \hrule
    \vspace{4mm}

    {\LARGE\scshape
        \dataset{}: Controlled Evaluation Across Observation Patterns\par
    }

    \vspace{2mm}

    {\small\bfseries
        Supplemental Material\par
    }

    \vspace{5mm}
    \hrule height 0.5pt
\end{center}

\printcontents[sections]{l}{0}{\setcounter{tocdepth}{2}}

\clearpage

\section{Benchmark Specification and Implementations}
\label{app:have}
This appendix specifies the tasks, generators, observations, losses, and baseline adaptations used to interpret the reported results.
Implementation references are paths within the public GitHub repository linked in the main text.
We distinguish the configurable software, the selected experimental protocol, and the newly verified outcomes of retained checkpoints.
An available configuration is not evidence that its experiment was executed; a complete score table is not evidence of a new prediction audit.
Appendix~\ref{app:validity} gives the artifact and scoring limitations.

\subsection{Task Contract and Information Permissions}
\label{app:have:tasks}
Let $c$ denote a complete physical condition, $u_t$ a complete state, $M$ a binary observation mask, and $G$ the geometry channel.
Here $M=1$ means observed and $G=1$ means solid or boundary geometry.
Recovery observes the solution itself, not the coefficient or source that generated it.
Rollout observes an initial state and predicts future states; for the selected temporal records, the stored condition is that initial state.
\begin{center}\footnotesize
\begin{tabular}{@{}llll@{}}\toprule
Task & Predictor input & Supervised target & Role in this paper \\ \midrule
Recovery & $M\odot u_t$, $M$, $G$ & complete $u_t$ & stationary experiments \\
Rollout & masked history, masks, $G$ & complete future states & temporal experiments \\
Forward & $M\odot c$, $M$, $G$ & complete solution & implemented extension \\
Inverse & $M\odot u_t$, $M$, $G$ & complete condition $c$ & implemented extension \\
\bottomrule\end{tabular}
\end{center}
The complete record is available to the dataset and loss code, but is not passed wholesale to the predictor.
Unobserved input entries are zero-filled; the mask distinguishes missing data from a measured physical zero.
Complete targets provide training supervision and are held separately for scoring at inference.
PINO additionally uses privileged \emph{training-loss} information: the coefficient or source for stationary residuals, and physical parameters and frame times for temporal residuals.
This does not give PINO complete inputs or future targets at inference, but makes its training information different from a purely data-driven baseline.
Third-party adapters must declare their own access; the interface does not audit arbitrary plugins.

The reported rollout is
\begin{equation}
\widehat u_1=F(M\odot u_0,M,G),\qquad
\widehat u_{h+1}=F(\widehat u_h,\mathbf 1,G),\quad h=1,2.
\end{equation}
The autoregressive wrapper in \nolinkurl{src/pdeobs/methods/neural.py} clears the initial mask after the first prediction; the one-step models supply an all-one mask when it is absent.
Predictions are fed back without detaching the computation graph.
The selected protocol fixes teacher forcing to zero: future truth enters the loss, never the recurrent input.
Three transitions between saved frames do not establish long-time forecasting stability.

\subsection{Physical Systems, Numerical Routes, and Regimes}
\label{app:have:pdes}
\paragraph{Equations and variables.}
On the unit square, the selected stationary systems are
\begin{equation}
-\nabla\cdot(a\nabla u)=f\quad\text{(Darcy)},\qquad
-\Delta u=f\quad\text{(Poisson)},\qquad
(-\Delta-k^2)u=f\quad\text{(Helmholtz)}.
\label{eq:app_stationary_pdes}
\end{equation}
The temporal systems use periodic boundaries in the reported subset:
\begin{align}
\partial_t u&=\alpha\Delta u, &&\text{Heat},\nonumber\\
\partial_t u&=0.008\,\Delta u+\rho(u-u^3), &&\text{reaction--diffusion},\nonumber\\
\partial_t u+u(\partial_{x_1}u+\partial_{x_2}u)&=\nu\Delta u, &&\text{scalar Burgers},\nonumber\\
\partial_t\omega+\boldsymbol v\cdot\nabla\omega&=\nu\Delta\omega+f, &&\text{Navier--Stokes}.
\label{eq:app_temporal_pdes}
\end{align}
Reaction--diffusion is Allen--Cahn-type and scalar, not a two-species system.
Burgers is a two-dimensional scalar equation, not a two-component velocity equation.
Navier--Stokes is stored and predicted as vorticity, with $-\Delta\psi=\omega$ and $\boldsymbol v=(\partial_{x_2}\psi,-\partial_{x_1}\psi)$.
Its fixed forcing is $0.1[\sin(2\pi(x_1+x_2))+\cos(2\pi(x_1+x_2))]$.
Darcy uses $f=\sin(2\pi x_1)\sin(2\pi x_2)+0.2\sin(4\pi x_1+0.3)\sin(2\pi x_2)$, centered to zero discrete mean.
The family modules under \nolinkurl{src/pdeobs/pdes/} define these equations and parameters.
\begin{table}[tbp]\centering\small
\caption{Evaluated task definitions. Each family uses 2,000 scalar fields or trajectories on a $128\times128$ grid. Temporal targets are frames 1--3 of 15 equally spaced frames, $t_h=hT/14$. All inputs are noise-free partial observations.}
\label{tab:tasks}
\begin{tabularx}{\linewidth}{@{}lYlc@{}}\toprule
Family & Prediction & Boundary & $T$\\ \midrule
Darcy & Solution-field recovery & Dirichlet & --\\
Poisson & Solution-field recovery & Dirichlet & --\\
Helmholtz & Solution-field recovery & Dirichlet & --\\
Heat & Initial field to three future states & Periodic & .25\\
Reaction--diffusion & Initial field to three future states & Periodic & .8\\
Burgers & Initial field to three future states & Periodic & .35\\
Navier--Stokes & Initial vorticity to three future states & Periodic & .4\\
\bottomrule\end{tabularx}\end{table}

\begin{table}[tbp]\centering\small
\caption{Parameters in low/medium/high order. Regime names are family-specific, not a common difficulty scale.}
\label{tab:app_regimes}
\begin{tabular}{@{}lll@{}}\toprule
Family & Parameter & Values \\ \midrule
Darcy & coefficient contrast $\chi$ & 2, 8, 32 \\
Poisson & source amplitude & 0.5, 1, 2 \\
Helmholtz & wavenumber $k$; damping & 0.5, 1, 2; zero \\
Heat & diffusivity $\alpha$ & 0.005, 0.02, 0.08 \\
Reaction--diffusion & rate $\rho$; diffusivity & 0.5, 1.5, 4; 0.008 \\
Burgers & viscosity $\nu$ & 0.04, 0.015, 0.004 \\
Navier--Stokes & viscosity; initial-vorticity scale & 0.02, 0.008, 0.0025; 2, 3, 4 \\
\bottomrule\end{tabular}\end{table}

\paragraph{Condition generation.}
The selected construction, \code{smooth\_grf} in \nolinkurl{src/pdeobs/settings.py}, filters standard-normal grid noise in Fourier space with
\begin{equation}
A(q_1,q_2)=\exp[-2(\pi\ell)^2HW(q_1^2+q_2^2)],\qquad \ell=0.22,
\label{eq:app_grf}
\end{equation}
where $q$ is frequency in cycles per grid cell, then centers and standardizes the field.
Darcy range-normalizes the latent field to $s\in[-1,1]$ and sets $a=\exp[\tfrac12\log(\chi)s]$.
Poisson multiplies it by the source amplitude; Helmholtz uses it directly as the source.
Heat and Burgers use it as the initial state, reaction--diffusion multiplies it by 0.8, and Navier--Stokes applies the regime-specific vorticity scale and removes the mean.
These are generation transforms, not normalizations fitted to test data.

\paragraph{Grid, integration, and stored times.}
The selected records use $128\times128$ cell-center coordinates $x_j=(j+\tfrac12)/128$.
For stationary Dirichlet cases, the outer stored rows and columns are zero and the system is solved on the interior.
Geometry marks this ring; the reported subset has no interior obstacle.
Numerical coordinates must not be confused with the coordinate features constructed internally by a neural adapter.

The kernels in \nolinkurl{src/pdeobs/pdes/numerics.py} assemble a sparse second-order flux operator with arithmetic face coefficients.
Darcy and Poisson use diagonally preconditioned conjugate gradients; Helmholtz uses MINRES for the negative reaction term.
The selected small Dirichlet wavenumbers do not establish a high-frequency, indefinite Helmholtz benchmark merely because MINRES is used.
The solvers request relative tolerance $10^{-9}$, record the attained residual, and raise on iterative failure.
Periodic Heat uses Fourier diffusion; reaction--diffusion uses Strang splitting with an analytic reaction update.
The selected smooth periodic Burgers cases use dealiased spectral advection with SSPRK2 stages and diffusion updates.
Periodic Navier--Stokes uses dealiased vorticity advection and Crank--Nicolson diffusion, with internal increments bounded by $10^{-4}$.
Numerical arrays are computed in float64 and the selected storage dtype is float32.

\nolinkurl{configs/dataset/numerics_full_t15.yaml} specifies dense temporal outputs for residual checks and selects 15 exact frames by an integer stride, not interpolation.
With final stored time $T$ in Table~\ref{tab:tasks}, the saved times are $t_h=hT/14$.
The benchmark observes saved frame 0 and predicts saved frames 1--3; solver-frame indices and physical times are retained in metadata.
It does not evaluate a forecast over the entire stored trajectory.

\paragraph{What quality checks establish.}
Generation checks array contracts, finiteness, geometry, boundary and initial-condition defects, transition replay, and family-specific discrete PDE residuals.
Default tolerances include finite fraction 1, geometry error $10^{-6}$, initial-condition defect $10^{-6}$, boundary defect $10^{-4}$, and first-transition replay defect $5\times10^{-6}$.
The selected numerical configuration inherits a \code{strict} profile with normalized PDE-residual threshold 0.05 and divergence threshold $10^{-6}$ where applicable.
These are discrete consistency checks, not bounds on error against an independent continuum solution.

The profiles are not interchangeable.
\code{report} records threshold outcomes but still rejects hard array, domain, or residual-contract failures; \code{strict} also rejects failing quality status.
The current \code{publication} path requires independently verified solver/calibration evidence and keeps publication readiness false because no trusted verification mechanism is configured.
Neither the profile name nor a checksum-shaped metadata field certifies publication quality.

\paragraph{Recorded production quality.}
The original generation summary, whose digest matches the dataset pin of all 441 archived settings, records 560,000 passing records, with no missing or invalid quality entries and 80,000 records per family.
Filtering its disjoint calibration groups to the evaluated boundary and \code{smooth\_grf} construction gives exactly 2,000 records per family (regimes 667/667/666).
Table~\ref{tab:production-qc} reports their recorded normalized discrete PDE residuals; every maximum is below 0.05.
The public artifact includes the full-precision extraction, source digest, operator identifiers and aggregation procedure in \nolinkurl{results/numerical_quality/}.
These are verified generation-time records, not a new solution of the full corpus or independent reference-solution error bounds. No grid-convergence study is inferred from a small residual.

\begin{table}[!ht]
\centering
\caption{Recorded production consistency for the evaluated slice (2,000 records per family). Residuals use the family-specific discrete operator and normalization, not a common continuum-error scale.}
\label{tab:production-qc}
\small
\begin{tabular}{lrr}
\toprule
PDE & Mean normalized residual & Maximum normalized residual \\
\midrule
Darcy & $1.14\times10^{-5}$ & $1.57\times10^{-5}$ \\
Poisson & $1.69\times10^{-5}$ & $2.84\times10^{-5}$ \\
Helmholtz & $1.80\times10^{-5}$ & $3.52\times10^{-5}$ \\
Heat & $1.32\times10^{-8}$ & $1.51\times10^{-8}$ \\
Reaction--diffusion & $3.39\times10^{-5}$ & $1.55\times10^{-4}$ \\
Burgers & $1.26\times10^{-2}$ & $3.85\times10^{-2}$ \\
Navier--Stokes & $5.08\times10^{-5}$ & $1.37\times10^{-4}$ \\
\bottomrule
\end{tabular}
\end{table}

\subsubsection{Discrete Updates and Production Residuals}
\label{app:numerical-details}
The parameters below describe the seven routes used in the evaluated slice, not every optional solver in the package. They are traced to the family generators and \nolinkurl{src/pdeobs/pdes/numerics.py}, with the original production calibration contexts retained in \nolinkurl{results/numerical_quality/}. The residual summary predates model training. No additional numerical solutions are generated for this analysis.

\paragraph{Elliptic systems.}
Let $h=1/128$. For Darcy, a face coefficient is the arithmetic average of its neighboring cell values, for example $a_{i+1/2,j}=(a_{i+1,j}+a_{i,j})/2$. The discrete operator \addedcitep{leveque2007finite} is
\begin{equation}\begin{aligned}
(A_a u)_{i,j}={}&-\frac{a_{i+1/2,j}(u_{i+1,j}-u_{i,j})-a_{i-1/2,j}(u_{i,j}-u_{i-1,j})}{h^2}\\
&-\frac{a_{i,j+1/2}(u_{i,j+1}-u_{i,j})-a_{i,j-1/2}(u_{i,j}-u_{i,j-1})}{h^2}.
\end{aligned}\end{equation}
The unknowns are the interior samples; prescribed zero outer-layer values enter the assembled boundary-value problem. Darcy solves $A_a u=f$, Poisson solves $A_1u=f$, and Helmholtz solves $(A_1-k^2I)u=f$. The diagonal preconditioner \addedcitep{saad2003iterative} divides by $\max(|A_{ii}|,10^{-12})$. The requested relative tolerance is $10^{-9}$; iteration caps are 6,000, 4,000, and 6,000, respectively. Conjugate gradients \addedcitep{hestenes1952cg} are used for Darcy/Poisson and MINRES \addedcitep{paige1975minres} for Helmholtz, whose small selected wavenumbers remain coercive on this Dirichlet domain. The stored solution is float32, whereas assembly and iteration use float64; the reported post-storage residual is therefore distinct from the solver's stopping tolerance.

\paragraph{Diffusion and reaction--diffusion.}
For periodic diffusion, a substep of length $\delta$ multiplies each Fourier coefficient by $\exp(-\alpha\delta|\xi|^2)$, with $\xi=2\pi\,\mathrm{fftfreq}(128,h)$. Heat uses this exact semidiscrete propagation \addedcitep{trefethen2000spectral,boyd2001spectral}. Reaction--diffusion alternates exact half reaction, exact Fourier diffusion, and exact half reaction \addedcitep{strang1968splitting}. The reaction flow is
\begin{equation}
\mathcal R_s(u)=\frac{u}{\sqrt{u^2+(1-u^2)e^{-2\rho s}}},\qquad
u^{n+1}=\mathcal R_{\delta/2}\!\left(e^{\kappa\delta\Delta_h}\mathcal R_{\delta/2}(u^n)\right).
\end{equation}
The code floors the square-root argument at float64 machine epsilon. Each dense output interval $\Delta_q$ is divided into $m=\max(1,\lceil\rho\Delta_q/0.18\rceil)$ substeps with $\delta=\Delta_q/m$. The selected regimes and output intervals give $m=1$; no state clipping is applied.

\paragraph{Burgers.}
The selected smooth periodic route differentiates spectrally and filters the nonlinear product to integer Fourier modes satisfying $|k_x|,|k_y|\le\lfloor128/3\rfloor$ \addedcitep{patterson1971spectral}. Writing the negative dealiased advection as $N(u)$, a substep forms \addedcitep{gottlieb1998tvd} $u^*=u^n+\delta N(u^n)$ and $\widetilde u=u^n+\frac{\delta}{2}[N(u^n)+N(u^*)]$, followed by $u^{n+1}=e^{\nu\delta\Delta_h}\widetilde u$. The step is
\begin{equation}
\delta=\min\!\left(\hbox{remaining output interval},\frac{0.35}{\max(\|u^n\|_\infty,10^{-12})(h^{-1}+h^{-1})}\right).
\end{equation}
The controller raises on nonfinite states or more than 10,000 substeps in one output interval. It does not clip the state. The package's Rusanov alternative is used for other constructions or boundaries, not for the smooth periodic slice plotted here.

\paragraph{Periodic vorticity.}
With $\widehat\psi=\widehat\omega/|\xi|^2$ and $\widehat\psi(0)=0$, velocity is obtained by Fourier differentiation. Let $\widehat N$ be the two-thirds-filtered transform of $\boldsymbol v\cdot\nabla\omega$. The update is
\begin{equation}
\widehat\omega^{n+1}=\frac{(1-\nu\delta|\xi|^2/2)\widehat\omega^n-\delta\widehat N^n+\delta\widehat f}{1+\nu\delta|\xi|^2/2}.
\end{equation}
An output interval uses $m=\max(1,\lceil\Delta_q/10^{-4}\rceil)$ substeps, $\delta=\Delta_q/m$; mean vorticity is removed at the interval boundary. Nonlinear advection and forcing are explicit, while diffusion is Crank--Nicolson \addedcitep{crank1947practical}. The forcing and regime-dependent initial amplitudes are given above; this is the vorticity route, not a velocity--pressure or obstacle-flow simulation.

\begin{table}[!ht]\centering\small
\caption{Time grids in the evaluated periodic slice. Dense quality frames include both endpoints; stored data select every indicated stride. Internal substeps are separate from both grids.}
\label{tab:numerical-time-grids}
\begin{tabular}{@{}lrrrrl@{}}\toprule
PDE & Final $T$ & Dense frames & Stride & Stored frames & Internal rule \\\midrule
Heat & 0.25 & 155 & 11 & 15 & Fourier propagation \\
Reaction--diffusion & 0.8 & 267 & 19 & 15 & $\rho\delta\le0.18$ \\
Burgers & 0.35 & 267 & 19 & 15 & CFL bound above \\
Navier--Stokes & 0.4 & 71 & 5 & 15 & $\delta\le10^{-4}$ \\\bottomrule
\end{tabular}\end{table}

\paragraph{What the plotted PDE loss measures.}
For each record, \nolinkurl{src/pdeobs/quality.py} computes
\begin{equation}
q=\frac{\mathrm{RMS}_{\Omega_a}(r)}{\sum_j\mathrm{RMS}_{\Omega_a}(a_j)},\qquad r=\sum_j s_j a_j,
\label{eq:production-residual}
\end{equation}
where $s_j\in\{-1,1\}$ are the equation signs and $\Omega_a$ contains valid stencils. If the denominator is at most $10^{-12}$, the code returns zero when the numerator is also at most $10^{-12}$, and otherwise divides by $10^{-12}$. For the selected stationary wall geometry, the outer ring and its one-cell stencil halo are excluded from quality scoring; periodic scoring uses all cells. This domain differs from the full-grid neural prediction metric.

The stationary components are the complete elliptic operator and its source. For the nonlinear temporal families, the quality checker uses a finite difference in time and midpoint states for diffusion, reaction or dealiased advection, with the original physical coefficients and forcing. The principal reported residual aggregates post-initial \emph{dense quality} intervals; the first dense transition is checked separately by replaying the same numerical update, with threshold $5\times10^{-6}$. For Heat the principal operator is instead the spectral semigroup defect $u^{n+1}-e^{\alpha\Delta_q\Delta_h}u^n$, normalized by the two state RMS values; the finite-difference strong-form defect is auxiliary. This explains why its scale should not be interpreted as a cross-family solution-accuracy ranking.

\begin{figure}[!ht]
\centering
\includegraphics[width=\linewidth]{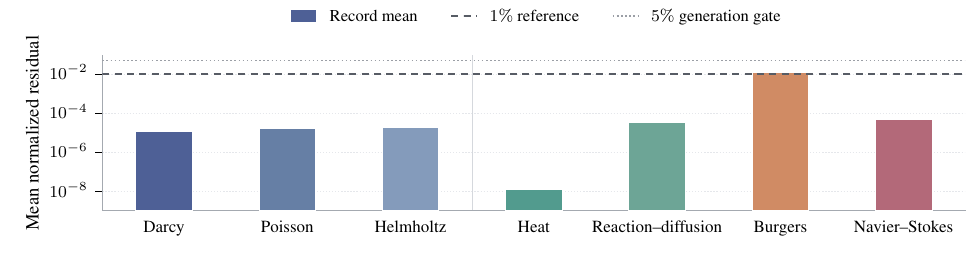}
\caption{\textbf{Mean recorded PDE residuals.} Each bar is the arithmetic mean over 2,000 selected records, shown on a logarithmic axis. Six family means are below $1\%$; Burgers is $1.26\%$. The dashed line marks the $1\%$ reference and the dotted line the original $5\%$ generation gate. Residual operators and normalizations are family-specific, with a semigroup defect for Heat; these values measure production discrete consistency, not independent solution error.}
\label{fig:production-residual}
\end{figure}

The bars in Figure~\ref{fig:production-residual} show the arithmetic mean of these per-record values over all 2,000 selected records per family, including both training and test identities. Six family means are below $1\%$; Burgers has mean $1.26\%$. The $1\%$ reference is not the configured $5\%$ generation gate, and no value is clipped to it. Extrema remain reported in Table~\ref{tab:production-qc}. These are neither training losses nor independent discretization-error estimates. Calibration groups are disjoint: counts and first moments are summed with their recorded sample weights, second moments recover the pooled variance, and extrema are taken across groups. The separate initial replay verifies implementation consistency, not an independent reference solution, and is not a replay over the longer stored $t_0\!\to t_1$ interval. Analytic/reference error and grid-convergence evidence for all seven production routes remains outside this result cut.

\subsection{Data Design, Splits, and Storage}
\label{app:have:data}
\paragraph{Design versus evaluated slice.}
The full software design has seven families, four boundary protocols, and ten condition constructions, with 2,000 records per combination.
The regimes \emph{partition} those records as 667/667/666; they do not multiply the count.
\revnew{The design therefore has $7\times4\times10\times2000=560{,}000$ records.}
The paper uses seven combinations: Dirichlet with \code{smooth\_grf} for stationary families and periodic with \code{smooth\_grf} for temporal families, totaling 14,000 records.
Other boundaries and constructions are implemented capabilities, not additional reported experiments.
Release documentation lists 3,360 HDF5 shards for the full design and 84 for the selected slice; these are inventory declarations, not a fresh full-corpus verification here.

\paragraph{Exact split.}
\nolinkurl{src/pdeobs/one_setting.py::stable_split} requires 2,000 records and all three regimes.
Within each regime it ranks identities by SHA-256 of \nolinkurl{seed|pde|boundary|setting|sample_id}, using seed 20260804.
Largest-remainder allocation assigns 67/67/66 test identities and leaves 600 training identities per regime: 1,800 training, no validation, and 200 test records.
The same identity lists are used across methods and observation views.
A loader label such as \code{iid} does not reconstruct this partition; the identity lists and split receipt must be retained.
Test identities must not be used for early stopping, hyperparameter selection, or checkpoint selection.

\paragraph{Canonical storage.}
Each HDF5 shard stores conditions of shape $[N,H,W,C_c]$, trajectories of shape $[N,T,H,W,C_s]$, and geometry of shape $[N,H,W,1]$.
The array keys are \code{condition}, \code{trajectory}, and \code{geometry}; each identity also has a UTF-8 JSON metadata record.
Here $C_c=C_s=1$, with $T=1$ for stationary cases and $T=15$ for temporal cases.
Metadata includes family, boundary, construction, regime, seeds, identity, frame indices, and physical times.
\nolinkurl{src/pdeobs/storage.py} writes compressed arrays and atomically finalizes temporary shards, with checksum, manifest, metadata, and quality sidecars.
Default compression is gzip level 4.
Identity derives from the generation specification, not a row's file position.
Regenerated records are not automatically the historical evaluated records: their manifests and identity receipts must agree.

\subsection{Observation Construction}
\label{app:have:obs}
Views are frozen in \nolinkurl{configs/paper/observations.yaml} and implemented by \nolinkurl{src/pdeobs/masks.py}.
For grid $\Omega_h$, the realized count is $K_{i,w}=\sum_{x\in\Omega_h}M_{i,w}(x)$ and density is $r_{i,w}=K_{i,w}/|\Omega_h|$.
Integer counts, rather than nominal percentages, define the equal-count groups.
\begin{table}[tbp]\centering\small
\caption{Exact observation counts out of 16,384 locations. Equal-count comparisons use R50/H/V/CL or BL/LI/BD, not all nominal 50\% masks together. A perimeter band on a periodic grid is not a physical wall.}
\label{tab:masks}
\begin{tabularx}{\linewidth}{@{}llrY@{}}\toprule
Code & Pattern & Count $K$ & Construction\\ \midrule
R50 & Random & 8192 & Uniform distinct locations\\
R65 & Random & 10650 & Uniform distinct locations\\
R80 & Random & 13107 & Uniform distinct locations\\
BL & Missing block & 8284 & Outside a missing $90\times90$ square\\
LI & Crossed lines & 8284 & Union of 38 rows and 38 columns\\
H & Horizontal lines & 8192 & 64 distinct rows\\
V & Vertical lines & 8192 & 64 distinct columns\\
BD & Perimeter band & 8284 & A 19-cell band\\
CL & Clustered points & 8192 & Four Gaussian clusters; distinct locations\\
\bottomrule\end{tabularx}\end{table}

R50, R65, and R80 sample distinct cells uniformly without replacement with the requested counts.
These are separate calls under the deterministic seed policy, not a constructed nested sequence; neither nesting nor statistical independence is assumed.
H and V choose 64 distinct rows or columns.
LI observes 38 rows and 38 columns, giving $38\cdot128+38\cdot128-38^2=8{,}284$ cells.
BL observes outside a non-wrapping $90\times90$ missing square whose top-left corner is randomly drawn from valid positions.
BD observes the deterministic width-19 band, also giving 8,284 cells.
CL samples 8,192 distinct cells using four Gaussian clusters, random centers in array coordinates, and standard deviation $0.08\cdot128$ cells.
Blocks and cluster distances do not wrap across periodic edges.

Factories operate on array locations and do not intersect observations with geometry.
A stationary Dirichlet record has 508 prescribed-zero outer-ring cells; selecting one counts as an observation.
BD includes that ring and therefore observes 7,776 interior cells.
Other views contain different boundary-cell counts, so equal totals need not imply equal numbers of unknown interior values.
The metric includes boundary entries; this fixed convention does not remove that information-content confound.
On periodic data, BD is a perimeter sampling layout, not a new physical wall.

\code{BenchmarkDataset} derives a stable mask seed from the campaign seed, \code{mask}, sample identity, and canonical factory name.
There is no epoch or method index, so a record's mask stays fixed across epochs and methods under a fixed view.
The \code{paper-row} path separates this seed from the model training seed. \revcode{The runner configuration exposes the same separation: \code{data.mask\_seed} pins the sensors independently of the campaign seed and \code{training.seed} seeds only initialization and batch order.}
Specifications carry namespace, parameters, version, and a content-derived identifier.
Changing a general API seed is not equivalent to changing only initialization.
The selected views are noiseless, stationary point observations; noise, moving sensors, and active sampling are not reported controls.

\subsection{Metric Definitions and Scoring Semantics}
\label{app:have:metrics}
For finite predictions and targets, the intended per-identity error and block mean are
\begin{equation}
e_i^{v,w}=\frac{\|\widehat y_i^{v,w}-y_i\|_2}{\max(\|y_i\|_2,10^{-12})},\qquad
E_{v,w}=\frac1{N_{\mathrm{test}}}\sum_{i\in\mathcal I_{\mathrm{test}}}e_i^{v,w},\qquad N_{\mathrm{test}}=200.
\label{eq:metric}
\end{equation}
The norm includes every stored spatial location, including observed cells, and jointly includes all three target frames for rollout.
Ratios are not percentages unless labelled.
For nine views $\mathcal V$,
\begin{equation}
D=\frac1{9}\sum_v E_{v,v},\qquad C=\frac1{72}\sum_{v\ne w}E_{v,w},
\label{eq:matched_transfer}
\end{equation}
and the destination-referenced contrast is
\begin{equation}
\Delta_{v\to w}=E_{v,w}-E_{w,w}.
\label{eq:destination}
\end{equation}
Both terms use the same test view and identities.
Over all transfers, mean $\Delta=C-D$; the contrast is useful directionally or within restricted groups, not as an independent third whole-matrix summary.
Complete-matrix summaries require every cell.

\paragraph{Separate diagnostics.}
Horizon-wise errors are
\begin{equation}
e_{i,h}^{v,w}=\frac{\|\widehat u_{i,h}^{v,w}-u_{i,h}\|_2}{\max(\|u_{i,h}\|_2,10^{-12})},\quad h=1,2,3.
\label{eq:app_horizon}
\end{equation}
Their mean is generally not the joint error.
For stationary recovery, set $R_i=\widehat u_i-u_i$ and $q_i=\max(\|u_i\|_2,10^{-12})$.
Then
\begin{equation}
e_{\mathrm{obs},i}=\frac{\|M_i\odot R_i\|_2}{q_i},\quad
e_{\mathrm{hid},i}=\frac{\|(1-M_i)\odot R_i\|_2}{q_i},\quad
e_i^2=e_{\mathrm{obs},i}^2+e_{\mathrm{hid},i}^2.
\label{eq:app_regional}
\end{equation}
This identity holds per record, not after separately averaging the errors.
Hidden-only normalization by $\|(1-M_i)\odot u_i\|_2$ is a different diagnostic.
Replacing observed predictions with supplied noiseless values leaves hidden predictions unchanged and cannot increase full-field recovery error.
\revnew{The stationary projection results are reported separately in \appref{app:controls}; projection cannot inject future truth into a forecast.}

\paragraph{Strict scorer versus archive.}
The \code{pdeobs-strict-v1} scorer uses an explicit contract fixing expected identities, shape, task, and target-frame indices.
It aligns by identity and rejects missing, extra, duplicate, nonfinite, or mismatched entries before float64 norms and ordinary arithmetic means.
Recovery also checks raw observations and masks against targets before projection.
These checks do not authenticate provenance, establish numerical fidelity, or prove generalization.
The implementation is in \nolinkurl{src/pdeobs/strict_score.py}.
The original campaign metrics remain a separate legacy archive. The current manuscript instead uses new predictions and strict scores for every retained checkpoint, with the terminal audit described in Appendix~\ref{app:validity}; original arrays are not retrospectively recovered.

\subsection{Baseline Adaptations and Training Recipes}
\label{app:have:models}
\paragraph{Reference architectures.}
The baselines are PDE-OBS adaptations, not interchangeable copies of upstream pipelines.
Their reference registry is \nolinkurl{configs/method/all_pde_source_faithful_registry.yaml}; exposed schemas and presets are in \nolinkurl{src/pdeobs/api/specs.py}.
U-FNO-2D has three Fourier and three U-Fourier blocks, width 36, 12 modes, and padding 8.
FNO uses spectral convolution with a parallel local path.
The CNO-inspired CNN uses fixed binomial filtering in an encoder--decoder; its scores are not those of an exact upstream CNO implementation.
DeepONet uses an ordered spatial CNN branch and a branch--trunk product.
GNOT uses linear attention and geometry-gated experts without upstream unit normalization.
Transolver uses physics-slice attention and a larger temporal preset.
PINO combines an FNO backbone with a trainer-side residual.
Registry revisions, licenses, and adaptation notes provide attribution, not a new independent equivalence or license audit.
\begin{table}[tbp]\centering\small
\caption{Frozen reference model sizes and nominal per-device batch sizes. These describe the reference registry, not a proof of identical effective batches or capacities in every execution attempt. Complex weights are counted as real scalars. Attempt-specific configurations remain necessary for reproduction.}
\label{tab:reference-models}
\begin{tabularx}{\linewidth}{@{}lYrr@{}}\toprule
Adaptation & Reference configuration & Parameters & Batch\\ \midrule
U-FNO-2D & Width 36; Fourier and U blocks & 5,049,545 & 4\\
FNO & Width 32; 12 modes; 4 layers & 2,366,145 & 8\\
CNO-inspired CNN & Width 32; filtered convolution & 103,617 & 8\\
DeepONet & Spatial CNN branch; latent 128 & 221,121 & 4\\
GNOT & Width 128; 3 layers; 2 experts & 2,099,975 & 4\\
Transolver (static) & Width 128; 8 layers; 64 slices & 2,811,457 & 4\\
Transolver (temporal) & Width 256; 8 layers; 32 slices & 11,196,737 & 2\\
PINO & FNO width 64; 20 modes; 5 layers & 32,798,273 & 4\\
\bottomrule\end{tabularx}\end{table}

\paragraph{Original recipe and later attempts.}
The original reference budget is 500 epochs over 1,800 training records, with no validation partition, plateau stopping, or mixed precision, and with the final checkpoint retained.
Its initial or peak learning rate is $10^{-3}$.
For minibatch $\mathcal B$,
\begin{equation}
\mathcal L_{\mathrm{data}}=\frac1{|\mathcal B|}\sum_{i\in\mathcal B}
\frac{\|\widehat y_i-y_i\|_2}{\max(\|y_i\|_2,10^{-12})},
\label{eq:app_loss}
\end{equation}
computed on float32 predictions and targets after flattening non-batch dimensions.
Table~\ref{tab:app_optimizers} gives reference defaults, not a universal description of the archive.
Actual configurations take precedence for later learning-rate, clipping, continuation, and stopping amendments.
Actual epochs, cohorts, stop reasons, and checkpoint identities remain in the results.
The public \code{paper-row} launcher executes \code{original500} and \code{demo}; historical fixed-200, patience, and recovery labels are not executable replays of those attempts in that launcher.
\begin{table}[tbp]\centering\footnotesize
\caption{Original optimization defaults. Step schedules halve the rate every 100 epochs; one-cycle schedules advance per batch with cosine annealing. Parentheses denote temporal Transolver settings.}
\label{tab:app_optimizers}
\setlength{\tabcolsep}{4pt}\begin{tabular}{@{}llrrll@{}}\toprule
Method & Optimizer & Weight decay & Batch & Schedule & Grad.\ clip \\ \midrule
U-FNO-2D & Adam & $10^{-4}$ & 4 & step & 1.0 \\
FNO & Adam & $10^{-4}$ & 8 & step & 1.0 \\
CNO-inspired CNN & AdamW & $10^{-6}$ & 8 & step & 1.0 \\
DeepONet & Adam & 0 & 4 & step & 1.0 \\
GNOT & AdamW & $5\times10^{-5}$ & 4 & one-cycle, 20\% warm-up & 1000 \\
Transolver & AdamW & $10^{-5}$ & 4 (2) & one-cycle, 30\% warm-up & 0.1 (none) \\
PINO & Adam & 0 & 4 & step & 1.0 \\
\bottomrule\end{tabular}\end{table}

\paragraph{PINO residual.}
The reference objective is $5\mathcal L_{\mathrm{data}}+\mathcal L_{\mathrm{phys}}$.
Complete targets enter the data term; the residual uses predictions and declared physical context.
For active cells $\mathcal A$,
\begin{equation}
\mathcal L_{\mathrm{phys}}=\frac1{|\mathcal A|}\sum_{p\in\mathcal A}(r_p/\sigma)^2,\qquad
\sigma=\max\!\left(\sum_j\operatorname{RMS}_{\mathcal A}(\tau_j),10^{-12}\right),
\label{eq:app_pino_residual}
\end{equation}
where a signed sum of PDE terms $\tau_j$ forms $r$; contributions are averaged over records and temporal pairs.
Stationary residuals use finite-difference/flux stencils, stored coefficients or sources, and a detached normalization scale.
The active mask excludes solid cells and their stencil neighbors, including the prescribed wall ring in the selected Dirichlet data.
Temporal residuals use spectral spatial derivatives of the midpoint between two predicted states and their finite-difference time derivative.
Only pairs $(\widehat u_1,\widehat u_2)$ and $(\widehat u_2,\widehat u_3)$ contribute, using recorded physical times; their normalization remains differentiable.
\nolinkurl{src/pdeobs/pino.py} implements distinct \nolinkurl{pino_static_fd_v1} and \nolinkurl{pino_rollout_spectral_v1} contracts, the latter periodic-only.
Mixed precision is rejected while a physics residual is active.

\paragraph{Unreported comparators.}
The software also contains interpolation, gappy POD, zero-fill, observed-mean, and persistence methods.
The method named \code{mean} fills using an example's observed mean, not a training-population mean.
These methods and forward/inverse interfaces contribute no result to the 441-setting learned grid.

\subsection{Software and Artifact Access}
\label{app:have:software}
The inspected package is version \revcode{0.2.1 (0.2.0 plus the observation-spec interface of \appref{app:provide:obs})} with review overlays and MIT-licensed code.
It requires Python 3.10 or newer, NumPy, SciPy $\geq1.12$, h5py, and PyYAML; learned training adds PyTorch through the \code{train} extra.
The package includes generators, adapters, configurations, tests, demonstrations, new per-record and aggregate scores, and the separate archived metrics, not the full tensor data or weight corpus.
Data licensing is separate.
Release documents point to separate data and model deposits, whose unauthenticated accessibility and inventory require independent checks; a README badge is not verification.
\revnew{The dated 26 September 2026 release report records unauthenticated access and file-digest agreement for the \href{https://huggingface.co/datasets/PDE-OBS/pdeobs-data}{data} and \href{https://huggingface.co/PDE-OBS/pdeobs-models}{model} deposits; inventory scope and reproduction limits are stated in \appref{app:validity}.}
\code{pdeobs download} requires an explicit checksum manifest and has no default endpoint.
Checksums verify bytes against a manifest, not the identity or anonymity of its publisher.
Appendices~\ref{app:provide} and~\ref{app:validity} distinguish runnable interfaces, aggregate arithmetic, inference, and training reproducibility.

\subsection{Baseline Adaptation Notes}
\label{app:have:adaptations}
\begin{movedblock}{Section~\ref{sec:baselines}}
\paragraph{U-FNO.}
U-FNO \citep{wen2022ufno} combines Fourier blocks with U-Net-style local branches \addedcitep{ronneberger2015unet}. We adapt its six-block composition from three-dimensional space--time to two spatial dimensions, adding mask, coordinate and geometry channels. It is explicitly a two-dimensional adaptation, not a replication of the original model.

\paragraph{FNO.}
FNO \citep{li2020fno} combines learned truncated Fourier multipliers with a pointwise path. We use a compact mask-channel variant; PINO below uses the same backbone family with a different width.

\paragraph{CNO-inspired CNN.}
Inspired by CNO \citep{raonic2023cno}, our encoder--bottleneck--decoder applies a fixed binomial low-pass filter before each learned convolution. It does not reproduce the original continuous--discrete construction and is therefore labelled an inspired variant.

\paragraph{DeepONet.}
DeepONet \citep{lu2021deeponet} combines an input branch and coordinate trunk by an inner product. A small convolutional branch on the masked field retains sensor ordering, replacing the original fixed sensor vector; a set branch remains available for legacy checkpoints.

\paragraph{GNOT.}
GNOT \citep{hao2023gnot} uses normalized linear cross-attention and geometry-gated mixture-of-expert layers. The structured-grid adaptation retains these mechanisms but omits upstream unit normalization, optimizing the physical relative-$L^2$ objective.

\paragraph{Transolver.}
Transolver \citep{wu2024transolver} attends over learned physics-aware slices instead of grid points. We use structured-grid slice attention with the published Darcy and Navier--Stokes dimensions; it is the only baseline with a separate temporal structure.

\paragraph{PINO.}
PINO \citep{li2021pino} augments FNO's data loss with stationary finite-difference or temporal spectral residuals. Coefficients, sources and physical parameters enter the training residual, not inference. Complete targets enter only the data term. Its wider backbone prevents interpreting the FNO--PINO contrast as an isolated effect of physics supervision.
\end{movedblock}

\section{Using and Extending the Released Code}
\label{app:provide}
This appendix separates a small software example from a formal benchmark run.
The public entry points are \code{from pdeobs import api} and the \code{pdeobs} command.
\nolinkurl{configs/paper/protocol.yaml} describes the study; it is not itself a pipeline configuration or job launcher.
Likewise, \code{preset="paper"} selects reference defaults, not a certified reproduction of a historical checkpoint.

\subsection{Records and Physical Inputs}
\label{app:provide:data}
\paragraph{Local generation.}
\nolinkurl{api.create_dataset} takes explicit \code{pde} and \code{out} arguments.
Options include boundary, construction, regime, count, resolution, seed, temporal frame counts, shard size, dtype, and numerical solver options.
Static records have one frame; temporal calls must specify at least two.
This convenience function uses the \code{report} quality profile, not the full numerical configuration's \code{strict} profile.
Advanced generation resolves quality and dense/stored frame choices from \nolinkurl{configs/dataset/numerics_full_t15.yaml} and its included configuration.
A short count or failed solve is an error, not permission to substitute new identities.

The ten constructions comprise three Gaussian-field smoothness levels; low-frequency and multi-frequency Fourier fields; Gaussian blobs; piecewise blocks; threshold level sets; dipole vortex pairs; and front/ring/shock fields.
Their exact registry keys are defined in \nolinkurl{src/pdeobs/settings.py}.
Gaussian-field correlation parameters are 0.22, 0.09, and 0.025.
Only \code{smooth\_grf} occurs in the reported grid.
The boundary labels \code{periodic}, \code{dirichlet}, \code{neumann}, and \code{robin\_obstacle} have family-dependent numerical meanings; they are not interchangeable solver settings.

\paragraph{Loading and verification.}
With \code{verify=True}, \code{api.load\_dataset} verifies local shards against completion sidecars.
With \code{source="manifest"}, it uses an identity manifest and verifies its shard hashes.
The separate \code{pdeobs download --manifest <manifest>} route consumes a release download manifest, a different format from an identity manifest.
The inspected \code{api.load\_dataset(..., source="download")} route raises rather than downloading.
Loading neither trains a model nor regenerates missing records.

\paragraph{External conditions.}
\nolinkurl{api.create_dataset_from_arrays} accepts a family, boundary, output path, complete physical arrays, and physical parameters.
It solves these specified problems with the existing kernels and writes canonical records.
It requires a complete problem specification; it does not infer missing coefficients.
These outputs are new data, not additional historical evidence.

\subsection{Observation Namespaces}
\label{app:provide:obs}
\nolinkurl{api.make_observation} resolves four namespaces.
\code{paper:R50} and its eight companions replay frozen definitions and reject overrides.
The paper namespace requires a $128\times128$ grid in the training interface.
The general namespace provides parameterized factories; custom delegates to a registered factory; stored reuses a supplied inference mask.
Names, namespaces, versions, and resolved parameters contribute to the observation identifier.
A general mask matching a paper mask numerically does not acquire paper provenance.
\begin{table}[tbp]\centering\footnotesize
\caption{General observation controls. Record realized locations and counts; a requested ratio need not equal realized density. Alternatives separated by \emph{or} are mutually exclusive.}
\label{tab:app_general_obs}
\setlength{\tabcolsep}{4pt}
\begin{tabularx}{\linewidth}{@{}l>{\raggedright\arraybackslash}X@{}}\toprule
Factory & Controls \\ \midrule
Random & \code{ratio} or \code{count}; distinct cells \\
Grid & \code{ratio} or \code{spacing}; optional \code{random\_offset} \\
Block & \code{missing\_fraction} or \code{block\_shape}; observe outside the block \\
Line / horizontal / vertical & \code{ratio} or \code{num\_lines}; orientation for general lines \\
Boundary & \code{width}; optional deterministic \code{count} within the band \\
Clustered & \code{ratio} or \code{count}; \code{clusters} and \code{spread} \\
Full & no pattern parameters; diagnostic complete observation \\
\revcode{Mixture} & \revcode{components joined by \code{+}, optional integer weights (\code{x2}); one component per record by stratum position} \\
\bottomrule\end{tabularx}\end{table}
\code{api.realized\_count(spec, (H,W))} evaluates the discrete count.
\revcode{A mask spec is a protocol name with the observed fraction in percent (\code{uniform 20}, \code{block 50}, \code{boundary 50}) or with explicit arguments (\code{line\_sensors(num\_lines=64, orientation=horizontal)}), optionally weighted and joined by \code{+} into a mixture. The weight-expanded component cycle is rotated by a seed derived from the mask seed, the canonical spec and the record's stratum and indexed by the original generation-time stratum position. Assignments are fixed across epochs; selecting the training subset does not preserve exact balance (\appref{app:mixed-control}). The per-record mask seed uses the resolved protocol name only, hence \code{uniform 50} equals R50 and a component's mask does not depend on the other components. Specs are accepted in \code{data.mask.protocol}, \code{evaluation.observation\_protocols}, \code{--obs} and \code{api.make\_observation}; the paper namespace still rejects them (\code{docs/observation\_specs.md}).}
The legacy key \code{random\_3pct} can carry ratio 0.5; its string alone does not determine density.
For a nested-density control, retain explicitly constructed nested masks and use stored inputs.
\revcode{A mixture assigns one component per identity by stratum position and does not resample identities across epochs; the mixed-pattern rows of \appref{app:mixed-control} were trained with the nine frozen views as components, each with its own arguments, so every record's sensors are those of the corresponding single-pattern row.}
No nested-mask experiment is reported.

\subsection{Models, Budgets, and Splits}
\label{app:provide:models}
Model schemas are inspectable with \code{api.list\_models()}, \nolinkurl{api.describe_model(name)}, and command-line description tools.
\code{api.resolve\_model} validates names, parameters, task compatibility, presets, and structural constraints.
\nolinkurl{api.check_grid_compatibility} identifies incompatible spectral modes or branch grids.
Overrides yield a custom configuration rather than an unchanged paper preset.
The training facade enforces batch size at least two for U-FNO's normalization-based branches.
Equal input/output contracts do not imply equal parameter counts.

\noindent\begin{minipage}{\linewidth}
\begin{lstlisting}[language=Python,caption={Configuration-only inspection; no trained weights are loaded.},label={lst:model_config}]
from pdeobs import api
cfg = api.resolve_model("transolver", task="rollout",
                        preset="paper")
api.describe_model("gnot")
api.check_grid_compatibility(cfg, (128, 128))
\end{lstlisting}
\end{minipage}

\code{api.train} requires an explicit epoch or optimizer-step budget.
The archive's epoch budgets must not be relabelled as optimizer steps.
The generic default is \code{split="all"}, so held-out evidence requires an explicit split.
Routes include the exact \code{paper} split, stored metadata, deterministic \code{holdout:<fraction>}, and explicit train/test lists.
The facade records those identities and does not evaluate its held-out subset during training.
The base trainer can accept a validation loader directly, but \code{api.train} supplies none; naming a monitor \code{val\_loss} does not create a validation set.
New development subsets must come from training identities, never the paper test set.

\code{TrainingConfig} exposes objective weights, optimizer, learning rate, weight decay, clipping, schedules, precision, checkpointing, and health checks.
These are capabilities, not permission to reinterpret old attempts.
Retain a historical model's recorded final checkpoint rather than selecting one from test performance.
\revcode{Three seeds are independent: the campaign \code{seed} fixes the identity split and, by default, the sensors; \code{data.mask\_seed} pins the sensors on its own; \code{training.seed} seeds initialization and batch order. Moving \code{training.seed} alone keeps the split and every mask of a row; the convenience API seed still affects masks and splits as well as initialization, so use the explicit keys or the formal \code{paper-row} separation when changing only the training seed.}
\nolinkurl{configs/paper/training_cohorts.yaml} describes historical stopping semantics but does not implement every continuation controller.

\subsection{Inference Inputs and Checkpoint Loading}
\label{app:provide:infer}
\nolinkurl{api.inference_input_from_dataset} returns an observation package and targets separately.
The standard predictor receives masked values, masks, and required geometry, not hidden targets or PINO's training-loss condition.
Custom \code{InferenceInput} arrays are channel-last: $[N,H,W,C]$ for recovery or $[N,T_{\mathrm{hist}},H,W,C]$ for rollout, with aligned one-channel binary masks.
The constructor can generate placeholder identities or accept absent times; these fallbacks cannot establish correspondence with paper-test identities and source-frame indices.
An executable grid/horizon or extrapolated time label is not a newly validated forecasting range.

\nolinkurl{api.load_predictor} restores a new-format artifact from \code{model.json} and its weights, checking recorded weight hashes and refusing missing weights or mismatched structures by default.
Legacy \code{last.pt} files use \nolinkurl{api.load_legacy_checkpoint} with explicit architecture and task.
That adapter strictly matches model-state keys, cross-checks available task/physics metadata, and records unknown provenance fields.
It does not authenticate missing identities or reconstruct omitted optimizer state.
The release documentation describes stripped weights-only checkpoints: these support inference, not exact full-state continuation.
A successful load is necessary but insufficient to reproduce a historical score.

\subsection{Small End-to-End Example}
\label{app:provide:api}
Listing~\ref{lst:api} checks the local CPU workflow on eight new records and two optimizer steps.
It uses a general mask because frozen paper views are not defined on the demonstration grid.
It is not a member of the 441-setting experiment; its score must not enter a paper table.
Use fresh output directories, which are not overwritten by default.

\noindent\begin{minipage}{\linewidth}
\begin{lstlisting}[language=Python,caption={A small CPU example. Its two held-out demo records are not the 200 paper-test identities.},label={lst:api}]
import torch
from pdeobs import api
torch.set_num_threads(1)
data = api.create_dataset(
    pde="poisson", boundary="dirichlet", out="demo-records",
    num_samples=8, resolution=16, seed=17)
obs = api.make_observation("general:random", ratio=0.5)
run = api.train(
    data, task="recovery", model={"name": "fno", "preset": "smoke"},
    observation=obs, split="holdout:0.25",
    budget={"max_steps": 2}, seed=17, device="cpu",
    out="demo-training")
predictor = api.load_predictor(run.artifact.path, device="cpu")
inputs, targets = api.inference_input_from_dataset(
    data, obs, task="recovery", sample_ids=run.split.test_ids, seed=17)
bundle = api.predict(predictor, inputs)
score = api.evaluate(bundle, targets, target_ids=inputs.sample_ids,
                     target_time_indices=bundle.time_indices)
assert score["status"] == "valid"
\end{lstlisting}
\end{minipage}

\subsection{Formal Evaluation and Reproduction Boundary}
\label{app:provide:formal}
The convenience \code{api.evaluate} call builds its default contract from the supplied bundle.
It checks that supplied set, but cannot detect paper identities that were never declared.
Its default scope is a user evaluation, not a paper performance estimate.
For a formal block, freeze an independent contract with all 200 test identities, expected shape, frame coordinates, dataset/identity-manifest binding, observation definition, and checkpoint identity before prediction.
For legacy loading, explicitly supply the loaded file's digest; a convenience \code{checkpoint\_id} is not automatically populated from every legacy provenance format.

\code{pdeobs strict-infer} and \code{pdeobs strict-score} consume explicit materials and retain predictions for rescoring.
The \code{paper-row} original-500 path writes the split, completion records, contracts, predictions, and nine test blocks.
It requires all 2,000 identities across three regimes, not one partial shard.
Its template is \nolinkurl{configs/paper/row_original500.yaml}; campaign/registry paths must remain valid if copied.
Running it trains a new model, rather than verifying an archived score.
The other executable cohort, \code{demo}, is deliberately reduced.
Historical fixed-200, min-120 patience, and recovered attempts require separately versioned attempt logic; their labels alone are not executable reproduction instructions.

\subsection{Extension Contracts}
\label{app:provide:extend}
Registries expose masks, condition constructions, PDE generators, methods, and metrics.
Extensions must declare tasks and information permissions, use supplied randomness, return the expected schema, and preserve resolved parameters.
Registration does not certify numerical correctness or leakage freedom.
Listing~\ref{lst:extend} defines a custom observation; other contracts are in \nolinkurl{docs/extending.md} and \nolinkurl{src/pdeobs/registry.py}.
Changed layouts must not reuse frozen paper labels.

\noindent\begin{minipage}{\linewidth}
\begin{lstlisting}[language=Python,caption={A custom mask, not a new result under a frozen paper-view label.},label={lst:extend}]
import numpy as np
from pdeobs import api
from pdeobs.registry import MASK_REGISTRY
@MASK_REGISTRY.register("alternating_rows")
def alternating_rows(shape, rng, **kwargs):
    mask = np.zeros(shape, dtype=bool)
    mask[::2, :] = True
    return mask
spec = api.make_observation("alternating_rows", namespace="custom")
assert api.realized_count(spec, (128, 128)) == 8192
\end{lstlisting}
\end{minipage}

\subsection{Local Inspection and Tests}
\label{app:provide:ai}
\code{AI\_START\_HERE.md} is the implementation and validation index; \code{AGENTS.md} states the repository rules.
These inspection commands do not train models:

\noindent\begin{minipage}{\linewidth}
\begin{lstlisting}[language=bash,caption={Read-only implementation and evidence lookup.},label={lst:ai_interfaces}]
python tools/ai_assist.py doctor --json
python tools/ai_assist.py context --task scoring --max-chars 24000
\end{lstlisting}
\end{minipage}

Regression tests cover facade validation, scoring, observation replay, and evidence bundles.
Report the environment and passed/failed/skipped scope; an archived receipt is not a newly rerun test.
CPU examples do not certify GPU behavior, full-corpus fidelity, or historical model performance.

\subsection{Data Format, Benchmark Access, Maintenance, and Extensibility}
\label{app:provide:software}
\begin{movedblock}{Section~\ref{sec:software}}
\paragraph{Data format.}
To reuse physical records across patterns, HDF5 shards \addedcitep{collette2013hdf5} store channel-last condition, trajectory, and geometry arrays separately from observation masks. JSON metadata records identity, family, boundary, condition setting, regime, seeds, and times; atomic writes and hash sidecars track completed shards. Masks are then constructed at access time from the identity and protocol. Two versioned packages keep inference separate from scoring: an observation-only input produces a prediction bundle, which is scored against separately supplied targets. External sensor data can use the prediction interface, but quantitative scoring still requires corresponding targets (\appref{app:have:data}, \appref{app:provide:infer}).

\paragraph{Benchmark access.}
Code, configurations, tests and documentation are available at \weblink{\coderepo}{our public GitHub repository} under the MIT code license; data licensing is separate. The result release contains full-precision scores, per-record errors and hash bindings, with historical scores archived separately. Raw tensors are not embedded in Git. Reproducing this evaluation requires the bound records and weights, not merely newly generated data. Their integrity was checked in the verification environment, but unauthenticated external access remains a separate release requirement (\appref{app:validity}).

\paragraph{Maintenance.}
Python and command-line interfaces use NumPy \addedcitep{harris2020numpy}, SciPy \addedcitep{virtanen2020scipy}, h5py, and PyYAML, with optional PyTorch training support \addedcitep{paszke2019pytorch}. Exploration uses a convenience configuration resolver, whereas frozen experimental rows have stricter guards. Versioned schemas, resolved configurations, regression tests, CPU demonstrations, and a \code{doctor} command support inspection (\appref{app:have:software}). These checks establish software behavior, not the availability of every artifact needed to replay a historical run.

\paragraph{Extensibility.}
Public registries extend observation views, condition constructions, prediction methods, PDE families, and metrics. A view returns a boolean mask, a method implements the observation-to-prediction interface, and a PDE generator returns a complete record. This separation lets new methods and views reuse existing records and scoring without redefining the physical task. To protect the reference protocol, the nine frozen labels cannot be redefined; optional-plugin failures are recorded separately from built-in baselines. Registration itself does not certify a plugin's scientific validity. API contracts, entry-point discovery, and read-only inspection tools, including AI-assisted commands, are documented in \appref{app:provide}.
\end{movedblock}

\section{Metric Validity and Reproducibility}
\label{app:validity}
\subsection{A new prediction-level result version}
\revnew{All main-grid results use the prediction evaluation identified as \code{20260925-v1}, covering 441 retained checkpoints and 3,969 observation-pattern blocks with 200 held-out records per PDE. The additional five MIX models use the 26 September strict prediction rescoring and main-grid specialist references described in \appref{app:mixed-control}.} The verification made no optimizer updates, selected no new checkpoints, and changed no training setting. The original scalar archive remains preserved for provenance, but supplies no value to the current tables, figures, or conclusions.

The frozen scorer checks all prediction and target entries for finiteness and recomputes relative-$L^2$ errors in float64 without NaN-skipping reductions. Complete contracts bind record identities, mask, geometry, shape, checkpoint and configuration digests, data split, and frame mapping. For forecasting, stored positions 1--3 are distinguished from each record's internal solver indices and physical times. All 3,969 blocks contain their full 200-record population; no invalid block or failed model is omitted. Large finite errors remain valid poor outcomes, not an exclusion criterion.

The terminal audit freshly hashes every prediction HDF5 file and authenticates its full-array finiteness receipt against those identical bytes. It independently checks all identities and shapes, regenerates masks from the frozen identity/seed chain, compares geometry with the source, and verifies physical-time mappings. It also rehashes all 84 selected data shards, released checkpoints and resolved configurations against their pinned manifests. Thus the audit does not substitute a scalar-finiteness check for the scorer's array check. It performs neither another model forward pass nor another training run. Prediction-file validity, score reproduction, and scientific quality remain separate outcomes.

\subsection{Score changes and uncertainty}
Of the 3,969 blocks, \NewStrictCount{} meet the original score-comparison tolerance ($\mathrm{atol}=10^{-7}$, $\mathrm{rtol}=10^{-4}$); an additional \NewMinorCount{} differ by at most 1\% relative to the historical value, and \NewLargeCount{} differ by more than 1\%. The 1\% reporting threshold was adopted after initial verification results, not preregistered as a scientific validity gate. All new valid scores are used, whether better or worse, with no repeat inference to obtain closer agreement. The full old/new comparison remains an audit artifact. The specific numerical causes of the differences have not been established.

Each cell reports mean $\pm$ sample standard deviation of 200 per-record errors, with $\mathrm{ddof}=1$. Pair-level $D$ and $C$ instead summarize the means of 9 matched and 72 cross-pattern cells. The main table pools 63/504 such cells per PDE across seven methods and labels that different population explicitly. These standard deviations describe record or view variability, not multiple training seeds, confidence intervals, or independent replications. Dynamic joint error uses one norm over all three frames; the three auxiliary horizon errors are never averaged to replace it. Counts, medians and derived ratios receive no invented standard deviation. Full precision is retained before uniform display rounding.

\paragraph{\revnew{Paired test-identity bootstrap.}}
\label{app:bootstrap}
\begin{newrevision}
We resample the 200 test identities within each PDE's low/medium/high regimes (67/67/66 draws with replacement), using one shared draw across all its models, observation views and forecasting horizons. Cell means and derived statistics are recomputed in each of 5,000 replicates (seed 20260926). The 95\% percentile interval for the median $C/D$ over 49 pairs is [10.07, 10.71], with point estimate 10.46; the fixed 13-pair subset gives 10.74 [10.44, 11.12]. Each of the 49 pair-level $C/D$ intervals lies above one. These are marginal, conditional test-identity intervals for fixed trained models, not simultaneous coverage guarantees, training-seed uncertainty, or corrections for retrospective selection. The full 7,542-statistic output and resampling specification are in \nolinkurl{results/revision_v2/bootstrap_intervals.csv} and \nolinkurl{bootstrap_config.json}; no new model inference is used by this resampling.
\end{newrevision}

\subsection{Data, weights, and public availability}
Released-file manifests map each original checkpoint digest to the stripped release-file digest, including the retained Heat/Transolver/R50 checkpoint at epoch 257. Removing optimizer, scaler, RNG, or identifying metadata changes file bytes; original and released digests must therefore not be treated as interchangeable. Verified mappings, compatible model structures, exact weights, observations, splits and physical records are all required for inference reproducibility. Weights-only files cannot restore full-state training.

\begin{newrevision}
The \href{https://huggingface.co/datasets/PDE-OBS/pdeobs-data}{PDE-OBS data deposit} and \href{https://huggingface.co/PDE-OBS/pdeobs-models}{model deposit} were republished from de-identified copies on 26 September 2026. Direct manifests identify the \href{https://huggingface.co/datasets/PDE-OBS/pdeobs-data/resolve/main/release_manifest.json}{paper data subset}, \href{https://huggingface.co/datasets/PDE-OBS/pdeobs-data/resolve/main/release_manifest_full.json}{full data resource}, and model groups \href{https://huggingface.co/PDE-OBS/pdeobs-models/resolve/main/models_manifest.cluster-A.json}{A}, \href{https://huggingface.co/PDE-OBS/pdeobs-models/resolve/main/models_manifest.cluster-B.json}{B}, and \href{https://huggingface.co/PDE-OBS/pdeobs-models/resolve/main/models_manifest.cluster-C.json}{C}. The dated release report records unchanged numerical arrays and model parameters, unauthenticated access, and agreement with declared file digests for 20,167 dataset files and 6,477 model-deposit files (large objects by SHA-256, other files by Git blob identifier). The \code{scrub-manifest.json} files map original to republished digests where metadata changes alter bytes; the \href{https://huggingface.co/datasets/PDE-OBS/pdeobs-data/resolve/main/scrub-manifest.json}{data mapping} is linked here. This dated availability check is distinct from permanent access or an end-to-end reproduction from downloaded artifacts; the latter smoke test is not reported as completed. The checkpoint inventory covers the 441 main-grid models, not evidence of download coverage for the five additional MIX checkpoints. New aggregate scores, per-record errors and bindings accompany the result release; raw prediction tensors are not embedded in Git.
\end{newrevision}

\subsection{Training provenance and reproducible arithmetic}
The common framework fixes physical records, observation rules, the supervised objective, rollout interface and final-checkpoint convention. Per-setting budgets, stopping rules, continuations and exceptions remain documented in \nolinkurl{docs/training_template_and_records.md}; they are not retrospectively harmonized. The fixed 13-pair, 117-model original500 sensitivity subset additionally matches complete recorded configurations within each pair, excluding only the intended observation mask and run-name/data-root fields (\appref{app:training-sensitivity}). Equal epochs alone, used by a broader 23-pair historical summary, do not establish this stronger match. The analysis remains single-seed and retrospective.

The public artifact's \nolinkurl{results/prediction_verification_20260924/} contains the new index, per-record errors, full-precision statistics, dataset and checkpoint hash bindings, configuration-subset recheck, and a file manifest. The public \nolinkurl{tools/prediction_analysis/} scripts rebuild $D$, $C$, directional differences, ratios, horizon summaries and tables from this version without reopening models or rerunning inference. Every finite adverse outcome is retained. The immutable old archive is not a fallback for a missing new result. Independent reference-solution error and grid-convergence checks remain separate from these prediction and production-residual checks.

\clearpage
\section{Configuration-Matched Original-Budget Subset}
\label{app:training-sensitivity}
To examine sensitivity to heterogeneous training histories, we select pairs whose nine checkpoints all belong to the original500 cohort and actually reached epoch 500. This metadata-based selection predates the new prediction scores and fixes 13 pairs, 117 models and 1,053 blocks. No model is retrained and no pair is selected for a favorable test outcome.

\paragraph{Complete configuration audit.}
The released \code{resolved.yaml} of each selected model is freshly hashed and checked against the earlier release mapping and configuration audit. After removing only the row name, observation-mask specification and filesystem data root, the complete remaining configuration agrees across all nine views within each pair. This includes architecture, batch, loss, optimizer, learning rate and schedule, precision, gradient controls, stopping and loader settings. Dataset, split, seed and repository bindings also agree. It does not establish identical hardware, execution time or bitwise-deterministic training. The earlier field-level audit remains in \nolinkurl{results/training_sensitivity/}; the new result cut records its hash and the fresh input recheck.

\paragraph{Selection and new-score results.}
The subset includes CNO on Burgers, Darcy, Heat, Helmholtz, Navier--Stokes and reaction--diffusion; FNO and PINO on Darcy and Helmholtz; U-FNO on Darcy; and DeepONet and Transolver on Helmholtz. Table~\ref{tab:original500-sensitivity} compares the same descriptive statistics, using only new predictions in both columns. The 13 blue rows in the consolidated Table~\ref{tab:all-pairs} supply every selected pair's absolute means and standard deviations without duplicating them in a separate table.
\begin{table}[H]
\centering
\caption{Sensitivity to fixed original500 training provenance. Counts and medians summarize dependent comparisons, not independent experimental replications. The metadata-defined subset is unchanged.}
\label{tab:original500-sensitivity}
\begin{tabular*}{\linewidth}{@{\extracolsep{\fill}}lrr@{}}
\toprule
Statistic & Full grid & Fixed subset \\
\midrule
Pairs & 49 & 13 \\
Pairs with $C>D$ & 49/49 & 13/13 \\
Median $C/D$ & 10.46 & 10.74 \\
Positive layout directions, group I & 545/588 & 151/156 \\
Positive layout directions, group II & 276/294 & 76/78 \\
R80 improves over R50, R50-trained & 13/49 & 0/13 \\
R80 improves over R50, R65-trained & 37/49 & 8/13 \\
R80 improves over R50, R80-trained & 49/49 & 13/13 \\
\bottomrule
\end{tabular*}
\end{table}

The matched-versus-transfer direction holds in \NewSubsetPositive{} of the 13 pairs. Effect magnitudes are not identical to the full grid: the subset's median R80/R50 ratio for R50-trained checkpoints is $\NewSubsetDensityFiftyMedian$, versus $\NewDensityFiftyMedian$ over all 49 pairs. This is a retrospective, single-seed, completion-conditioned sensitivity analysis, not a randomized architecture comparison or evidence that omitted models behave identically.

\clearpage
\section{\revnew{Completed Diagnostics and Further Controls}}
\label{app:controls}
\revnew{Prediction-level verification enables the stationary diagnostics below without retraining. We distinguish these completed analyses and the MIX study from the further controls still needed for stronger interpretation.}

\paragraph{Stationary observed and hidden errors.}
For a noise-free stationary record, let $r=\widehat u-u$ and $d=\max(\|u\|_2,10^{-12})$. Define $e_{\rm obs}=\|M\odot r\|_2/d$ and $e_{\rm hid}=\|(1-M)\odot r\|_2/d$. Then $e^2=e_{\rm obs}^2+e_{\rm hid}^2$ per record. Squaring the separately averaged errors does not preserve this identity. A hidden-only relative error normalized by $\|(1-M)\odot u\|_2$ is a different metric and must be labelled separately.

The projection
\begin{equation}
\widehat u_{\rm dc}=M\odot u+(1-M)\odot\widehat u
\end{equation}
leaves hidden predictions unchanged and eliminates observed-point error. Full-field squared error, and thus the per-record full-field relative error, cannot increase. This algebraic diagnostic does not use hidden values to repair predictions. It applies to stationary noiseless recovery only: injecting true future values into a forecast would violate the task.

\begin{newrevision}
These diagnostics were computed from the retained predictions for all 1,701 stationary blocks (189 matched, 1,512 cross-pattern), each with 200 records. For each block the observed squared-error share is $\sum_i e_{{\rm obs},i}^2/\sum_i e_i^2$, not a fraction of incorrect grid points. Its median is 0.43 across matched blocks and 0.46 across cross-pattern blocks. The median ratio of projected to raw block-mean error is 0.74 and 0.73, respectively, with an overall range of 0.19--1.00. Matched Transolver blocks show a smaller projection effect (median ratio 0.96; observed squared-error share 0.04). Record-level checks verify the squared-error decomposition, equality of projected error to the hidden contribution, and non-increasing error. Thus some raw error lies at already observed points, but projection leaves hidden predictions unchanged and does not explain their error. Raw predictions remain the headline results; projection is a separately labelled diagnostic, recorded in \nolinkurl{results/revision_v2/stationary_diagnostics_summary.csv} and \nolinkurl{stationary_diagnostics_checks.json}.
\end{newrevision}

\paragraph{Simple and mixed-pattern baselines.}
A training-population mean, interpolation, or a training-only POD basis can test whether smooth stationary fields are easily reconstructed without the learned interface.
The software's existing \code{mean} method is instead an observed-mean fill, and must not be labelled a training-population baseline.
Interpolation alone does not predict temporal evolution; a forecasting comparator must state its temporal rule.
\revnew{Five completed MIX models, their actual epochs and update counts, and the distinction between all-nine descriptive and Poisson recipe-matched comparisons are reported in \appref{app:mixed-control}. The formal-protocol simple baselines remain unexecuted; no result is inferred from their implementation.}

\paragraph{Forecast ambiguity and density.}
A full-input versus partial-input, hidden-parameter versus parameter-conditioned comparison can separate sources of temporal uncertainty. Conditioning on a recorded physical coefficient is an explicitly different task, not an unreported change to ordinary inference. Nested random masks isolate the addition of observations by preserving the smaller set. These controls require new evidence before any mechanistic claim.

\paragraph{Uncertainty and sensitivity.}
\revnew{The paired test-identity bootstrap is reported in \appref{app:bootstrap}; independent training seeds are still needed for training-seed uncertainty.} Output variance and sensitivity to changed visible values can diagnose an input-insensitive model. Near-equal aggregate scores alone neither prove collapse nor demonstrate useful robustness.

\clearpage
\section{Complete Prediction-Level Results}
\label{app:distributions}
All 49 PDE--method matrices use the new complete prediction-level evaluation, retaining every finite adverse result. The tables expand Figure~\ref{fig:cross-observation} without selecting a different subset. Each block contains 200 held-out identities and passes the file-validity checks in \appref{app:validity}; this does not imply small prediction error. Means are arithmetic, standard deviations use $\mathrm{ddof}=1$, and the 90th percentile uses linear interpolation. \revnew{These tables use two decimal places with matching mean/SD precision; counts remain integers. MIX errors are explicitly scaled to percentages in \appref{app:mixed-control}, while these grid errors and all ratios retain their stated units.} All calculations precede display rounding. A displayed $0.00$ can represent a small positive value, not exact zero; the released scores retain full precision.

\subsection{PDE-level overview}
Table~\ref{tab:main-summary} complements the shared absolute-error color scale in Figure~\ref{fig:cross-observation}. It retains variability across all seven methods and observation cells rather than assigning a winner. These pooled descriptive summaries do not equalize training histories or model capacities.
\begin{table}[H]
\centering
\caption{Absolute error and transfer sensitivity from new predictions. Matched/cross-pattern entries are mean $\pm$ sample SD over 63/504 cell means per PDE (seven methods); the ratio is the median of seven pairwise $C/D$ values. These are not error percentages, seed uncertainty, or an architecture ranking.}\label{tab:main-summary}
\begin{tabular*}{\linewidth}{@{\extracolsep{\fill}}lrrr@{}}
\toprule
 & \multicolumn{2}{c}{Relative-$L^2$ test error} & Transfer ratio \\
\cmidrule(lr){2-3}\cmidrule(l){4-4}
PDE & Matched & Cross-pattern & Median $C/D$ \\
\midrule
\rowcolor{PDEGray}\multicolumn{4}{l}{\textcolor{PDEInk}{Stationary solution recovery}} \\
Darcy & $0.08 \pm 0.10$ & $0.80 \pm 2.40$ & 13.45 \\
Poisson & $0.06 \pm 0.07$ & $1.50 \pm 5.71$ & 16.58 \\
Helmholtz & $0.05 \pm 0.14$ & $4.91 \pm 30.45$ & 28.61 \\
\rowcolor{PDEGray}\multicolumn{4}{l}{\textcolor{PDEInk}{Temporal state forecasting}} \\
Heat & $0.09 \pm 0.07$ & $1.16 \pm 6.77$ & 3.62 \\
Reaction--diffusion & $0.11 \pm 0.12$ & $1.19 \pm 5.60$ & 3.36 \\
Burgers & $0.10 \pm 0.11$ & $0.51 \pm 1.65$ & 3.63 \\
Navier--Stokes & $0.04 \pm 0.04$ & $0.34 \pm 0.50$ & 7.39 \\
\bottomrule
\end{tabular*}
\end{table}

\clearpage

\subsection{Training cohorts and aggregation}
\begin{table}[H]
\centering
\caption{Retained training-provenance groups. Codes identify histories, not scientific quality; every retained model contributes all nine test views.}\label{tab:result-cohorts}
\begin{tabular*}{\linewidth}{@{\extracolsep{\fill}}llr@{}}
\toprule
Code & Recorded cohort & Models \\
\midrule
O500 & original500 & 257 \\
INT & interrupted\_or\_recovered & 28 \\
M120 & budget200\_min120 & 44 \\
P10 & budget200\_patience10\_no\_floor & 10 \\
R500 & original500\_recovered & 66 \\
F200 & fixed200\_recovered & 36 \\
\bottomrule
\end{tabular*}
\end{table}

For each matrix, $D$ averages nine matched cell means and $C$ averages 72 cross-pattern cell means. Their standard deviations describe those 9/72 values, not seed uncertainty. The main-text median $C/D$ weights each of the 49 matrices equally, not each physical record pooled across PDEs. Each colored cell in Figure~\ref{fig:cross-observation} is one $E_{v,w}$, not a ratio or a median across methods. The logarithmic color scale preserves all 3,969 positive finite cell means without clipping; their rounded values and sample SDs appear below. No error or ratio is winsorized.

\begin{table}[H]
\centering
\caption{Layout and density contrasts. The first group counts $E_{v,w}>E_{w,w}$ and summarizes $E_{v,w}/E_{w,w}$; the second counts $E_{v,w}<E_{v,\mathrm{R50}}$ and summarizes $E_{v,w}/E_{v,\mathrm{R50}}$. Density masks are sampled separately and are not constructed as nested sets; statistical independence is not assumed. Counts are integers; ratios are dimensionless.}
\label{tab:equal-count}
\label{tab:density-directions}
\begin{tabular*}{\linewidth}{@{\extracolsep{\fill}}llrr@{}}
\toprule
Condition & Comparison & Count / total & Median ratio \\
\midrule
\rowcolor{PDEGray}\multicolumn{4}{l}{Equal-count layout: positive destination-referenced difference}\\
$K=8,192$ & Recovery & 238/252 & 7.53 \\
$K=8,192$ & Rollout & 307/336 & 2.97 \\
$K=8,284$ & Recovery & 122/126 & 12.37 \\
$K=8,284$ & Rollout & 154/168 & 3.76 \\
\midrule
\rowcolor{PDEGray}\multicolumn{4}{l}{Density: lower error than the same checkpoint at R50}\\
Train R50 & Test R65 & 15/49 & 1.13 \\
Train R50 & Test R80 & 13/49 & 1.58 \\
Train R65 & Test R65 & 48/49 & 0.61 \\
Train R65 & Test R80 & 37/49 & 0.93 \\
Train R80 & Test R65 & 49/49 & 0.56 \\
Train R80 & Test R80 & 49/49 & 0.29 \\
\bottomrule
\end{tabular*}
\end{table}

\clearpage
\subsection{All pair-level summaries}
\begingroup
\setlength{\tabcolsep}{4pt}
\setlength{\LTcapwidth}{\linewidth}
\begin{longtable}{@{}lrrrrrr@{}}
\caption{All 49 PDE--method summaries. Relative-$L^2$ errors are displayed to two decimals. $D$/$C$ are mean $\pm$ sample SD across 9/72 cell means; the last three columns describe the 72 cross-pattern means. Blue retains the 13 fixed original500 pairs, not a winner or a uniform-budget architecture ranking. All medians, ratios, and tails are calculated before rounding.}\label{tab:all-pairs}\\
\toprule
 & \multicolumn{2}{c}{Mean $\pm$ SD} &  & \multicolumn{3}{c}{Cross-pattern cell means} \\
\cmidrule(lr){2-3}\cmidrule(l){5-7}
Method & Matched $D$ & Cross $C$ & $C/D$ & Median & 90th pct. & Max. \\
\midrule
\endfirsthead
\multicolumn{7}{l}{\tablename~\thetable{} (continued)}\\
\toprule
 & \multicolumn{2}{c}{Mean $\pm$ SD} &  & \multicolumn{3}{c}{Cross-pattern cell means} \\
\cmidrule(lr){2-3}\cmidrule(l){5-7}
Method & Matched $D$ & Cross $C$ & $C/D$ & Median & 90th pct. & Max. \\
\midrule
\endhead
\midrule
\multicolumn{7}{r}{Continued on the next page}\\
\endfoot
\bottomrule
\endlastfoot
\rowcolor{PDEGray}\multicolumn{7}{l}{\textcolor{PDEInk}{Darcy}\label{tab:pair-darcy}}\\*
\rowcolor{PDEFocus}
U-FNO & $0.26 \pm 0.10$ & $0.51 \pm 0.32$ & 2.01 & 0.36 & 1.07 & 1.49 \\*
\rowcolor{PDEFocus}
FNO & $0.08 \pm 0.11$ & $0.81 \pm 0.86$ & 10.46 & 0.54 & 1.80 & 4.39 \\*
\rowcolor{PDEFocus}
CNO & $0.04 \pm 0.05$ & $0.48 \pm 0.53$ & 10.74 & 0.37 & 0.82 & 3.63 \\*
DeepONet & $0.06 \pm 0.04$ & $1.58 \pm 2.95$ & 26.63 & 0.44 & 3.67 & 14.26 \\*
GNOT & $0.06 \pm 0.04$ & $0.74 \pm 3.21$ & 13.45 & 0.10 & 0.54 & 26.18 \\*
Transolver & $0.03 \pm 0.05$ & $0.51 \pm 0.46$ & 15.30 & 0.46 & 1.10 & 2.33 \\*
\rowcolor{PDEFocus}
PINO & $0.03 \pm 0.04$ & $0.99 \pm 4.42$ & 28.83 & 0.36 & 1.04 & 37.80 \\
\rowcolor{PDEGray}\multicolumn{7}{l}{\textcolor{PDEInk}{Poisson}\label{tab:pair-poisson}}\\*
U-FNO & $0.12 \pm 0.10$ & $0.93 \pm 1.14$ & 8.01 & 0.47 & 3.07 & 4.11 \\*
FNO & $0.04 \pm 0.05$ & $0.70 \pm 0.70$ & 16.58 & 0.49 & 1.45 & 3.01 \\*
CNO & $0.04 \pm 0.05$ & $0.48 \pm 0.45$ & 11.96 & 0.35 & 1.13 & 1.85 \\*
DeepONet & $0.06 \pm 0.07$ & $1.79 \pm 3.10$ & 30.16 & 0.53 & 6.49 & 13.65 \\*
GNOT & $0.06 \pm 0.09$ & $5.20 \pm 14.19$ & 91.12 & 0.18 & 9.66 & 85.57 \\*
Transolver & $0.04 \pm 0.05$ & $0.80 \pm 0.62$ & 18.85 & 0.67 & 1.70 & 2.60 \\*
PINO & $0.04 \pm 0.04$ & $0.60 \pm 0.46$ & 14.90 & 0.50 & 1.06 & 2.25 \\
\rowcolor{PDEGray}\multicolumn{7}{l}{\textcolor{PDEInk}{Helmholtz}\label{tab:pair-helmholtz}}\\*
U-FNO & $0.17 \pm 0.34$ & $0.78 \pm 0.67$ & 4.70 & 0.53 & 1.63 & 2.89 \\*
\rowcolor{PDEFocus}
FNO & $0.02 \pm 0.02$ & $0.61 \pm 0.56$ & 28.61 & 0.53 & 1.38 & 2.12 \\*
\rowcolor{PDEFocus}
CNO & $0.03 \pm 0.04$ & $0.45 \pm 0.36$ & 15.35 & 0.40 & 0.90 & 1.15 \\*
\rowcolor{PDEFocus}
DeepONet & $0.05 \pm 0.05$ & $1.71 \pm 3.12$ & 34.16 & 0.48 & 4.54 & 15.28 \\*
GNOT & $0.05 \pm 0.07$ & $29.75 \pm 76.34$ & 575.98 & 0.16 & 127.29 & 494.98 \\*
\rowcolor{PDEFocus}
Transolver & $0.02 \pm 0.02$ & $0.58 \pm 0.39$ & 32.52 & 0.52 & 0.99 & 1.93 \\*
\rowcolor{PDEFocus}
PINO & $0.04 \pm 0.03$ & $0.53 \pm 0.39$ & 13.00 & 0.35 & 1.13 & 1.42 \\
\rowcolor{PDEGray}\multicolumn{7}{l}{\textcolor{PDEInk}{Heat}\label{tab:pair-heat}}\\*
U-FNO & $0.17 \pm 0.12$ & $0.33 \pm 0.23$ & 1.94 & 0.24 & 0.65 & 0.93 \\*
FNO & $0.05 \pm 0.01$ & $0.20 \pm 0.20$ & 3.60 & 0.09 & 0.53 & 0.80 \\*
\rowcolor{PDEFocus}
CNO & $0.07 \pm 0.02$ & $0.27 \pm 0.23$ & 3.83 & 0.18 & 0.63 & 0.79 \\*
DeepONet & $0.10 \pm 0.04$ & $6.40 \pm 17.06$ & 65.14 & 0.73 & 17.37 & 93.63 \\*
GNOT & $0.10 \pm 0.02$ & $0.39 \pm 0.78$ & 4.01 & 0.10 & 0.63 & 4.59 \\*
Transolver & $0.10 \pm 0.09$ & $0.36 \pm 0.24$ & 3.62 & 0.34 & 0.74 & 0.87 \\*
PINO & $0.06 \pm 0.01$ & $0.22 \pm 0.20$ & 3.53 & 0.11 & 0.55 & 0.66 \\
\rowcolor{PDEGray}\multicolumn{7}{l}{\textcolor{PDEInk}{Reaction--diffusion}\label{tab:pair-reaction_diffusion}}\\*
U-FNO & $0.14 \pm 0.07$ & $0.31 \pm 0.31$ & 2.25 & 0.22 & 0.58 & 1.72 \\*
FNO & $0.06 \pm 0.01$ & $0.20 \pm 0.21$ & 3.36 & 0.10 & 0.43 & 0.97 \\*
\rowcolor{PDEFocus}
CNO & $0.07 \pm 0.02$ & $0.25 \pm 0.24$ & 3.58 & 0.12 & 0.62 & 0.84 \\*
DeepONet & $0.11 \pm 0.03$ & $4.72 \pm 12.99$ & 42.79 & 0.53 & 12.91 & 77.24 \\*
GNOT & $0.11 \pm 0.04$ & $2.20 \pm 5.96$ & 20.05 & 0.13 & 6.23 & 33.23 \\*
Transolver & $0.20 \pm 0.29$ & $0.44 \pm 0.32$ & 2.19 & 0.36 & 0.96 & 1.06 \\*
PINO & $0.07 \pm 0.01$ & $0.22 \pm 0.20$ & 3.29 & 0.10 & 0.57 & 0.67 \\
\rowcolor{PDEGray}\multicolumn{7}{l}{\textcolor{PDEInk}{Burgers}\label{tab:pair-burgers}}\\*
U-FNO & $0.21 \pm 0.26$ & $0.32 \pm 0.31$ & 1.50 & 0.16 & 0.80 & 1.23 \\*
FNO & $0.05 \pm 0.01$ & $0.19 \pm 0.18$ & 3.70 & 0.08 & 0.49 & 0.61 \\*
\rowcolor{PDEFocus}
CNO & $0.07 \pm 0.03$ & $0.26 \pm 0.24$ & 3.89 & 0.14 & 0.66 & 0.71 \\*
DeepONet & $0.12 \pm 0.03$ & $1.99 \pm 4.03$ & 17.11 & 0.67 & 5.32 & 20.33 \\*
GNOT & $0.09 \pm 0.03$ & $0.28 \pm 0.35$ & 3.06 & 0.10 & 0.73 & 2.15 \\*
Transolver & $0.10 \pm 0.06$ & $0.33 \pm 0.23$ & 3.18 & 0.30 & 0.69 & 0.86 \\*
PINO & $0.05 \pm 0.01$ & $0.20 \pm 0.19$ & 3.63 & 0.10 & 0.54 & 0.58 \\
\rowcolor{PDEGray}\multicolumn{7}{l}{\textcolor{PDEInk}{Navier--Stokes}\label{tab:pair-navier_stokes}}\\*
U-FNO & $0.03 \pm 0.03$ & $0.18 \pm 0.18$ & 7.09 & 0.07 & 0.47 & 0.55 \\*
FNO & $0.01 \pm 0.01$ & $0.18 \pm 0.19$ & 17.47 & 0.08 & 0.51 & 0.56 \\*
\rowcolor{PDEFocus}
CNO & $0.04 \pm 0.04$ & $0.29 \pm 0.28$ & 7.18 & 0.14 & 0.69 & 1.01 \\*
DeepONet & $0.07 \pm 0.05$ & $0.76 \pm 0.96$ & 11.31 & 0.52 & 1.99 & 4.18 \\*
GNOT & $0.06 \pm 0.06$ & $0.17 \pm 0.17$ & 2.65 & 0.08 & 0.45 & 0.63 \\*
Transolver & $0.05 \pm 0.04$ & $0.40 \pm 0.30$ & 7.39 & 0.37 & 0.79 & 0.94 \\*
PINO & $0.01 \pm 0.02$ & $0.43 \pm 0.56$ & 38.18 & 0.20 & 1.27 & 2.12 \\
\end{longtable}
\endgroup

\clearpage
\subsection{All forecasting horizons}
\label{app:horizons}
The 28 temporal matrices each include three horizon diagnostics. The mean $\pm$ sample SD of 9 matched or 72 cross-pattern cell means is reported at every horizon. These auxiliary errors are not averaged to obtain the joint three-frame score. Three predicted transitions do not establish long-term stability.
\begingroup
\setlength{\tabcolsep}{4pt}
\setlength{\LTcapwidth}{\linewidth}
\begin{longtable}{@{}lrrrrrr@{}}
\caption{All forecasting horizons in one grouped table. Each entry reports mean $\pm$ sample SD of 9 matched or 72 cross-pattern cell means on one line, with two decimals in both terms. Gray retains the method grouping. These relative-$L^2$ errors are auxiliary per-frame scores, not components to average into the joint rollout score.}\label{tab:all-horizons}\\
\toprule
 & \multicolumn{2}{c}{Horizon 1} & \multicolumn{2}{c}{Horizon 2} & \multicolumn{2}{c}{Horizon 3} \\
\cmidrule(lr){2-3}\cmidrule(lr){4-5}\cmidrule(l){6-7}
Method & $D_1$ & $C_1$ & $D_2$ & $C_2$ & $D_3$ & $C_3$ \\
\midrule
\endfirsthead
\multicolumn{7}{l}{\tablename~\thetable{} (continued)}\\
\toprule
 & \multicolumn{2}{c}{Horizon 1} & \multicolumn{2}{c}{Horizon 2} & \multicolumn{2}{c}{Horizon 3} \\
\cmidrule(lr){2-3}\cmidrule(lr){4-5}\cmidrule(l){6-7}
Method & $D_1$ & $C_1$ & $D_2$ & $C_2$ & $D_3$ & $C_3$ \\
\midrule
\endhead
\midrule
\multicolumn{7}{r}{Continued on the next page}\\
\endfoot
\bottomrule
\endlastfoot
\rowcolor{PDEGray}\multicolumn{7}{l}{\textcolor{PDEInk}{Heat}\label{tab:horizon-heat}}\\*
\cellcolor{PDEGray}U-FNO & \PDEMeanSD{0.16}{0.12} & \PDEMeanSD{0.33}{0.24} & \PDEMeanSD{0.16}{0.12} & \PDEMeanSD{0.32}{0.23} & \PDEMeanSD{0.18}{0.11} & \PDEMeanSD{0.32}{0.22} \\*
\cellcolor{PDEGray}FNO & \PDEMeanSD{0.03}{0.01} & \PDEMeanSD{0.19}{0.21} & \PDEMeanSD{0.05}{0.01} & \PDEMeanSD{0.19}{0.21} & \PDEMeanSD{0.08}{0.01} & \PDEMeanSD{0.20}{0.20} \\*
\cellcolor{PDEGray}CNO & \PDEMeanSD{0.05}{0.03} & \PDEMeanSD{0.26}{0.25} & \PDEMeanSD{0.06}{0.02} & \PDEMeanSD{0.26}{0.23} & \PDEMeanSD{0.09}{0.02} & \PDEMeanSD{0.27}{0.22} \\*
\cellcolor{PDEGray}DeepONet & \PDEMeanSD{0.08}{0.04} & \PDEMeanSD{1.40}{1.97} & \PDEMeanSD{0.09}{0.04} & \PDEMeanSD{2.89}{6.12} & \PDEMeanSD{0.12}{0.04} & \PDEMeanSD{10.66}{29.76} \\*
\cellcolor{PDEGray}GNOT & \PDEMeanSD{0.08}{0.02} & \PDEMeanSD{0.25}{0.29} & \PDEMeanSD{0.09}{0.01} & \PDEMeanSD{0.30}{0.46} & \PDEMeanSD{0.11}{0.01} & \PDEMeanSD{0.50}{1.24} \\*
\cellcolor{PDEGray}Transolver & \PDEMeanSD{0.08}{0.12} & \PDEMeanSD{0.36}{0.24} & \PDEMeanSD{0.09}{0.09} & \PDEMeanSD{0.35}{0.24} & \PDEMeanSD{0.11}{0.08} & \PDEMeanSD{0.35}{0.23} \\*
\cellcolor{PDEGray}PINO & \PDEMeanSD{0.03}{0.01} & \PDEMeanSD{0.20}{0.21} & \PDEMeanSD{0.06}{0.01} & \PDEMeanSD{0.21}{0.20} & \PDEMeanSD{0.08}{0.01} & \PDEMeanSD{0.23}{0.19} \\
\rowcolor{PDEGray}\multicolumn{7}{l}{\textcolor{PDEInk}{Reaction--diffusion}\label{tab:horizon-reaction_diffusion}}\\*
\cellcolor{PDEGray}U-FNO & \PDEMeanSD{0.14}{0.06} & \PDEMeanSD{0.30}{0.27} & \PDEMeanSD{0.14}{0.07} & \PDEMeanSD{0.31}{0.33} & \PDEMeanSD{0.14}{0.07} & \PDEMeanSD{0.32}{0.33} \\*
\cellcolor{PDEGray}FNO & \PDEMeanSD{0.04}{0.01} & \PDEMeanSD{0.21}{0.22} & \PDEMeanSD{0.06}{0.01} & \PDEMeanSD{0.19}{0.20} & \PDEMeanSD{0.07}{0.01} & \PDEMeanSD{0.19}{0.20} \\*
\cellcolor{PDEGray}CNO & \PDEMeanSD{0.06}{0.03} & \PDEMeanSD{0.25}{0.25} & \PDEMeanSD{0.07}{0.02} & \PDEMeanSD{0.25}{0.24} & \PDEMeanSD{0.08}{0.02} & \PDEMeanSD{0.25}{0.23} \\*
\cellcolor{PDEGray}DeepONet & \PDEMeanSD{0.09}{0.04} & \PDEMeanSD{1.45}{2.64} & \PDEMeanSD{0.11}{0.03} & \PDEMeanSD{2.72}{6.56} & \PDEMeanSD{0.13}{0.03} & \PDEMeanSD{7.48}{21.69} \\*
\cellcolor{PDEGray}GNOT & \PDEMeanSD{0.10}{0.05} & \PDEMeanSD{0.63}{1.25} & \PDEMeanSD{0.11}{0.04} & \PDEMeanSD{2.93}{8.66} & \PDEMeanSD{0.12}{0.04} & \PDEMeanSD{1.94}{5.05} \\*
\cellcolor{PDEGray}Transolver & \PDEMeanSD{0.19}{0.28} & \PDEMeanSD{0.44}{0.30} & \PDEMeanSD{0.20}{0.30} & \PDEMeanSD{0.43}{0.32} & \PDEMeanSD{0.21}{0.30} & \PDEMeanSD{0.44}{0.33} \\*
\cellcolor{PDEGray}PINO & \PDEMeanSD{0.05}{0.01} & \PDEMeanSD{0.20}{0.21} & \PDEMeanSD{0.07}{0.01} & \PDEMeanSD{0.21}{0.20} & \PDEMeanSD{0.08}{0.01} & \PDEMeanSD{0.23}{0.21} \\
\rowcolor{PDEGray}\multicolumn{7}{l}{\textcolor{PDEInk}{Burgers}\label{tab:horizon-burgers}}\\*
\cellcolor{PDEGray}U-FNO & \PDEMeanSD{0.16}{0.15} & \PDEMeanSD{0.28}{0.26} & \PDEMeanSD{0.19}{0.23} & \PDEMeanSD{0.30}{0.29} & \PDEMeanSD{0.27}{0.38} & \PDEMeanSD{0.36}{0.38} \\*
\cellcolor{PDEGray}FNO & \PDEMeanSD{0.02}{0.01} & \PDEMeanSD{0.18}{0.19} & \PDEMeanSD{0.04}{0.01} & \PDEMeanSD{0.17}{0.18} & \PDEMeanSD{0.07}{0.01} & \PDEMeanSD{0.19}{0.17} \\*
\cellcolor{PDEGray}CNO & \PDEMeanSD{0.04}{0.04} & \PDEMeanSD{0.26}{0.25} & \PDEMeanSD{0.06}{0.03} & \PDEMeanSD{0.25}{0.24} & \PDEMeanSD{0.09}{0.02} & \PDEMeanSD{0.27}{0.22} \\*
\cellcolor{PDEGray}DeepONet & \PDEMeanSD{0.07}{0.04} & \PDEMeanSD{1.28}{1.79} & \PDEMeanSD{0.10}{0.04} & \PDEMeanSD{1.68}{3.12} & \PDEMeanSD{0.16}{0.03} & \PDEMeanSD{2.67}{6.15} \\*
\cellcolor{PDEGray}GNOT & \PDEMeanSD{0.07}{0.03} & \PDEMeanSD{0.25}{0.30} & \PDEMeanSD{0.08}{0.03} & \PDEMeanSD{0.26}{0.35} & \PDEMeanSD{0.12}{0.02} & \PDEMeanSD{0.30}{0.37} \\*
\cellcolor{PDEGray}Transolver & \PDEMeanSD{0.07}{0.05} & \PDEMeanSD{0.33}{0.23} & \PDEMeanSD{0.10}{0.06} & \PDEMeanSD{0.32}{0.23} & \PDEMeanSD{0.13}{0.07} & \PDEMeanSD{0.34}{0.23} \\*
\cellcolor{PDEGray}PINO & \PDEMeanSD{0.02}{0.01} & \PDEMeanSD{0.18}{0.20} & \PDEMeanSD{0.05}{0.01} & \PDEMeanSD{0.19}{0.19} & \PDEMeanSD{0.08}{0.01} & \PDEMeanSD{0.22}{0.18} \\
\rowcolor{PDEGray}\multicolumn{7}{l}{\textcolor{PDEInk}{Navier--Stokes}\label{tab:horizon-navier_stokes}}\\*
\cellcolor{PDEGray}U-FNO & \PDEMeanSD{0.03}{0.03} & \PDEMeanSD{0.19}{0.18} & \PDEMeanSD{0.02}{0.03} & \PDEMeanSD{0.18}{0.18} & \PDEMeanSD{0.03}{0.03} & \PDEMeanSD{0.18}{0.18} \\*
\cellcolor{PDEGray}FNO & \PDEMeanSD{0.01}{0.01} & \PDEMeanSD{0.19}{0.19} & \PDEMeanSD{0.01}{0.01} & \PDEMeanSD{0.17}{0.19} & \PDEMeanSD{0.01}{0.01} & \PDEMeanSD{0.17}{0.19} \\*
\cellcolor{PDEGray}CNO & \PDEMeanSD{0.04}{0.05} & \PDEMeanSD{0.30}{0.30} & \PDEMeanSD{0.04}{0.04} & \PDEMeanSD{0.28}{0.28} & \PDEMeanSD{0.04}{0.04} & \PDEMeanSD{0.28}{0.27} \\*
\cellcolor{PDEGray}DeepONet & \PDEMeanSD{0.06}{0.05} & \PDEMeanSD{0.79}{1.02} & \PDEMeanSD{0.07}{0.05} & \PDEMeanSD{0.71}{0.85} & \PDEMeanSD{0.07}{0.05} & \PDEMeanSD{0.75}{0.97} \\*
\cellcolor{PDEGray}GNOT & \PDEMeanSD{0.06}{0.06} & \PDEMeanSD{0.18}{0.18} & \PDEMeanSD{0.06}{0.05} & \PDEMeanSD{0.16}{0.16} & \PDEMeanSD{0.07}{0.06} & \PDEMeanSD{0.16}{0.15} \\*
\cellcolor{PDEGray}Transolver & \PDEMeanSD{0.05}{0.04} & \PDEMeanSD{0.39}{0.29} & \PDEMeanSD{0.05}{0.04} & \PDEMeanSD{0.40}{0.30} & \PDEMeanSD{0.06}{0.04} & \PDEMeanSD{0.41}{0.31} \\*
\cellcolor{PDEGray}PINO & \PDEMeanSD{0.01}{0.02} & \PDEMeanSD{0.34}{0.46} & \PDEMeanSD{0.01}{0.02} & \PDEMeanSD{0.43}{0.58} & \PDEMeanSD{0.01}{0.02} & \PDEMeanSD{0.48}{0.63} \\
\end{longtable}
\endgroup

\clearpage
\subsection{All 49 train-view by test-view matrices}
\label{app:all-matrices}
Rows are training patterns and columns are test patterns, ordered R50, H, V, CL, BL, LI, BD, R65, R80. Each complete $9\times9$ matrix is shown in one table without splitting its columns. Each cell reports mean $\pm$ sample SD of 200 per-record relative-$L^2$ errors on one line, with two decimal places in both terms. Blue identifies matched observations, not a best-method ranking. Captions record the actual epoch and cohort of every training-pattern checkpoint using Table~\ref{tab:result-cohorts}; row order also determines the order of the epoch/cohort list. No budgets are silently mixed. CNO denotes the CNO-inspired adaptation. Full-precision per-record scores accompany the artifact; no cell is an estimate or placeholder.
\begingroup
\setlength{\tabcolsep}{1pt}
\begin{table}[H]
\centering
\caption{Darcy / U-FNO. All 81 cells: mean $\pm$ SD on one line ($n=200$); blue marks matched views. All training views: 500/O500.}
\label{tab:matrix-darcy-ufno-2d}
\PDETableFont
\begin{tabular*}{\linewidth}{@{\extracolsep{\fill}}lrrrrrrrrr@{}}
\toprule
 & \multicolumn{9}{c}{Test observation pattern} \\
\cmidrule(l){2-10}
Train & R50 & H & V & CL & BL & LI & BD & R65 & R80 \\
\midrule
R50 & \cellcolor{PDEFocus}\PDEMeanSD{0.18}{0.03} & \PDEMeanSD{0.25}{0.06} & \PDEMeanSD{0.24}{0.06} & \PDEMeanSD{0.76}{0.15} & \PDEMeanSD{0.76}{0.07} & \PDEMeanSD{0.24}{0.05} & \PDEMeanSD{0.73}{0.06} & \PDEMeanSD{0.31}{0.02} & \PDEMeanSD{0.57}{0.04} \\
H & \PDEMeanSD{0.32}{0.05} & \cellcolor{PDEFocus}\PDEMeanSD{0.32}{0.05} & \PDEMeanSD{0.38}{0.07} & \PDEMeanSD{1.08}{0.30} & \PDEMeanSD{1.49}{0.30} & \PDEMeanSD{0.35}{0.07} & \PDEMeanSD{1.47}{0.26} & \PDEMeanSD{0.31}{0.06} & \PDEMeanSD{0.32}{0.05} \\
V & \PDEMeanSD{0.36}{0.18} & \PDEMeanSD{0.36}{0.18} & \cellcolor{PDEFocus}\PDEMeanSD{0.36}{0.18} & \PDEMeanSD{0.36}{0.18} & \PDEMeanSD{0.36}{0.18} & \PDEMeanSD{0.36}{0.18} & \PDEMeanSD{0.36}{0.18} & \PDEMeanSD{0.36}{0.18} & \PDEMeanSD{0.36}{0.18} \\
CL & \PDEMeanSD{0.27}{0.14} & \PDEMeanSD{0.27}{0.14} & \PDEMeanSD{0.27}{0.14} & \cellcolor{PDEFocus}\PDEMeanSD{0.30}{0.14} & \PDEMeanSD{0.30}{0.15} & \PDEMeanSD{0.27}{0.14} & \PDEMeanSD{0.31}{0.16} & \PDEMeanSD{0.29}{0.12} & \PDEMeanSD{0.34}{0.10} \\
BL & \PDEMeanSD{0.36}{0.18} & \PDEMeanSD{0.36}{0.18} & \PDEMeanSD{0.36}{0.18} & \PDEMeanSD{0.36}{0.18} & \cellcolor{PDEFocus}\PDEMeanSD{0.36}{0.18} & \PDEMeanSD{0.36}{0.18} & \PDEMeanSD{0.36}{0.18} & \PDEMeanSD{0.36}{0.18} & \PDEMeanSD{0.36}{0.18} \\
LI & \PDEMeanSD{0.08}{0.02} & \PDEMeanSD{0.09}{0.02} & \PDEMeanSD{0.09}{0.02} & \PDEMeanSD{0.57}{0.14} & \PDEMeanSD{0.62}{0.09} & \cellcolor{PDEFocus}\PDEMeanSD{0.09}{0.02} & \PDEMeanSD{0.66}{0.08} & \PDEMeanSD{0.28}{0.02} & \PDEMeanSD{0.53}{0.03} \\
\bottomrule
\end{tabular*}
\end{table}
\begin{table}[H]
\centering
\caption{Darcy / FNO. All 81 cells: mean $\pm$ SD on one line ($n=200$); blue marks matched views. All training views: 500/O500.}
\label{tab:matrix-darcy-fno}
\PDETableFont
\begin{tabular*}{\linewidth}{@{\extracolsep{\fill}}lrrrrrrrrr@{}}
\toprule
 & \multicolumn{9}{c}{Test observation pattern} \\
\cmidrule(l){2-10}
Train & R50 & H & V & CL & BL & LI & BD & R65 & R80 \\
\midrule
R50 & \cellcolor{PDEFocus}\PDEMeanSD{0.02}{0.01} & \PDEMeanSD{0.16}{0.07} & \PDEMeanSD{0.20}{0.09} & \PDEMeanSD{1.87}{0.50} & \PDEMeanSD{1.81}{0.33} & \PDEMeanSD{0.14}{0.05} & \PDEMeanSD{1.73}{0.29} & \PDEMeanSD{0.39}{0.04} & \PDEMeanSD{1.23}{0.16} \\
H & \PDEMeanSD{0.15}{0.07} & \cellcolor{PDEFocus}\PDEMeanSD{0.03}{0.02} & \PDEMeanSD{1.44}{0.81} & \PDEMeanSD{2.98}{0.98} & \PDEMeanSD{4.28}{0.80} & \PDEMeanSD{0.99}{0.61} & \PDEMeanSD{4.39}{0.81} & \PDEMeanSD{0.34}{0.06} & \PDEMeanSD{0.48}{0.09} \\
V & \PDEMeanSD{0.07}{0.02} & \PDEMeanSD{0.48}{0.23} & \cellcolor{PDEFocus}\PDEMeanSD{0.02}{0.01} & \PDEMeanSD{2.52}{0.61} & \PDEMeanSD{2.17}{0.41} & \PDEMeanSD{0.36}{0.19} & \PDEMeanSD{2.21}{0.40} & \PDEMeanSD{0.88}{0.12} & \PDEMeanSD{0.95}{0.16} \\
CL & \PDEMeanSD{0.20}{0.05} & \PDEMeanSD{0.20}{0.06} & \PDEMeanSD{0.19}{0.06} & \cellcolor{PDEFocus}\PDEMeanSD{0.14}{0.09} & \PDEMeanSD{0.26}{0.10} & \PDEMeanSD{0.20}{0.06} & \PDEMeanSD{0.35}{0.07} & \PDEMeanSD{0.26}{0.03} & \PDEMeanSD{0.35}{0.05} \\
BL & \PDEMeanSD{0.48}{0.12} & \PDEMeanSD{1.05}{0.45} & \PDEMeanSD{0.80}{0.30} & \PDEMeanSD{0.60}{0.16} & \cellcolor{PDEFocus}\PDEMeanSD{0.08}{0.04} & \PDEMeanSD{0.76}{0.31} & \PDEMeanSD{0.07}{0.04} & \PDEMeanSD{0.42}{0.12} & \PDEMeanSD{0.37}{0.09} \\
LI & \PDEMeanSD{0.02}{0.01} & \PDEMeanSD{0.03}{0.02} & \PDEMeanSD{0.03}{0.01} & \PDEMeanSD{0.59}{0.18} & \PDEMeanSD{0.56}{0.12} & \cellcolor{PDEFocus}\PDEMeanSD{0.03}{0.01} & \PDEMeanSD{0.61}{0.09} & \PDEMeanSD{0.16}{0.02} & \PDEMeanSD{0.22}{0.02} \\
\bottomrule
\end{tabular*}
\end{table}
\begin{table}[H]
\centering
\caption{Darcy / CNO. All 81 cells: mean $\pm$ SD on one line ($n=200$); blue marks matched views. All training views: 500/O500.}
\label{tab:matrix-darcy-cno}
\PDETableFont
\begin{tabular*}{\linewidth}{@{\extracolsep{\fill}}lrrrrrrrrr@{}}
\toprule
 & \multicolumn{9}{c}{Test observation pattern} \\
\cmidrule(l){2-10}
Train & R50 & H & V & CL & BL & LI & BD & R65 & R80 \\
\midrule
R50 & \cellcolor{PDEFocus}\PDEMeanSD{0.01}{0.00} & \PDEMeanSD{0.10}{0.05} & \PDEMeanSD{0.09}{0.04} & \PDEMeanSD{0.56}{0.15} & \PDEMeanSD{0.73}{0.11} & \PDEMeanSD{0.08}{0.04} & \PDEMeanSD{0.80}{0.06} & \PDEMeanSD{1.77}{0.76} & \PDEMeanSD{3.63}{0.45} \\
H & \PDEMeanSD{0.06}{0.01} & \cellcolor{PDEFocus}\PDEMeanSD{0.01}{0.01} & \PDEMeanSD{0.19}{0.04} & \PDEMeanSD{0.38}{0.19} & \PDEMeanSD{0.55}{0.13} & \PDEMeanSD{0.12}{0.03} & \PDEMeanSD{0.59}{0.07} & \PDEMeanSD{0.93}{0.09} & \PDEMeanSD{0.96}{0.13} \\
V & \PDEMeanSD{0.02}{0.00} & \PDEMeanSD{0.08}{0.03} & \cellcolor{PDEFocus}\PDEMeanSD{0.02}{0.01} & \PDEMeanSD{0.36}{0.15} & \PDEMeanSD{0.48}{0.10} & \PDEMeanSD{0.05}{0.01} & \PDEMeanSD{0.51}{0.07} & \PDEMeanSD{0.05}{0.01} & \PDEMeanSD{0.08}{0.01} \\
CL & \PDEMeanSD{0.07}{0.01} & \PDEMeanSD{0.07}{0.02} & \PDEMeanSD{0.08}{0.02} & \cellcolor{PDEFocus}\PDEMeanSD{0.11}{0.09} & \PDEMeanSD{0.29}{0.16} & \PDEMeanSD{0.07}{0.02} & \PDEMeanSD{0.36}{0.15} & \PDEMeanSD{0.08}{0.01} & \PDEMeanSD{0.10}{0.02} \\
BL & \PDEMeanSD{0.48}{0.11} & \PDEMeanSD{0.38}{0.08} & \PDEMeanSD{0.37}{0.08} & \PDEMeanSD{0.26}{0.10} & \cellcolor{PDEFocus}\PDEMeanSD{0.15}{0.09} & \PDEMeanSD{0.39}{0.08} & \PDEMeanSD{0.14}{0.09} & \PDEMeanSD{0.57}{0.14} & \PDEMeanSD{0.63}{0.16} \\
LI & \PDEMeanSD{0.01}{0.00} & \PDEMeanSD{0.01}{0.01} & \PDEMeanSD{0.01}{0.00} & \PDEMeanSD{0.38}{0.15} & \PDEMeanSD{0.57}{0.09} & \cellcolor{PDEFocus}\PDEMeanSD{0.01}{0.00} & \PDEMeanSD{0.61}{0.06} & \PDEMeanSD{0.82}{0.17} & \PDEMeanSD{2.03}{0.27} \\
\bottomrule
\end{tabular*}
\end{table}
\begin{table}[H]
\centering
\caption{Darcy / DeepONet. All 81 cells: mean $\pm$ SD on one line ($n=200$); blue marks matched views. Epoch/cohort in row order: 500/O500, 500/O500, 290/INT, 500/O500, 500/O500, 500/O500, 500/O500, 500/O500, 500/O500.}
\label{tab:matrix-darcy-deeponet}
\PDETableFont
\begin{tabular*}{\linewidth}{@{\extracolsep{\fill}}lrrrrrrrrr@{}}
\toprule
 & \multicolumn{9}{c}{Test observation pattern} \\
\cmidrule(l){2-10}
Train & R50 & H & V & CL & BL & LI & BD & R65 & R80 \\
\midrule
R50 & \cellcolor{PDEFocus}\PDEMeanSD{0.04}{0.02} & \PDEMeanSD{0.14}{0.04} & \PDEMeanSD{0.16}{0.04} & \PDEMeanSD{0.58}{0.11} & \PDEMeanSD{0.71}{0.08} & \PDEMeanSD{0.11}{0.03} & \PDEMeanSD{0.74}{0.04} & \PDEMeanSD{0.14}{0.01} & \PDEMeanSD{0.20}{0.02} \\
H & \PDEMeanSD{0.26}{0.06} & \cellcolor{PDEFocus}\PDEMeanSD{0.04}{0.02} & \PDEMeanSD{0.69}{0.24} & \PDEMeanSD{0.80}{0.26} & \PDEMeanSD{0.80}{0.17} & \PDEMeanSD{0.36}{0.13} & \PDEMeanSD{0.69}{0.12} & \PDEMeanSD{0.34}{0.07} & \PDEMeanSD{0.40}{0.07} \\
V & \PDEMeanSD{0.33}{0.02} & \PDEMeanSD{0.80}{0.32} & \cellcolor{PDEFocus}\PDEMeanSD{0.05}{0.02} & \PDEMeanSD{0.48}{0.15} & \PDEMeanSD{0.44}{0.14} & \PDEMeanSD{0.45}{0.20} & \PDEMeanSD{0.32}{0.10} & \PDEMeanSD{0.18}{0.02} & \PDEMeanSD{0.15}{0.03} \\
CL & \PDEMeanSD{0.24}{0.06} & \PDEMeanSD{2.88}{1.08} & \PDEMeanSD{1.59}{0.44} & \cellcolor{PDEFocus}\PDEMeanSD{0.14}{0.07} & \PDEMeanSD{0.29}{0.10} & \PDEMeanSD{0.58}{0.19} & \PDEMeanSD{0.31}{0.07} & \PDEMeanSD{0.22}{0.06} & \PDEMeanSD{0.23}{0.06} \\
BL & \PDEMeanSD{8.56}{1.10} & \PDEMeanSD{12.22}{2.47} & \PDEMeanSD{14.26}{2.39} & \PDEMeanSD{1.87}{0.77} & \cellcolor{PDEFocus}\PDEMeanSD{0.10}{0.06} & \PDEMeanSD{13.56}{2.00} & \PDEMeanSD{0.08}{0.04} & \PDEMeanSD{7.52}{0.96} & \PDEMeanSD{4.92}{0.68} \\
LI & \PDEMeanSD{0.53}{0.03} & \PDEMeanSD{0.16}{0.03} & \PDEMeanSD{0.17}{0.03} & \PDEMeanSD{0.42}{0.11} & \PDEMeanSD{0.43}{0.10} & \cellcolor{PDEFocus}\PDEMeanSD{0.04}{0.02} & \PDEMeanSD{0.43}{0.07} & \PDEMeanSD{0.41}{0.02} & \PDEMeanSD{0.25}{0.02} \\
\bottomrule
\end{tabular*}
\end{table}
\begin{table}[H]
\centering
\caption{Darcy / GNOT. All 81 cells: mean $\pm$ SD on one line ($n=200$); blue marks matched views. Epoch/cohort in row order: 500/O500, 185/INT, 182/INT, 184/INT, 500/O500, 500/O500, 185/INT, 500/O500, 500/O500.}
\label{tab:matrix-darcy-gnot}
\PDETableFont
\begin{tabular*}{\linewidth}{@{\extracolsep{\fill}}lrrrrrrrrr@{}}
\toprule
 & \multicolumn{9}{c}{Test observation pattern} \\
\cmidrule(l){2-10}
Train & R50 & H & V & CL & BL & LI & BD & R65 & R80 \\
\midrule
R50 & \cellcolor{PDEFocus}\PDEMeanSD{0.01}{0.01} & \PDEMeanSD{0.03}{0.01} & \PDEMeanSD{0.03}{0.01} & \PDEMeanSD{0.54}{1.57} & \PDEMeanSD{2.47}{7.11} & \PDEMeanSD{0.03}{0.01} & \PDEMeanSD{0.33}{0.24} & \PDEMeanSD{0.01}{0.01} & \PDEMeanSD{0.01}{0.01} \\
H & \PDEMeanSD{0.08}{0.01} & \cellcolor{PDEFocus}\PDEMeanSD{0.08}{0.01} & \PDEMeanSD{0.10}{0.02} & \PDEMeanSD{0.31}{0.20} & \PDEMeanSD{0.35}{0.12} & \PDEMeanSD{0.09}{0.02} & \PDEMeanSD{0.35}{0.09} & \PDEMeanSD{0.08}{0.01} & \PDEMeanSD{0.08}{0.01} \\
V & \PDEMeanSD{0.04}{0.01} & \PDEMeanSD{0.06}{0.02} & \cellcolor{PDEFocus}\PDEMeanSD{0.05}{0.01} & \PDEMeanSD{0.26}{0.18} & \PDEMeanSD{0.33}{0.12} & \PDEMeanSD{0.05}{0.02} & \PDEMeanSD{0.34}{0.07} & \PDEMeanSD{0.04}{0.01} & \PDEMeanSD{0.04}{0.01} \\
CL & \PDEMeanSD{0.06}{0.03} & \PDEMeanSD{0.06}{0.03} & \PDEMeanSD{0.06}{0.03} & \cellcolor{PDEFocus}\PDEMeanSD{0.10}{0.07} & \PDEMeanSD{0.23}{0.10} & \PDEMeanSD{0.06}{0.03} & \PDEMeanSD{0.24}{0.07} & \PDEMeanSD{0.06}{0.03} & \PDEMeanSD{0.06}{0.03} \\
BL & \PDEMeanSD{0.27}{0.07} & \PDEMeanSD{0.28}{0.07} & \PDEMeanSD{0.27}{0.07} & \PDEMeanSD{0.41}{0.17} & \cellcolor{PDEFocus}\PDEMeanSD{0.11}{0.08} & \PDEMeanSD{0.28}{0.07} & \PDEMeanSD{0.08}{0.07} & \PDEMeanSD{0.27}{0.07} & \PDEMeanSD{0.27}{0.07} \\
LI & \PDEMeanSD{0.01}{0.01} & \PDEMeanSD{0.02}{0.01} & \PDEMeanSD{0.02}{0.01} & \PDEMeanSD{0.84}{4.49} & \PDEMeanSD{0.74}{0.83} & \cellcolor{PDEFocus}\PDEMeanSD{0.02}{0.01} & \PDEMeanSD{0.40}{0.11} & \PDEMeanSD{0.01}{0.01} & \PDEMeanSD{0.01}{0.01} \\
\bottomrule
\end{tabular*}
\end{table}
\begin{table}[H]
\centering
\caption{Darcy / Transolver. All 81 cells: mean $\pm$ SD on one line ($n=200$); blue marks matched views. Epoch/cohort in row order: 500/O500, 500/O500, 500/O500, 500/O500, 500/O500, 500/O500, 200/M120, 500/O500, 500/O500.}
\label{tab:matrix-darcy-transolver}
\PDETableFont
\begin{tabular*}{\linewidth}{@{\extracolsep{\fill}}lrrrrrrrrr@{}}
\toprule
 & \multicolumn{9}{c}{Test observation pattern} \\
\cmidrule(l){2-10}
Train & R50 & H & V & CL & BL & LI & BD & R65 & R80 \\
\midrule
R50 & \cellcolor{PDEFocus}\PDEMeanSD{0.00}{0.00} & \PDEMeanSD{0.52}{0.16} & \PDEMeanSD{0.60}{0.20} & \PDEMeanSD{1.31}{0.36} & \PDEMeanSD{1.53}{0.31} & \PDEMeanSD{0.53}{0.11} & \PDEMeanSD{1.55}{0.26} & \PDEMeanSD{0.05}{0.01} & \PDEMeanSD{0.13}{0.02} \\
H & \PDEMeanSD{0.19}{0.03} & \cellcolor{PDEFocus}\PDEMeanSD{0.00}{0.01} & \PDEMeanSD{0.29}{0.04} & \PDEMeanSD{0.33}{0.13} & \PDEMeanSD{0.43}{0.10} & \PDEMeanSD{0.10}{0.02} & \PDEMeanSD{0.46}{0.08} & \PDEMeanSD{0.17}{0.03} & \PDEMeanSD{0.15}{0.03} \\
V & \PDEMeanSD{0.40}{0.04} & \PDEMeanSD{0.46}{0.05} & \cellcolor{PDEFocus}\PDEMeanSD{0.01}{0.00} & \PDEMeanSD{0.64}{0.19} & \PDEMeanSD{0.73}{0.15} & \PDEMeanSD{0.31}{0.04} & \PDEMeanSD{0.80}{0.10} & \PDEMeanSD{0.34}{0.04} & \PDEMeanSD{0.27}{0.03} \\
CL & \PDEMeanSD{0.02}{0.00} & \PDEMeanSD{0.04}{0.02} & \PDEMeanSD{0.04}{0.02} & \cellcolor{PDEFocus}\PDEMeanSD{0.06}{0.06} & \PDEMeanSD{0.19}{0.10} & \PDEMeanSD{0.04}{0.01} & \PDEMeanSD{0.18}{0.09} & \PDEMeanSD{0.04}{0.01} & \PDEMeanSD{0.06}{0.01} \\
BL & \PDEMeanSD{0.69}{0.17} & \PDEMeanSD{0.82}{0.21} & \PDEMeanSD{0.66}{0.18} & \PDEMeanSD{0.54}{0.24} & \cellcolor{PDEFocus}\PDEMeanSD{0.08}{0.05} & \PDEMeanSD{0.57}{0.16} & \PDEMeanSD{0.06}{0.04} & \PDEMeanSD{0.61}{0.15} & \PDEMeanSD{0.55}{0.13} \\
LI & \PDEMeanSD{0.09}{0.02} & \PDEMeanSD{0.01}{0.01} & \PDEMeanSD{0.01}{0.01} & \PDEMeanSD{0.37}{0.16} & \PDEMeanSD{0.70}{0.10} & \cellcolor{PDEFocus}\PDEMeanSD{0.00}{0.00} & \PDEMeanSD{0.75}{0.09} & \PDEMeanSD{0.07}{0.01} & \PDEMeanSD{0.15}{0.03} \\
\bottomrule
\end{tabular*}
\end{table}
\begin{table}[H]
\centering
\caption{Darcy / PINO. All 81 cells: mean $\pm$ SD on one line ($n=200$); blue marks matched views. All training views: 500/O500.}
\label{tab:matrix-darcy-pino}
\PDETableFont
\begin{tabular*}{\linewidth}{@{\extracolsep{\fill}}lrrrrrrrrr@{}}
\toprule
 & \multicolumn{9}{c}{Test observation pattern} \\
\cmidrule(l){2-10}
Train & R50 & H & V & CL & BL & LI & BD & R65 & R80 \\
\midrule
R50 & \cellcolor{PDEFocus}\PDEMeanSD{0.02}{0.01} & \PDEMeanSD{0.08}{0.06} & \PDEMeanSD{0.08}{0.05} & \PDEMeanSD{0.92}{0.36} & \PDEMeanSD{1.34}{0.24} & \PDEMeanSD{0.07}{0.05} & \PDEMeanSD{1.50}{0.22} & \PDEMeanSD{0.29}{0.01} & \PDEMeanSD{0.57}{0.03} \\
H & \PDEMeanSD{0.10}{0.02} & \cellcolor{PDEFocus}\PDEMeanSD{0.02}{0.01} & \PDEMeanSD{0.24}{0.07} & \PDEMeanSD{1.05}{0.28} & \PDEMeanSD{1.09}{0.15} & \PDEMeanSD{0.17}{0.06} & \PDEMeanSD{1.10}{0.08} & \PDEMeanSD{0.29}{0.02} & \PDEMeanSD{0.53}{0.04} \\
V & \PDEMeanSD{0.50}{0.12} & \PDEMeanSD{0.25}{0.08} & \cellcolor{PDEFocus}\PDEMeanSD{0.02}{0.01} & \PDEMeanSD{37.80}{27.49} & \PDEMeanSD{2.20}{1.19} & \PDEMeanSD{0.27}{0.11} & \PDEMeanSD{2.73}{0.67} & \PDEMeanSD{0.53}{0.11} & \PDEMeanSD{0.63}{0.09} \\
CL & \PDEMeanSD{0.18}{0.08} & \PDEMeanSD{0.15}{0.07} & \PDEMeanSD{0.15}{0.07} & \cellcolor{PDEFocus}\PDEMeanSD{0.14}{0.09} & \PDEMeanSD{0.29}{0.10} & \PDEMeanSD{0.15}{0.07} & \PDEMeanSD{0.34}{0.08} & \PDEMeanSD{0.13}{0.05} & \PDEMeanSD{0.16}{0.05} \\
BL & \PDEMeanSD{0.38}{0.07} & \PDEMeanSD{0.30}{0.09} & \PDEMeanSD{0.31}{0.10} & \PDEMeanSD{0.40}{0.12} & \cellcolor{PDEFocus}\PDEMeanSD{0.04}{0.04} & \PDEMeanSD{0.29}{0.08} & \PDEMeanSD{0.03}{0.03} & \PDEMeanSD{0.31}{0.07} & \PDEMeanSD{0.24}{0.06} \\
LI & \PDEMeanSD{0.04}{0.01} & \PDEMeanSD{0.02}{0.02} & \PDEMeanSD{0.02}{0.01} & \PDEMeanSD{0.45}{0.13} & \PDEMeanSD{0.44}{0.11} & \cellcolor{PDEFocus}\PDEMeanSD{0.02}{0.01} & \PDEMeanSD{0.49}{0.08} & \PDEMeanSD{0.26}{0.02} & \PDEMeanSD{0.53}{0.05} \\
\bottomrule
\end{tabular*}
\end{table}
\begin{table}[H]
\centering
\caption{Poisson / U-FNO. All 81 cells: mean $\pm$ SD on one line ($n=200$); blue marks matched views. Epoch/cohort in row order: 130/M120, 130/M120, 137/M120, 200/M120, 200/M120, 134/M120, 200/M120, 151/M120, 163/M120.}
\label{tab:matrix-poisson-ufno-2d}
\PDETableFont
\begin{tabular*}{\linewidth}{@{\extracolsep{\fill}}lrrrrrrrrr@{}}
\toprule
 & \multicolumn{9}{c}{Test observation pattern} \\
\cmidrule(l){2-10}
Train & R50 & H & V & CL & BL & LI & BD & R65 & R80 \\
\midrule
R50 & \cellcolor{PDEFocus}\PDEMeanSD{0.18}{0.11} & \PDEMeanSD{0.20}{0.13} & \PDEMeanSD{0.21}{0.13} & \PDEMeanSD{2.69}{1.76} & \PDEMeanSD{3.28}{2.08} & \PDEMeanSD{0.19}{0.11} & \PDEMeanSD{3.11}{1.98} & \PDEMeanSD{0.12}{0.06} & \PDEMeanSD{1.00}{0.64} \\
H & \PDEMeanSD{0.50}{0.14} & \cellcolor{PDEFocus}\PDEMeanSD{0.08}{0.05} & \PDEMeanSD{0.57}{0.11} & \PDEMeanSD{0.49}{0.15} & \PDEMeanSD{0.73}{0.21} & \PDEMeanSD{0.24}{0.06} & \PDEMeanSD{0.77}{0.20} & \PDEMeanSD{0.46}{0.14} & \PDEMeanSD{0.42}{0.12} \\
V & \PDEMeanSD{0.62}{0.24} & \PDEMeanSD{0.61}{0.24} & \cellcolor{PDEFocus}\PDEMeanSD{0.06}{0.03} & \PDEMeanSD{0.74}{0.28} & \PDEMeanSD{0.90}{0.36} & \PDEMeanSD{0.35}{0.13} & \PDEMeanSD{0.93}{0.35} & \PDEMeanSD{0.60}{0.24} & \PDEMeanSD{0.62}{0.24} \\
CL & \PDEMeanSD{0.06}{0.03} & \PDEMeanSD{0.07}{0.02} & \PDEMeanSD{0.07}{0.03} & \cellcolor{PDEFocus}\PDEMeanSD{0.05}{0.04} & \PDEMeanSD{0.38}{0.15} & \PDEMeanSD{0.05}{0.02} & \PDEMeanSD{0.39}{0.14} & \PDEMeanSD{0.05}{0.02} & \PDEMeanSD{0.04}{0.01} \\
BL & \PDEMeanSD{0.31}{0.05} & \PDEMeanSD{0.33}{0.10} & \PDEMeanSD{0.34}{0.09} & \PDEMeanSD{0.36}{0.18} & \cellcolor{PDEFocus}\PDEMeanSD{0.07}{0.04} & \PDEMeanSD{0.32}{0.07} & \PDEMeanSD{0.05}{0.02} & \PDEMeanSD{0.25}{0.05} & \PDEMeanSD{0.34}{0.18} \\
LI & \PDEMeanSD{0.20}{0.07} & \PDEMeanSD{0.18}{0.08} & \PDEMeanSD{0.17}{0.08} & \PDEMeanSD{0.77}{0.30} & \PDEMeanSD{1.01}{0.44} & \cellcolor{PDEFocus}\PDEMeanSD{0.16}{0.07} & \PDEMeanSD{1.05}{0.45} & \PDEMeanSD{0.26}{0.12} & \PDEMeanSD{0.48}{0.24} \\
\bottomrule
\end{tabular*}
\end{table}
\begin{table}[H]
\centering
\caption{Poisson / FNO. All 81 cells: mean $\pm$ SD on one line ($n=200$); blue marks matched views. Epoch/cohort in row order: 433/INT, 500/O500, 500/O500, 262/INT, 130/M120, 480/INT, 130/M120, 500/O500, 461/INT.}
\label{tab:matrix-poisson-fno}
\PDETableFont
\begin{tabular*}{\linewidth}{@{\extracolsep{\fill}}lrrrrrrrrr@{}}
\toprule
 & \multicolumn{9}{c}{Test observation pattern} \\
\cmidrule(l){2-10}
Train & R50 & H & V & CL & BL & LI & BD & R65 & R80 \\
\midrule
R50 & \cellcolor{PDEFocus}\PDEMeanSD{0.01}{0.00} & \PDEMeanSD{0.05}{0.03} & \PDEMeanSD{0.05}{0.03} & \PDEMeanSD{0.47}{0.19} & \PDEMeanSD{0.81}{0.24} & \PDEMeanSD{0.04}{0.02} & \PDEMeanSD{0.88}{0.24} & \PDEMeanSD{0.15}{0.04} & \PDEMeanSD{0.29}{0.08} \\
H & \PDEMeanSD{0.06}{0.03} & \cellcolor{PDEFocus}\PDEMeanSD{0.02}{0.01} & \PDEMeanSD{0.36}{0.23} & \PDEMeanSD{2.14}{1.37} & \PDEMeanSD{2.99}{1.92} & \PDEMeanSD{0.25}{0.16} & \PDEMeanSD{3.01}{1.94} & \PDEMeanSD{0.19}{0.10} & \PDEMeanSD{0.51}{0.23} \\
V & \PDEMeanSD{0.03}{0.01} & \PDEMeanSD{0.18}{0.09} & \cellcolor{PDEFocus}\PDEMeanSD{0.02}{0.01} & \PDEMeanSD{0.65}{0.31} & \PDEMeanSD{1.21}{0.56} & \PDEMeanSD{0.12}{0.06} & \PDEMeanSD{1.26}{0.58} & \PDEMeanSD{0.11}{0.05} & \PDEMeanSD{0.25}{0.13} \\
CL & \PDEMeanSD{0.08}{0.03} & \PDEMeanSD{0.08}{0.03} & \PDEMeanSD{0.09}{0.03} & \cellcolor{PDEFocus}\PDEMeanSD{0.08}{0.04} & \PDEMeanSD{0.47}{0.21} & \PDEMeanSD{0.08}{0.03} & \PDEMeanSD{0.54}{0.18} & \PDEMeanSD{0.07}{0.04} & \PDEMeanSD{0.11}{0.04} \\
BL & \PDEMeanSD{0.67}{0.17} & \PDEMeanSD{0.86}{0.41} & \PDEMeanSD{0.98}{0.58} & \PDEMeanSD{0.71}{0.33} & \cellcolor{PDEFocus}\PDEMeanSD{0.14}{0.09} & \PDEMeanSD{0.86}{0.40} & \PDEMeanSD{0.10}{0.05} & \PDEMeanSD{0.34}{0.09} & \PDEMeanSD{0.29}{0.08} \\
LI & \PDEMeanSD{0.02}{0.00} & \PDEMeanSD{0.02}{0.01} & \PDEMeanSD{0.02}{0.01} & \PDEMeanSD{0.38}{0.14} & \PDEMeanSD{0.73}{0.15} & \cellcolor{PDEFocus}\PDEMeanSD{0.01}{0.00} & \PDEMeanSD{0.77}{0.14} & \PDEMeanSD{0.21}{0.14} & \PDEMeanSD{0.53}{0.33} \\
\bottomrule
\end{tabular*}
\end{table}
\begin{table}[H]
\centering
\caption{Poisson / CNO. All 81 cells: mean $\pm$ SD on one line ($n=200$); blue marks matched views. Epoch/cohort in row order: 275/INT, 500/O500, 500/O500, 500/O500, 500/O500, 500/O500, 500/O500, 500/O500, 500/O500.}
\label{tab:matrix-poisson-cno}
\PDETableFont
\begin{tabular*}{\linewidth}{@{\extracolsep{\fill}}lrrrrrrrrr@{}}
\toprule
 & \multicolumn{9}{c}{Test observation pattern} \\
\cmidrule(l){2-10}
Train & R50 & H & V & CL & BL & LI & BD & R65 & R80 \\
\midrule
R50 & \cellcolor{PDEFocus}\PDEMeanSD{0.02}{0.01} & \PDEMeanSD{0.13}{0.09} & \PDEMeanSD{0.13}{0.09} & \PDEMeanSD{1.14}{0.53} & \PDEMeanSD{1.53}{0.64} & \PDEMeanSD{0.12}{0.06} & \PDEMeanSD{1.53}{0.63} & \PDEMeanSD{0.30}{0.07} & \PDEMeanSD{1.35}{0.41} \\
H & \PDEMeanSD{0.02}{0.01} & \cellcolor{PDEFocus}\PDEMeanSD{0.01}{0.00} & \PDEMeanSD{0.06}{0.02} & \PDEMeanSD{0.43}{0.21} & \PDEMeanSD{0.87}{0.31} & \PDEMeanSD{0.04}{0.01} & \PDEMeanSD{0.92}{0.31} & \PDEMeanSD{0.15}{0.03} & \PDEMeanSD{0.34}{0.08} \\
V & \PDEMeanSD{0.01}{0.01} & \PDEMeanSD{0.05}{0.02} & \cellcolor{PDEFocus}\PDEMeanSD{0.01}{0.00} & \PDEMeanSD{0.39}{0.17} & \PDEMeanSD{0.80}{0.14} & \PDEMeanSD{0.03}{0.01} & \PDEMeanSD{0.81}{0.10} & \PDEMeanSD{0.12}{0.01} & \PDEMeanSD{0.25}{0.03} \\
CL & \PDEMeanSD{0.09}{0.04} & \PDEMeanSD{0.08}{0.02} & \PDEMeanSD{0.07}{0.02} & \cellcolor{PDEFocus}\PDEMeanSD{0.12}{0.07} & \PDEMeanSD{0.74}{0.26} & \PDEMeanSD{0.08}{0.02} & \PDEMeanSD{0.83}{0.27} & \PDEMeanSD{0.14}{0.05} & \PDEMeanSD{0.20}{0.06} \\
BL & \PDEMeanSD{0.64}{0.35} & \PDEMeanSD{0.68}{0.31} & \PDEMeanSD{0.65}{0.29} & \PDEMeanSD{0.81}{0.26} & \cellcolor{PDEFocus}\PDEMeanSD{0.12}{0.07} & \PDEMeanSD{0.62}{0.31} & \PDEMeanSD{0.09}{0.04} & \PDEMeanSD{0.65}{0.40} & \PDEMeanSD{0.66}{0.41} \\
LI & \PDEMeanSD{0.01}{0.01} & \PDEMeanSD{0.01}{0.01} & \PDEMeanSD{0.01}{0.01} & \PDEMeanSD{0.50}{0.25} & \PDEMeanSD{1.06}{0.40} & \cellcolor{PDEFocus}\PDEMeanSD{0.01}{0.01} & \PDEMeanSD{1.08}{0.39} & \PDEMeanSD{0.06}{0.03} & \PDEMeanSD{0.10}{0.05} \\
\bottomrule
\end{tabular*}
\end{table}
\begin{table}[H]
\centering
\caption{Poisson / DeepONet. All 81 cells: mean $\pm$ SD on one line ($n=200$); blue marks matched views. Epoch/cohort in row order: 500/O500, 500/O500, 500/O500, 66/INT, 500/O500, 500/O500, 500/O500, 500/O500, 500/O500.}
\label{tab:matrix-poisson-deeponet}
\PDETableFont
\begin{tabular*}{\linewidth}{@{\extracolsep{\fill}}lrrrrrrrrr@{}}
\toprule
 & \multicolumn{9}{c}{Test observation pattern} \\
\cmidrule(l){2-10}
Train & R50 & H & V & CL & BL & LI & BD & R65 & R80 \\
\midrule
R50 & \cellcolor{PDEFocus}\PDEMeanSD{0.02}{0.01} & \PDEMeanSD{0.16}{0.04} & \PDEMeanSD{0.15}{0.05} & \PDEMeanSD{0.59}{0.11} & \PDEMeanSD{0.89}{0.09} & \PDEMeanSD{0.13}{0.03} & \PDEMeanSD{0.91}{0.09} & \PDEMeanSD{0.06}{0.02} & \PDEMeanSD{0.19}{0.04} \\
H & \PDEMeanSD{0.19}{0.08} & \cellcolor{PDEFocus}\PDEMeanSD{0.03}{0.01} & \PDEMeanSD{0.29}{0.14} & \PDEMeanSD{0.65}{0.27} & \PDEMeanSD{0.83}{0.17} & \PDEMeanSD{0.16}{0.11} & \PDEMeanSD{0.81}{0.15} & \PDEMeanSD{0.10}{0.05} & \PDEMeanSD{0.09}{0.03} \\
V & \PDEMeanSD{0.17}{0.06} & \PDEMeanSD{0.24}{0.09} & \cellcolor{PDEFocus}\PDEMeanSD{0.03}{0.01} & \PDEMeanSD{0.54}{0.19} & \PDEMeanSD{0.78}{0.16} & \PDEMeanSD{0.11}{0.07} & \PDEMeanSD{0.76}{0.16} & \PDEMeanSD{0.12}{0.06} & \PDEMeanSD{0.10}{0.06} \\
CL & \PDEMeanSD{0.60}{0.24} & \PDEMeanSD{1.44}{0.55} & \PDEMeanSD{0.29}{0.12} & \cellcolor{PDEFocus}\PDEMeanSD{0.23}{0.09} & \PDEMeanSD{0.61}{0.24} & \PDEMeanSD{0.94}{0.36} & \PDEMeanSD{0.65}{0.24} & \PDEMeanSD{0.52}{0.20} & \PDEMeanSD{0.38}{0.16} \\
BL & \PDEMeanSD{7.26}{2.82} & \PDEMeanSD{13.65}{7.26} & \PDEMeanSD{11.65}{5.08} & \PDEMeanSD{1.62}{0.95} & \cellcolor{PDEFocus}\PDEMeanSD{0.13}{0.08} & \PDEMeanSD{11.44}{5.05} & \PDEMeanSD{0.07}{0.03} & \PDEMeanSD{6.52}{2.16} & \PDEMeanSD{4.46}{1.19} \\
LI & \PDEMeanSD{0.34}{0.06} & \PDEMeanSD{0.11}{0.03} & \PDEMeanSD{0.13}{0.03} & \PDEMeanSD{0.47}{0.15} & \PDEMeanSD{0.77}{0.14} & \cellcolor{PDEFocus}\PDEMeanSD{0.03}{0.01} & \PDEMeanSD{0.77}{0.13} & \PDEMeanSD{0.26}{0.05} & \PDEMeanSD{0.16}{0.04} \\
\bottomrule
\end{tabular*}
\end{table}
\begin{table}[H]
\centering
\caption{Poisson / GNOT. All 81 cells: mean $\pm$ SD on one line ($n=200$); blue marks matched views. Epoch/cohort in row order: 500/O500, 500/O500, 200/P10, 500/O500, 200/M120, 500/O500, 200/M120, 200/M120, 135/INT.}
\label{tab:matrix-poisson-gnot}
\PDETableFont
\begin{tabular*}{\linewidth}{@{\extracolsep{\fill}}lrrrrrrrrr@{}}
\toprule
 & \multicolumn{9}{c}{Test observation pattern} \\
\cmidrule(l){2-10}
Train & R50 & H & V & CL & BL & LI & BD & R65 & R80 \\
\midrule
R50 & \cellcolor{PDEFocus}\PDEMeanSD{0.01}{0.00} & \PDEMeanSD{0.02}{0.01} & \PDEMeanSD{0.02}{0.01} & \PDEMeanSD{15.43}{33.93} & \PDEMeanSD{71.17}{92.13} & \PDEMeanSD{0.02}{0.01} & \PDEMeanSD{85.57}{55.23} & \PDEMeanSD{0.01}{0.00} & \PDEMeanSD{0.01}{0.00} \\
H & \PDEMeanSD{0.01}{0.00} & \cellcolor{PDEFocus}\PDEMeanSD{0.01}{0.00} & \PDEMeanSD{0.05}{0.04} & \PDEMeanSD{6.17}{17.29} & \PDEMeanSD{26.94}{23.73} & \PDEMeanSD{0.02}{0.01} & \PDEMeanSD{33.87}{19.62} & \PDEMeanSD{0.01}{0.00} & \PDEMeanSD{0.01}{0.00} \\
V & \PDEMeanSD{0.02}{0.01} & \PDEMeanSD{0.08}{0.04} & \cellcolor{PDEFocus}\PDEMeanSD{0.01}{0.01} & \PDEMeanSD{6.43}{9.06} & \PDEMeanSD{4.12}{7.29} & \PDEMeanSD{0.06}{0.03} & \PDEMeanSD{0.92}{0.38} & \PDEMeanSD{0.01}{0.00} & \PDEMeanSD{0.01}{0.00} \\
CL & \PDEMeanSD{0.02}{0.01} & \PDEMeanSD{0.02}{0.01} & \PDEMeanSD{0.02}{0.01} & \cellcolor{PDEFocus}\PDEMeanSD{0.04}{0.03} & \PDEMeanSD{0.38}{0.16} & \PDEMeanSD{0.02}{0.01} & \PDEMeanSD{0.34}{0.12} & \PDEMeanSD{0.02}{0.01} & \PDEMeanSD{0.02}{0.01} \\
BL & \PDEMeanSD{0.27}{0.14} & \PDEMeanSD{0.30}{0.15} & \PDEMeanSD{0.29}{0.15} & \PDEMeanSD{0.60}{0.27} & \cellcolor{PDEFocus}\PDEMeanSD{0.29}{0.16} & \PDEMeanSD{0.29}{0.16} & \PDEMeanSD{0.26}{0.16} & \PDEMeanSD{0.27}{0.14} & \PDEMeanSD{0.27}{0.15} \\
LI & \PDEMeanSD{0.01}{0.00} & \PDEMeanSD{0.01}{0.00} & \PDEMeanSD{0.01}{0.00} & \PDEMeanSD{6.39}{13.21} & \PDEMeanSD{20.64}{25.29} & \cellcolor{PDEFocus}\PDEMeanSD{0.01}{0.00} & \PDEMeanSD{8.91}{5.64} & \PDEMeanSD{0.01}{0.00} & \PDEMeanSD{0.01}{0.00} \\
\bottomrule
\end{tabular*}
\end{table}
\begin{table}[H]
\centering
\caption{Poisson / Transolver. All 81 cells: mean $\pm$ SD on one line ($n=200$); blue marks matched views. Epoch/cohort in row order: 500/O500, 500/O500, 189/INT, 500/O500, 500/O500, 191/INT, 500/O500, 500/O500, 500/O500.}
\label{tab:matrix-poisson-transolver}
\PDETableFont
\begin{tabular*}{\linewidth}{@{\extracolsep{\fill}}lrrrrrrrrr@{}}
\toprule
 & \multicolumn{9}{c}{Test observation pattern} \\
\cmidrule(l){2-10}
Train & R50 & H & V & CL & BL & LI & BD & R65 & R80 \\
\midrule
R50 & \cellcolor{PDEFocus}\PDEMeanSD{0.00}{0.00} & \PDEMeanSD{0.29}{0.11} & \PDEMeanSD{0.36}{0.17} & \PDEMeanSD{0.66}{0.22} & \PDEMeanSD{1.98}{1.31} & \PDEMeanSD{0.29}{0.11} & \PDEMeanSD{2.17}{1.37} & \PDEMeanSD{0.08}{0.02} & \PDEMeanSD{0.31}{0.07} \\
H & \PDEMeanSD{1.74}{0.86} & \cellcolor{PDEFocus}\PDEMeanSD{0.00}{0.01} & \PDEMeanSD{0.73}{0.14} & \PDEMeanSD{0.62}{0.20} & \PDEMeanSD{1.00}{0.43} & \PDEMeanSD{0.32}{0.08} & \PDEMeanSD{1.02}{0.33} & \PDEMeanSD{1.88}{0.96} & \PDEMeanSD{1.44}{0.87} \\
V & \PDEMeanSD{0.50}{0.08} & \PDEMeanSD{0.65}{0.12} & \cellcolor{PDEFocus}\PDEMeanSD{0.06}{0.03} & \PDEMeanSD{0.83}{0.36} & \PDEMeanSD{1.05}{0.39} & \PDEMeanSD{0.34}{0.07} & \PDEMeanSD{1.09}{0.38} & \PDEMeanSD{0.40}{0.08} & \PDEMeanSD{0.35}{0.09} \\
CL & \PDEMeanSD{0.07}{0.04} & \PDEMeanSD{0.13}{0.06} & \PDEMeanSD{0.12}{0.05} & \cellcolor{PDEFocus}\PDEMeanSD{0.07}{0.04} & \PDEMeanSD{0.69}{0.19} & \PDEMeanSD{0.08}{0.02} & \PDEMeanSD{0.68}{0.15} & \PDEMeanSD{0.08}{0.02} & \PDEMeanSD{0.11}{0.04} \\
BL & \PDEMeanSD{1.25}{0.59} & \PDEMeanSD{1.04}{0.43} & \PDEMeanSD{0.99}{0.39} & \PDEMeanSD{0.93}{0.47} & \cellcolor{PDEFocus}\PDEMeanSD{0.15}{0.09} & \PDEMeanSD{0.99}{0.37} & \PDEMeanSD{0.10}{0.06} & \PDEMeanSD{1.32}{0.62} & \PDEMeanSD{1.35}{0.64} \\
LI & \PDEMeanSD{0.15}{0.04} & \PDEMeanSD{0.05}{0.02} & \PDEMeanSD{0.05}{0.03} & \PDEMeanSD{0.55}{0.21} & \PDEMeanSD{0.80}{0.14} & \cellcolor{PDEFocus}\PDEMeanSD{0.05}{0.02} & \PDEMeanSD{0.83}{0.10} & \PDEMeanSD{0.17}{0.07} & \PDEMeanSD{0.21}{0.08} \\
\bottomrule
\end{tabular*}
\end{table}
\begin{table}[H]
\centering
\caption{Poisson / PINO. All 81 cells: mean $\pm$ SD on one line ($n=200$); blue marks matched views. Epoch/cohort in row order: 500/O500, 500/O500, 500/O500, 130/M120, 500/O500, 500/O500, 500/O500, 500/O500, 500/O500.}
\label{tab:matrix-poisson-pino}
\PDETableFont
\begin{tabular*}{\linewidth}{@{\extracolsep{\fill}}lrrrrrrrrr@{}}
\toprule
 & \multicolumn{9}{c}{Test observation pattern} \\
\cmidrule(l){2-10}
Train & R50 & H & V & CL & BL & LI & BD & R65 & R80 \\
\midrule
R50 & \cellcolor{PDEFocus}\PDEMeanSD{0.04}{0.01} & \PDEMeanSD{0.16}{0.07} & \PDEMeanSD{0.15}{0.06} & \PDEMeanSD{0.66}{0.16} & \PDEMeanSD{0.95}{0.20} & \PDEMeanSD{0.14}{0.05} & \PDEMeanSD{1.05}{0.18} & \PDEMeanSD{0.28}{0.05} & \PDEMeanSD{0.55}{0.14} \\
H & \PDEMeanSD{0.14}{0.08} & \cellcolor{PDEFocus}\PDEMeanSD{0.01}{0.01} & \PDEMeanSD{0.23}{0.08} & \PDEMeanSD{1.05}{0.57} & \PDEMeanSD{1.73}{1.19} & \PDEMeanSD{0.16}{0.07} & \PDEMeanSD{2.25}{1.45} & \PDEMeanSD{0.30}{0.07} & \PDEMeanSD{0.52}{0.10} \\
V & \PDEMeanSD{0.14}{0.08} & \PDEMeanSD{0.25}{0.12} & \cellcolor{PDEFocus}\PDEMeanSD{0.02}{0.01} & \PDEMeanSD{1.31}{0.78} & \PDEMeanSD{1.68}{0.83} & \PDEMeanSD{0.16}{0.08} & \PDEMeanSD{2.02}{1.00} & \PDEMeanSD{0.29}{0.08} & \PDEMeanSD{0.52}{0.12} \\
CL & \PDEMeanSD{0.38}{0.03} & \PDEMeanSD{0.37}{0.05} & \PDEMeanSD{0.38}{0.06} & \cellcolor{PDEFocus}\PDEMeanSD{0.13}{0.08} & \PDEMeanSD{0.45}{0.16} & \PDEMeanSD{0.37}{0.04} & \PDEMeanSD{0.45}{0.18} & \PDEMeanSD{0.24}{0.03} & \PDEMeanSD{0.11}{0.03} \\
BL & \PDEMeanSD{0.76}{0.08} & \PDEMeanSD{0.63}{0.20} & \PDEMeanSD{0.83}{0.23} & \PDEMeanSD{0.83}{0.21} & \cellcolor{PDEFocus}\PDEMeanSD{0.07}{0.05} & \PDEMeanSD{0.61}{0.19} & \PDEMeanSD{0.05}{0.03} & \PDEMeanSD{0.69}{0.09} & \PDEMeanSD{0.64}{0.10} \\
LI & \PDEMeanSD{0.05}{0.02} & \PDEMeanSD{0.04}{0.02} & \PDEMeanSD{0.04}{0.02} & \PDEMeanSD{0.52}{0.16} & \PDEMeanSD{0.77}{0.19} & \cellcolor{PDEFocus}\PDEMeanSD{0.03}{0.02} & \PDEMeanSD{0.84}{0.19} & \PDEMeanSD{0.24}{0.05} & \PDEMeanSD{0.49}{0.12} \\
\bottomrule
\end{tabular*}
\end{table}
\begin{table}[H]
\centering
\caption{Helmholtz / U-FNO. All 81 cells: mean $\pm$ SD on one line ($n=200$); blue marks matched views. Epoch/cohort in row order: 132/M120, 133/M120, 135/M120, 138/M120, 200/M120, 134/M120, 200/M120, 155/M120, 500/O500.}
\label{tab:matrix-helmholtz-ufno-2d}
\PDETableFont
\begin{tabular*}{\linewidth}{@{\extracolsep{\fill}}lrrrrrrrrr@{}}
\toprule
 & \multicolumn{9}{c}{Test observation pattern} \\
\cmidrule(l){2-10}
Train & R50 & H & V & CL & BL & LI & BD & R65 & R80 \\
\midrule
R50 & \cellcolor{PDEFocus}\PDEMeanSD{1.07}{0.29} & \PDEMeanSD{1.01}{0.27} & \PDEMeanSD{1.02}{0.28} & \PDEMeanSD{1.14}{0.37} & \PDEMeanSD{1.37}{0.51} & \PDEMeanSD{1.03}{0.28} & \PDEMeanSD{1.34}{0.51} & \PDEMeanSD{0.95}{0.26} & \PDEMeanSD{0.45}{0.11} \\
H & \PDEMeanSD{0.47}{0.01} & \cellcolor{PDEFocus}\PDEMeanSD{0.02}{0.01} & \PDEMeanSD{0.68}{0.06} & \PDEMeanSD{0.52}{0.16} & \PDEMeanSD{0.76}{0.23} & \PDEMeanSD{0.26}{0.03} & \PDEMeanSD{0.80}{0.20} & \PDEMeanSD{0.44}{0.01} & \PDEMeanSD{0.40}{0.02} \\
V & \PDEMeanSD{0.47}{0.02} & \PDEMeanSD{0.67}{0.06} & \cellcolor{PDEFocus}\PDEMeanSD{0.03}{0.01} & \PDEMeanSD{0.60}{0.18} & \PDEMeanSD{0.78}{0.27} & \PDEMeanSD{0.28}{0.05} & \PDEMeanSD{0.81}{0.25} & \PDEMeanSD{0.46}{0.02} & \PDEMeanSD{0.47}{0.03} \\
CL & \PDEMeanSD{0.04}{0.02} & \PDEMeanSD{0.09}{0.03} & \PDEMeanSD{0.12}{0.07} & \cellcolor{PDEFocus}\PDEMeanSD{0.07}{0.03} & \PDEMeanSD{0.38}{0.17} & \PDEMeanSD{0.08}{0.03} & \PDEMeanSD{0.33}{0.13} & \PDEMeanSD{0.06}{0.02} & \PDEMeanSD{0.08}{0.03} \\
BL & \PDEMeanSD{0.63}{0.43} & \PDEMeanSD{1.36}{0.47} & \PDEMeanSD{1.42}{0.48} & \PDEMeanSD{0.29}{0.10} & \cellcolor{PDEFocus}\PDEMeanSD{0.06}{0.03} & \PDEMeanSD{1.24}{0.52} & \PDEMeanSD{0.05}{0.02} & \PDEMeanSD{0.52}{0.37} & \PDEMeanSD{0.27}{0.21} \\
LI & \PDEMeanSD{0.17}{0.05} & \PDEMeanSD{0.04}{0.01} & \PDEMeanSD{0.04}{0.01} & \PDEMeanSD{0.48}{0.15} & \PDEMeanSD{0.74}{0.13} & \cellcolor{PDEFocus}\PDEMeanSD{0.03}{0.01} & \PDEMeanSD{0.77}{0.12} & \PDEMeanSD{0.14}{0.04} & \PDEMeanSD{0.17}{0.06} \\
\bottomrule
\end{tabular*}
\end{table}
\begin{table}[H]
\centering
\caption{Helmholtz / FNO. All 81 cells: mean $\pm$ SD on one line ($n=200$); blue marks matched views. All training views: 500/O500.}
\label{tab:matrix-helmholtz-fno}
\PDETableFont
\begin{tabular*}{\linewidth}{@{\extracolsep{\fill}}lrrrrrrrrr@{}}
\toprule
 & \multicolumn{9}{c}{Test observation pattern} \\
\cmidrule(l){2-10}
Train & R50 & H & V & CL & BL & LI & BD & R65 & R80 \\
\midrule
R50 & \cellcolor{PDEFocus}\PDEMeanSD{0.01}{0.00} & \PDEMeanSD{0.07}{0.04} & \PDEMeanSD{0.08}{0.05} & \PDEMeanSD{1.16}{0.41} & \PDEMeanSD{1.60}{0.65} & \PDEMeanSD{0.05}{0.04} & \PDEMeanSD{1.59}{0.65} & \PDEMeanSD{0.13}{0.03} & \PDEMeanSD{0.44}{0.10} \\
H & \PDEMeanSD{0.02}{0.00} & \cellcolor{PDEFocus}\PDEMeanSD{0.01}{0.00} & \PDEMeanSD{0.12}{0.05} & \PDEMeanSD{0.70}{0.29} & \PDEMeanSD{1.25}{0.40} & \PDEMeanSD{0.08}{0.03} & \PDEMeanSD{1.29}{0.39} & \PDEMeanSD{0.27}{0.09} & \PDEMeanSD{0.59}{0.20} \\
V & \PDEMeanSD{0.03}{0.01} & \PDEMeanSD{0.19}{0.08} & \cellcolor{PDEFocus}\PDEMeanSD{0.01}{0.00} & \PDEMeanSD{0.68}{0.22} & \PDEMeanSD{1.11}{0.29} & \PDEMeanSD{0.12}{0.05} & \PDEMeanSD{1.12}{0.28} & \PDEMeanSD{0.10}{0.02} & \PDEMeanSD{0.26}{0.07} \\
CL & \PDEMeanSD{0.04}{0.00} & \PDEMeanSD{0.04}{0.01} & \PDEMeanSD{0.04}{0.01} & \cellcolor{PDEFocus}\PDEMeanSD{0.04}{0.02} & \PDEMeanSD{0.33}{0.17} & \PDEMeanSD{0.04}{0.01} & \PDEMeanSD{0.42}{0.21} & \PDEMeanSD{0.07}{0.02} & \PDEMeanSD{0.13}{0.04} \\
BL & \PDEMeanSD{0.53}{0.14} & \PDEMeanSD{0.57}{0.20} & \PDEMeanSD{0.70}{0.32} & \PDEMeanSD{0.79}{0.28} & \cellcolor{PDEFocus}\PDEMeanSD{0.06}{0.03} & \PDEMeanSD{0.55}{0.18} & \PDEMeanSD{0.04}{0.02} & \PDEMeanSD{0.56}{0.15} & \PDEMeanSD{0.69}{0.19} \\
LI & \PDEMeanSD{0.02}{0.01} & \PDEMeanSD{0.02}{0.01} & \PDEMeanSD{0.02}{0.01} & \PDEMeanSD{0.53}{0.21} & \PDEMeanSD{0.81}{0.30} & \cellcolor{PDEFocus}\PDEMeanSD{0.02}{0.01} & \PDEMeanSD{0.82}{0.29} & \PDEMeanSD{0.12}{0.03} & \PDEMeanSD{0.36}{0.08} \\
\bottomrule
\end{tabular*}
\end{table}
\begin{table}[H]
\centering
\caption{Helmholtz / CNO. All 81 cells: mean $\pm$ SD on one line ($n=200$); blue marks matched views. All training views: 500/O500.}
\label{tab:matrix-helmholtz-cno}
\PDETableFont
\begin{tabular*}{\linewidth}{@{\extracolsep{\fill}}lrrrrrrrrr@{}}
\toprule
 & \multicolumn{9}{c}{Test observation pattern} \\
\cmidrule(l){2-10}
Train & R50 & H & V & CL & BL & LI & BD & R65 & R80 \\
\midrule
R50 & \cellcolor{PDEFocus}\PDEMeanSD{0.00}{0.00} & \PDEMeanSD{0.07}{0.05} & \PDEMeanSD{0.08}{0.05} & \PDEMeanSD{0.49}{0.15} & \PDEMeanSD{0.85}{0.09} & \PDEMeanSD{0.07}{0.03} & \PDEMeanSD{0.90}{0.06} & \PDEMeanSD{0.11}{0.01} & \PDEMeanSD{0.22}{0.04} \\
H & \PDEMeanSD{0.02}{0.01} & \cellcolor{PDEFocus}\PDEMeanSD{0.01}{0.00} & \PDEMeanSD{0.09}{0.03} & \PDEMeanSD{0.42}{0.21} & \PDEMeanSD{0.82}{0.23} & \PDEMeanSD{0.05}{0.01} & \PDEMeanSD{0.86}{0.22} & \PDEMeanSD{0.20}{0.03} & \PDEMeanSD{0.43}{0.07} \\
V & \PDEMeanSD{0.02}{0.00} & \PDEMeanSD{0.07}{0.02} & \cellcolor{PDEFocus}\PDEMeanSD{0.01}{0.00} & \PDEMeanSD{0.44}{0.20} & \PDEMeanSD{0.85}{0.24} & \PDEMeanSD{0.04}{0.01} & \PDEMeanSD{0.90}{0.23} & \PDEMeanSD{0.12}{0.02} & \PDEMeanSD{0.20}{0.04} \\
CL & \PDEMeanSD{0.15}{0.05} & \PDEMeanSD{0.15}{0.05} & \PDEMeanSD{0.14}{0.04} & \cellcolor{PDEFocus}\PDEMeanSD{0.09}{0.05} & \PDEMeanSD{0.57}{0.22} & \PDEMeanSD{0.14}{0.05} & \PDEMeanSD{0.63}{0.16} & \PDEMeanSD{0.17}{0.04} & \PDEMeanSD{0.22}{0.05} \\
BL & \PDEMeanSD{0.95}{0.34} & \PDEMeanSD{1.00}{0.34} & \PDEMeanSD{1.08}{0.40} & \PDEMeanSD{0.74}{0.30} & \cellcolor{PDEFocus}\PDEMeanSD{0.09}{0.05} & \PDEMeanSD{1.00}{0.34} & \PDEMeanSD{0.07}{0.03} & \PDEMeanSD{0.90}{0.32} & \PDEMeanSD{0.90}{0.32} \\
LI & \PDEMeanSD{0.01}{0.00} & \PDEMeanSD{0.01}{0.00} & \PDEMeanSD{0.01}{0.00} & \PDEMeanSD{0.37}{0.17} & \PDEMeanSD{0.78}{0.10} & \cellcolor{PDEFocus}\PDEMeanSD{0.01}{0.00} & \PDEMeanSD{0.83}{0.07} & \PDEMeanSD{0.13}{0.01} & \PDEMeanSD{0.25}{0.02} \\
\bottomrule
\end{tabular*}
\end{table}
\begin{table}[H]
\centering
\caption{Helmholtz / DeepONet. All 81 cells: mean $\pm$ SD on one line ($n=200$); blue marks matched views. All training views: 500/O500.}
\label{tab:matrix-helmholtz-deeponet}
\PDETableFont
\begin{tabular*}{\linewidth}{@{\extracolsep{\fill}}lrrrrrrrrr@{}}
\toprule
 & \multicolumn{9}{c}{Test observation pattern} \\
\cmidrule(l){2-10}
Train & R50 & H & V & CL & BL & LI & BD & R65 & R80 \\
\midrule
R50 & \cellcolor{PDEFocus}\PDEMeanSD{0.02}{0.01} & \PDEMeanSD{0.14}{0.04} & \PDEMeanSD{0.13}{0.05} & \PDEMeanSD{0.63}{0.13} & \PDEMeanSD{0.92}{0.10} & \PDEMeanSD{0.11}{0.03} & \PDEMeanSD{0.93}{0.09} & \PDEMeanSD{0.08}{0.03} & \PDEMeanSD{0.17}{0.06} \\
H & \PDEMeanSD{0.22}{0.03} & \cellcolor{PDEFocus}\PDEMeanSD{0.02}{0.01} & \PDEMeanSD{0.34}{0.08} & \PDEMeanSD{0.53}{0.16} & \PDEMeanSD{0.84}{0.14} & \PDEMeanSD{0.18}{0.08} & \PDEMeanSD{0.81}{0.12} & \PDEMeanSD{0.11}{0.02} & \PDEMeanSD{0.04}{0.01} \\
V & \PDEMeanSD{0.31}{0.09} & \PDEMeanSD{0.38}{0.14} & \cellcolor{PDEFocus}\PDEMeanSD{0.02}{0.01} & \PDEMeanSD{0.51}{0.16} & \PDEMeanSD{0.82}{0.22} & \PDEMeanSD{0.19}{0.09} & \PDEMeanSD{0.82}{0.23} & \PDEMeanSD{0.30}{0.09} & \PDEMeanSD{0.30}{0.09} \\
CL & \PDEMeanSD{0.28}{0.09} & \PDEMeanSD{3.01}{0.82} & \PDEMeanSD{3.12}{1.11} & \cellcolor{PDEFocus}\PDEMeanSD{0.16}{0.06} & \PDEMeanSD{0.78}{0.28} & \PDEMeanSD{1.43}{0.50} & \PDEMeanSD{0.89}{0.19} & \PDEMeanSD{0.23}{0.06} & \PDEMeanSD{0.23}{0.08} \\
BL & \PDEMeanSD{7.60}{3.38} & \PDEMeanSD{15.28}{7.37} & \PDEMeanSD{13.81}{6.76} & \PDEMeanSD{1.60}{0.80} & \cellcolor{PDEFocus}\PDEMeanSD{0.10}{0.07} & \PDEMeanSD{12.74}{5.56} & \PDEMeanSD{0.06}{0.02} & \PDEMeanSD{6.77}{3.09} & \PDEMeanSD{4.59}{2.21} \\
LI & \PDEMeanSD{0.37}{0.04} & \PDEMeanSD{0.13}{0.03} & \PDEMeanSD{0.14}{0.03} & \PDEMeanSD{0.48}{0.14} & \PDEMeanSD{0.77}{0.13} & \cellcolor{PDEFocus}\PDEMeanSD{0.03}{0.01} & \PDEMeanSD{0.78}{0.13} & \PDEMeanSD{0.25}{0.03} & \PDEMeanSD{0.11}{0.02} \\
\bottomrule
\end{tabular*}
\end{table}
\begin{table}[H]
\centering
\caption{Helmholtz / GNOT. All 81 cells: mean $\pm$ SD on one line ($n=200$); blue marks matched views. Epoch/cohort in row order: 500/O500, 500/O500, 500/O500, 500/O500, 200/M120, 45/P10, 500/O500, 200/P10, 500/O500.}
\label{tab:matrix-helmholtz-gnot}
\scriptsize
\renewcommand{\PDEMeanSDFont}{\scriptsize}
\setlength{\tabcolsep}{0.5pt}
\begin{tabular*}{\linewidth}{@{\extracolsep{\fill}}lrrrrrrrrr@{}}
\toprule
 & \multicolumn{9}{c}{Test observation pattern} \\
\cmidrule(l){2-10}
Train & R50 & H & V & CL & BL & LI & BD & R65 & R80 \\
\midrule
R50 & \cellcolor{PDEFocus}\PDEMeanSD{0.01}{0.00} & \PDEMeanSD{0.04}{0.12} & \PDEMeanSD{0.05}{0.33} & \PDEMeanSD{132.35}{117.02} & \PDEMeanSD{81.74}{74.92} & \PDEMeanSD{0.02}{0.01} & \PDEMeanSD{13.44}{11.45} & \PDEMeanSD{0.01}{0.00} & \PDEMeanSD{0.00}{0.00} \\
H & \PDEMeanSD{0.01}{0.00} & \cellcolor{PDEFocus}\PDEMeanSD{0.01}{0.00} & \PDEMeanSD{0.07}{0.14} & \PDEMeanSD{65.50}{98.78} & \PDEMeanSD{62.51}{54.10} & \PDEMeanSD{0.03}{0.01} & \PDEMeanSD{51.71}{16.29} & \PDEMeanSD{0.01}{0.00} & \PDEMeanSD{0.01}{0.00} \\
V & \PDEMeanSD{0.01}{0.00} & \PDEMeanSD{0.16}{0.44} & \cellcolor{PDEFocus}\PDEMeanSD{0.01}{0.00} & \PDEMeanSD{20.08}{37.63} & \PDEMeanSD{13.43}{8.04} & \PDEMeanSD{0.03}{0.01} & \PDEMeanSD{11.85}{4.28} & \PDEMeanSD{0.01}{0.00} & \PDEMeanSD{0.01}{0.00} \\
CL & \PDEMeanSD{0.02}{0.01} & \PDEMeanSD{0.02}{0.01} & \PDEMeanSD{0.02}{0.01} & \cellcolor{PDEFocus}\PDEMeanSD{0.03}{0.02} & \PDEMeanSD{0.49}{0.13} & \PDEMeanSD{0.02}{0.01} & \PDEMeanSD{0.49}{0.13} & \PDEMeanSD{0.02}{0.01} & \PDEMeanSD{0.02}{0.01} \\
BL & \PDEMeanSD{0.28}{0.11} & \PDEMeanSD{0.29}{0.11} & \PDEMeanSD{0.29}{0.11} & \PDEMeanSD{0.51}{0.39} & \cellcolor{PDEFocus}\PDEMeanSD{0.20}{0.15} & \PDEMeanSD{0.29}{0.11} & \PDEMeanSD{0.16}{0.12} & \PDEMeanSD{0.29}{0.11} & \PDEMeanSD{0.28}{0.11} \\
LI & \PDEMeanSD{0.14}{0.04} & \PDEMeanSD{0.15}{0.05} & \PDEMeanSD{0.14}{0.05} & \PDEMeanSD{0.37}{0.15} & \PDEMeanSD{1.40}{0.39} & \cellcolor{PDEFocus}\PDEMeanSD{0.14}{0.05} & \PDEMeanSD{1.36}{0.36} & \PDEMeanSD{0.14}{0.04} & \PDEMeanSD{0.14}{0.04} \\
\bottomrule
\end{tabular*}
\end{table}
\begin{table}[H]
\centering
\caption{Helmholtz / Transolver. All 81 cells: mean $\pm$ SD on one line ($n=200$); blue marks matched views. All training views: 500/O500.}
\label{tab:matrix-helmholtz-transolver}
\PDETableFont
\begin{tabular*}{\linewidth}{@{\extracolsep{\fill}}lrrrrrrrrr@{}}
\toprule
 & \multicolumn{9}{c}{Test observation pattern} \\
\cmidrule(l){2-10}
Train & R50 & H & V & CL & BL & LI & BD & R65 & R80 \\
\midrule
R50 & \cellcolor{PDEFocus}\PDEMeanSD{0.00}{0.00} & \PDEMeanSD{0.31}{0.10} & \PDEMeanSD{0.31}{0.13} & \PDEMeanSD{0.77}{0.21} & \PDEMeanSD{1.37}{0.39} & \PDEMeanSD{0.30}{0.09} & \PDEMeanSD{1.38}{0.37} & \PDEMeanSD{0.07}{0.01} & \PDEMeanSD{0.12}{0.03} \\
H & \PDEMeanSD{0.58}{0.05} & \cellcolor{PDEFocus}\PDEMeanSD{0.00}{0.00} & \PDEMeanSD{0.81}{0.22} & \PDEMeanSD{0.73}{0.33} & \PDEMeanSD{1.93}{0.89} & \PDEMeanSD{0.42}{0.05} & \PDEMeanSD{1.91}{0.77} & \PDEMeanSD{0.54}{0.11} & \PDEMeanSD{0.48}{0.25} \\
V & \PDEMeanSD{0.49}{0.07} & \PDEMeanSD{0.71}{0.08} & \cellcolor{PDEFocus}\PDEMeanSD{0.00}{0.00} & \PDEMeanSD{0.55}{0.16} & \PDEMeanSD{0.83}{0.21} & \PDEMeanSD{0.34}{0.06} & \PDEMeanSD{0.84}{0.19} & \PDEMeanSD{0.46}{0.08} & \PDEMeanSD{0.46}{0.10} \\
CL & \PDEMeanSD{0.34}{0.12} & \PDEMeanSD{0.56}{0.19} & \PDEMeanSD{0.46}{0.17} & \cellcolor{PDEFocus}\PDEMeanSD{0.05}{0.04} & \PDEMeanSD{0.60}{0.20} & \PDEMeanSD{0.55}{0.20} & \PDEMeanSD{0.60}{0.21} & \PDEMeanSD{0.30}{0.11} & \PDEMeanSD{0.37}{0.11} \\
BL & \PDEMeanSD{0.56}{0.15} & \PDEMeanSD{0.52}{0.15} & \PDEMeanSD{0.53}{0.16} & \PDEMeanSD{0.58}{0.20} & \cellcolor{PDEFocus}\PDEMeanSD{0.07}{0.05} & \PDEMeanSD{0.49}{0.14} & \PDEMeanSD{0.05}{0.03} & \PDEMeanSD{0.64}{0.17} & \PDEMeanSD{0.71}{0.20} \\
LI & \PDEMeanSD{0.29}{0.04} & \PDEMeanSD{0.02}{0.01} & \PDEMeanSD{0.07}{0.03} & \PDEMeanSD{0.73}{0.30} & \PDEMeanSD{1.43}{0.60} & \cellcolor{PDEFocus}\PDEMeanSD{0.00}{0.00} & \PDEMeanSD{1.57}{0.60} & \PDEMeanSD{0.27}{0.04} & \PDEMeanSD{0.26}{0.03} \\
\bottomrule
\end{tabular*}
\end{table}
\begin{table}[H]
\centering
\caption{Helmholtz / PINO. All 81 cells: mean $\pm$ SD on one line ($n=200$); blue marks matched views. All training views: 500/O500.}
\label{tab:matrix-helmholtz-pino}
\PDETableFont
\begin{tabular*}{\linewidth}{@{\extracolsep{\fill}}lrrrrrrrrr@{}}
\toprule
 & \multicolumn{9}{c}{Test observation pattern} \\
\cmidrule(l){2-10}
Train & R50 & H & V & CL & BL & LI & BD & R65 & R80 \\
\midrule
R50 & \cellcolor{PDEFocus}\PDEMeanSD{0.04}{0.01} & \PDEMeanSD{0.12}{0.04} & \PDEMeanSD{0.11}{0.05} & \PDEMeanSD{0.56}{0.18} & \PDEMeanSD{0.98}{0.17} & \PDEMeanSD{0.10}{0.04} & \PDEMeanSD{1.12}{0.15} & \PDEMeanSD{0.20}{0.03} & \PDEMeanSD{0.36}{0.05} \\
H & \PDEMeanSD{0.09}{0.04} & \cellcolor{PDEFocus}\PDEMeanSD{0.02}{0.01} & \PDEMeanSD{0.27}{0.13} & \PDEMeanSD{0.79}{0.15} & \PDEMeanSD{1.03}{0.20} & \PDEMeanSD{0.18}{0.08} & \PDEMeanSD{1.14}{0.23} & \PDEMeanSD{0.18}{0.03} & \PDEMeanSD{0.32}{0.04} \\
V & \PDEMeanSD{0.11}{0.04} & \PDEMeanSD{0.27}{0.11} & \cellcolor{PDEFocus}\PDEMeanSD{0.02}{0.01} & \PDEMeanSD{0.77}{0.16} & \PDEMeanSD{1.22}{0.30} & \PDEMeanSD{0.19}{0.08} & \PDEMeanSD{1.28}{0.32} & \PDEMeanSD{0.19}{0.03} & \PDEMeanSD{0.34}{0.04} \\
CL & \PDEMeanSD{0.28}{0.03} & \PDEMeanSD{0.28}{0.05} & \PDEMeanSD{0.27}{0.06} & \cellcolor{PDEFocus}\PDEMeanSD{0.11}{0.05} & \PDEMeanSD{0.44}{0.16} & \PDEMeanSD{0.27}{0.04} & \PDEMeanSD{0.44}{0.17} & \PDEMeanSD{0.18}{0.02} & \PDEMeanSD{0.11}{0.02} \\
BL & \PDEMeanSD{0.69}{0.11} & \PDEMeanSD{0.67}{0.12} & \PDEMeanSD{0.74}{0.18} & \PDEMeanSD{0.81}{0.15} & \cellcolor{PDEFocus}\PDEMeanSD{0.08}{0.06} & \PDEMeanSD{0.65}{0.16} & \PDEMeanSD{0.06}{0.03} & \PDEMeanSD{0.64}{0.12} & \PDEMeanSD{0.61}{0.13} \\
LI & \PDEMeanSD{0.05}{0.02} & \PDEMeanSD{0.04}{0.02} & \PDEMeanSD{0.04}{0.02} & \PDEMeanSD{0.52}{0.18} & \PDEMeanSD{0.85}{0.17} & \cellcolor{PDEFocus}\PDEMeanSD{0.04}{0.02} & \PDEMeanSD{0.96}{0.15} & \PDEMeanSD{0.15}{0.03} & \PDEMeanSD{0.30}{0.04} \\
\bottomrule
\end{tabular*}
\end{table}
\begin{table}[H]
\centering
\caption{Heat / U-FNO. All 81 cells: mean $\pm$ SD on one line ($n=200$); blue marks matched views. Epoch/cohort in row order: 500/R500, 500/R500, 500/R500, 500/R500, 500/O500, 500/R500, 500/R500, 500/R500, 500/R500.}
\label{tab:matrix-heat-ufno-2d}
\PDETableFont
\begin{tabular*}{\linewidth}{@{\extracolsep{\fill}}lrrrrrrrrr@{}}
\toprule
 & \multicolumn{9}{c}{Test observation pattern} \\
\cmidrule(l){2-10}
Train & R50 & H & V & CL & BL & LI & BD & R65 & R80 \\
\midrule
R50 & \cellcolor{PDEFocus}\PDEMeanSD{0.36}{0.07} & \PDEMeanSD{0.31}{0.06} & \PDEMeanSD{0.33}{0.06} & \PDEMeanSD{0.55}{0.10} & \PDEMeanSD{0.61}{0.10} & \PDEMeanSD{0.32}{0.06} & \PDEMeanSD{0.60}{0.11} & \PDEMeanSD{0.19}{0.05} & \PDEMeanSD{0.10}{0.02} \\
H & \PDEMeanSD{0.33}{0.02} & \cellcolor{PDEFocus}\PDEMeanSD{0.35}{0.03} & \PDEMeanSD{0.42}{0.03} & \PDEMeanSD{0.61}{0.09} & \PDEMeanSD{0.67}{0.10} & \PDEMeanSD{0.37}{0.03} & \PDEMeanSD{0.65}{0.10} & \PDEMeanSD{0.15}{0.02} & \PDEMeanSD{0.06}{0.04} \\
V & \PDEMeanSD{0.15}{0.04} & \PDEMeanSD{0.26}{0.05} & \cellcolor{PDEFocus}\PDEMeanSD{0.07}{0.03} & \PDEMeanSD{0.89}{0.25} & \PDEMeanSD{0.93}{0.19} & \PDEMeanSD{0.16}{0.04} & \PDEMeanSD{0.92}{0.19} & \PDEMeanSD{0.14}{0.05} & \PDEMeanSD{0.10}{0.06} \\
CL & \PDEMeanSD{0.13}{0.07} & \PDEMeanSD{0.13}{0.06} & \PDEMeanSD{0.14}{0.07} & \cellcolor{PDEFocus}\PDEMeanSD{0.15}{0.06} & \PDEMeanSD{0.14}{0.05} & \PDEMeanSD{0.13}{0.06} & \PDEMeanSD{0.14}{0.05} & \PDEMeanSD{0.08}{0.06} & \PDEMeanSD{0.09}{0.04} \\
BL & \PDEMeanSD{0.31}{0.04} & \PDEMeanSD{0.27}{0.05} & \PDEMeanSD{0.28}{0.05} & \PDEMeanSD{0.24}{0.07} & \cellcolor{PDEFocus}\PDEMeanSD{0.11}{0.03} & \PDEMeanSD{0.27}{0.05} & \PDEMeanSD{0.11}{0.03} & \PDEMeanSD{0.17}{0.05} & \PDEMeanSD{0.11}{0.05} \\
LI & \PDEMeanSD{0.20}{0.06} & \PDEMeanSD{0.24}{0.06} & \PDEMeanSD{0.22}{0.06} & \PDEMeanSD{0.37}{0.09} & \PDEMeanSD{0.36}{0.07} & \cellcolor{PDEFocus}\PDEMeanSD{0.22}{0.06} & \PDEMeanSD{0.35}{0.06} & \PDEMeanSD{0.09}{0.06} & \PDEMeanSD{0.06}{0.05} \\
\bottomrule
\end{tabular*}
\end{table}
\begin{table}[H]
\centering
\caption{Heat / FNO. All 81 cells: mean $\pm$ SD on one line ($n=200$); blue marks matched views. Epoch/cohort in row order: 500/O500, 500/O500, 500/O500, 500/O500, 500/O500, 500/INT, 500/O500, 500/O500, 500/O500.}
\label{tab:matrix-heat-fno}
\PDETableFont
\begin{tabular*}{\linewidth}{@{\extracolsep{\fill}}lrrrrrrrrr@{}}
\toprule
 & \multicolumn{9}{c}{Test observation pattern} \\
\cmidrule(l){2-10}
Train & R50 & H & V & CL & BL & LI & BD & R65 & R80 \\
\midrule
R50 & \cellcolor{PDEFocus}\PDEMeanSD{0.05}{0.05} & \PDEMeanSD{0.06}{0.04} & \PDEMeanSD{0.06}{0.04} & \PDEMeanSD{0.47}{0.17} & \PDEMeanSD{0.56}{0.16} & \PDEMeanSD{0.05}{0.05} & \PDEMeanSD{0.56}{0.17} & \PDEMeanSD{0.11}{0.04} & \PDEMeanSD{0.08}{0.04} \\
H & \PDEMeanSD{0.05}{0.05} & \cellcolor{PDEFocus}\PDEMeanSD{0.05}{0.05} & \PDEMeanSD{0.07}{0.04} & \PDEMeanSD{0.42}{0.13} & \PDEMeanSD{0.49}{0.10} & \PDEMeanSD{0.06}{0.04} & \PDEMeanSD{0.49}{0.11} & \PDEMeanSD{0.05}{0.05} & \PDEMeanSD{0.06}{0.04} \\
V & \PDEMeanSD{0.05}{0.04} & \PDEMeanSD{0.07}{0.04} & \cellcolor{PDEFocus}\PDEMeanSD{0.05}{0.04} & \PDEMeanSD{0.38}{0.11} & \PDEMeanSD{0.46}{0.11} & \PDEMeanSD{0.06}{0.04} & \PDEMeanSD{0.45}{0.11} & \PDEMeanSD{0.05}{0.04} & \PDEMeanSD{0.06}{0.04} \\
CL & \PDEMeanSD{0.07}{0.06} & \PDEMeanSD{0.07}{0.05} & \PDEMeanSD{0.07}{0.04} & \cellcolor{PDEFocus}\PDEMeanSD{0.07}{0.04} & \PDEMeanSD{0.08}{0.03} & \PDEMeanSD{0.06}{0.05} & \PDEMeanSD{0.08}{0.03} & \PDEMeanSD{0.11}{0.05} & \PDEMeanSD{0.08}{0.06} \\
BL & \PDEMeanSD{0.08}{0.03} & \PDEMeanSD{0.11}{0.04} & \PDEMeanSD{0.11}{0.04} & \PDEMeanSD{0.17}{0.05} & \cellcolor{PDEFocus}\PDEMeanSD{0.06}{0.05} & \PDEMeanSD{0.10}{0.03} & \PDEMeanSD{0.06}{0.05} & \PDEMeanSD{0.07}{0.03} & \PDEMeanSD{0.06}{0.04} \\
LI & \PDEMeanSD{0.05}{0.05} & \PDEMeanSD{0.05}{0.05} & \PDEMeanSD{0.05}{0.05} & \PDEMeanSD{0.31}{0.10} & \PDEMeanSD{0.39}{0.10} & \cellcolor{PDEFocus}\PDEMeanSD{0.05}{0.05} & \PDEMeanSD{0.38}{0.10} & \PDEMeanSD{0.07}{0.05} & \PDEMeanSD{0.08}{0.05} \\
\bottomrule
\end{tabular*}
\end{table}
\begin{table}[H]
\centering
\caption{Heat / CNO. All 81 cells: mean $\pm$ SD on one line ($n=200$); blue marks matched views. All training views: 500/O500.}
\label{tab:matrix-heat-cno}
\PDETableFont
\begin{tabular*}{\linewidth}{@{\extracolsep{\fill}}lrrrrrrrrr@{}}
\toprule
 & \multicolumn{9}{c}{Test observation pattern} \\
\cmidrule(l){2-10}
Train & R50 & H & V & CL & BL & LI & BD & R65 & R80 \\
\midrule
R50 & \cellcolor{PDEFocus}\PDEMeanSD{0.06}{0.04} & \PDEMeanSD{0.18}{0.09} & \PDEMeanSD{0.21}{0.10} & \PDEMeanSD{0.62}{0.09} & \PDEMeanSD{0.66}{0.08} & \PDEMeanSD{0.19}{0.07} & \PDEMeanSD{0.65}{0.08} & \PDEMeanSD{0.06}{0.04} & \PDEMeanSD{0.06}{0.04} \\
H & \PDEMeanSD{0.05}{0.05} & \cellcolor{PDEFocus}\PDEMeanSD{0.05}{0.05} & \PDEMeanSD{0.14}{0.07} & \PDEMeanSD{0.46}{0.12} & \PDEMeanSD{0.59}{0.10} & \PDEMeanSD{0.08}{0.03} & \PDEMeanSD{0.59}{0.11} & \PDEMeanSD{0.05}{0.04} & \PDEMeanSD{0.05}{0.04} \\
V & \PDEMeanSD{0.06}{0.04} & \PDEMeanSD{0.18}{0.08} & \cellcolor{PDEFocus}\PDEMeanSD{0.06}{0.05} & \PDEMeanSD{0.48}{0.11} & \PDEMeanSD{0.55}{0.10} & \PDEMeanSD{0.13}{0.06} & \PDEMeanSD{0.54}{0.10} & \PDEMeanSD{0.06}{0.04} & \PDEMeanSD{0.06}{0.05} \\
CL & \PDEMeanSD{0.09}{0.04} & \PDEMeanSD{0.09}{0.05} & \PDEMeanSD{0.08}{0.04} & \cellcolor{PDEFocus}\PDEMeanSD{0.12}{0.05} & \PDEMeanSD{0.30}{0.12} & \PDEMeanSD{0.08}{0.04} & \PDEMeanSD{0.33}{0.13} & \PDEMeanSD{0.08}{0.04} & \PDEMeanSD{0.07}{0.05} \\
BL & \PDEMeanSD{0.10}{0.04} & \PDEMeanSD{0.14}{0.04} & \PDEMeanSD{0.18}{0.06} & \PDEMeanSD{0.54}{0.16} & \cellcolor{PDEFocus}\PDEMeanSD{0.10}{0.04} & \PDEMeanSD{0.15}{0.04} & \PDEMeanSD{0.09}{0.03} & \PDEMeanSD{0.08}{0.04} & \PDEMeanSD{0.07}{0.04} \\
LI & \PDEMeanSD{0.05}{0.04} & \PDEMeanSD{0.06}{0.04} & \PDEMeanSD{0.06}{0.04} & \PDEMeanSD{0.43}{0.11} & \PDEMeanSD{0.52}{0.09} & \cellcolor{PDEFocus}\PDEMeanSD{0.05}{0.04} & \PDEMeanSD{0.51}{0.09} & \PDEMeanSD{0.05}{0.04} & \PDEMeanSD{0.05}{0.04} \\
\bottomrule
\end{tabular*}
\end{table}
\begin{table}[H]
\centering
\caption{Heat / DeepONet. All 81 cells: mean $\pm$ SD on one line ($n=200$); blue marks matched views. Epoch/cohort in row order: 500/O500, 500/O500, 500/O500, 500/R500, 500/R500, 500/O500, 500/R500, 500/R500, 500/O500.}
\label{tab:matrix-heat-deeponet}
\PDETableFont
\begin{tabular*}{\linewidth}{@{\extracolsep{\fill}}lrrrrrrrrr@{}}
\toprule
 & \multicolumn{9}{c}{Test observation pattern} \\
\cmidrule(l){2-10}
Train & R50 & H & V & CL & BL & LI & BD & R65 & R80 \\
\midrule
R50 & \cellcolor{PDEFocus}\PDEMeanSD{0.08}{0.03} & \PDEMeanSD{0.52}{0.13} & \PDEMeanSD{0.52}{0.13} & \PDEMeanSD{1.04}{0.19} & \PDEMeanSD{1.12}{0.21} & \PDEMeanSD{0.46}{0.10} & \PDEMeanSD{1.02}{0.17} & \PDEMeanSD{0.21}{0.09} & \PDEMeanSD{0.12}{0.07} \\
H & \PDEMeanSD{0.26}{0.07} & \cellcolor{PDEFocus}\PDEMeanSD{0.09}{0.03} & \PDEMeanSD{0.85}{0.40} & \PDEMeanSD{0.63}{0.16} & \PDEMeanSD{0.71}{0.12} & \PDEMeanSD{0.50}{0.30} & \PDEMeanSD{0.64}{0.12} & \PDEMeanSD{0.29}{0.13} & \PDEMeanSD{0.28}{0.12} \\
V & \PDEMeanSD{0.49}{0.13} & \PDEMeanSD{1.06}{0.42} & \cellcolor{PDEFocus}\PDEMeanSD{0.09}{0.04} & \PDEMeanSD{0.65}{0.14} & \PDEMeanSD{0.65}{0.12} & \PDEMeanSD{0.61}{0.25} & \PDEMeanSD{0.61}{0.11} & \PDEMeanSD{0.67}{0.20} & \PDEMeanSD{0.53}{0.18} \\
CL & \PDEMeanSD{0.25}{0.11} & \PDEMeanSD{1.06}{1.38} & \PDEMeanSD{8.75}{10.37} & \cellcolor{PDEFocus}\PDEMeanSD{0.20}{0.07} & \PDEMeanSD{0.49}{0.15} & \PDEMeanSD{1.18}{1.76} & \PDEMeanSD{0.46}{0.14} & \PDEMeanSD{0.24}{0.06} & \PDEMeanSD{0.21}{0.06} \\
BL & \PDEMeanSD{6.02}{4.26} & \PDEMeanSD{17.45}{9.95} & \PDEMeanSD{17.62}{10.32} & \PDEMeanSD{0.70}{0.29} & \cellcolor{PDEFocus}\PDEMeanSD{0.11}{0.05} & \PDEMeanSD{16.67}{10.00} & \PDEMeanSD{0.10}{0.04} & \PDEMeanSD{4.37}{3.08} & \PDEMeanSD{1.84}{1.00} \\
LI & \PDEMeanSD{0.27}{0.06} & \PDEMeanSD{0.19}{0.09} & \PDEMeanSD{0.18}{0.09} & \PDEMeanSD{0.69}{0.16} & \PDEMeanSD{0.80}{0.20} & \cellcolor{PDEFocus}\PDEMeanSD{0.08}{0.03} & \PDEMeanSD{0.75}{0.15} & \PDEMeanSD{0.20}{0.06} & \PDEMeanSD{0.21}{0.06} \\
\bottomrule
\end{tabular*}
\end{table}
\begin{table}[H]
\centering
\caption{Heat / GNOT. All 81 cells: mean $\pm$ SD on one line ($n=200$); blue marks matched views. Epoch/cohort in row order: 200/P10, 200/F200, 200/F200, 200/F200, 200/F200, 200/F200, 200/F200, 200/P10, 200/F200.}
\label{tab:matrix-heat-gnot}
\PDETableFont
\begin{tabular*}{\linewidth}{@{\extracolsep{\fill}}lrrrrrrrrr@{}}
\toprule
 & \multicolumn{9}{c}{Test observation pattern} \\
\cmidrule(l){2-10}
Train & R50 & H & V & CL & BL & LI & BD & R65 & R80 \\
\midrule
R50 & \cellcolor{PDEFocus}\PDEMeanSD{0.07}{0.03} & \PDEMeanSD{0.10}{0.03} & \PDEMeanSD{0.10}{0.03} & \PDEMeanSD{0.51}{0.17} & \PDEMeanSD{0.48}{0.16} & \PDEMeanSD{0.10}{0.03} & \PDEMeanSD{0.50}{0.16} & \PDEMeanSD{0.07}{0.03} & \PDEMeanSD{0.07}{0.03} \\
H & \PDEMeanSD{0.09}{0.04} & \cellcolor{PDEFocus}\PDEMeanSD{0.10}{0.03} & \PDEMeanSD{0.15}{0.05} & \PDEMeanSD{3.38}{3.97} & \PDEMeanSD{4.59}{4.14} & \PDEMeanSD{0.12}{0.04} & \PDEMeanSD{3.56}{3.87} & \PDEMeanSD{0.09}{0.04} & \PDEMeanSD{0.09}{0.04} \\
V & \PDEMeanSD{0.09}{0.03} & \PDEMeanSD{0.12}{0.04} & \cellcolor{PDEFocus}\PDEMeanSD{0.09}{0.03} & \PDEMeanSD{0.74}{0.71} & \PDEMeanSD{0.63}{0.24} & \PDEMeanSD{0.10}{0.03} & \PDEMeanSD{0.63}{0.20} & \PDEMeanSD{0.09}{0.03} & \PDEMeanSD{0.09}{0.03} \\
CL & \PDEMeanSD{0.07}{0.05} & \PDEMeanSD{0.08}{0.05} & \PDEMeanSD{0.08}{0.05} & \cellcolor{PDEFocus}\PDEMeanSD{0.13}{0.05} & \PDEMeanSD{0.23}{0.08} & \PDEMeanSD{0.08}{0.04} & \PDEMeanSD{0.29}{0.08} & \PDEMeanSD{0.07}{0.05} & \PDEMeanSD{0.07}{0.05} \\
BL & \PDEMeanSD{0.08}{0.05} & \PDEMeanSD{0.10}{0.04} & \PDEMeanSD{0.10}{0.04} & \PDEMeanSD{0.34}{0.11} & \cellcolor{PDEFocus}\PDEMeanSD{0.10}{0.04} & \PDEMeanSD{0.10}{0.04} & \PDEMeanSD{0.10}{0.04} & \PDEMeanSD{0.08}{0.05} & \PDEMeanSD{0.08}{0.05} \\
LI & \PDEMeanSD{0.10}{0.03} & \PDEMeanSD{0.11}{0.04} & \PDEMeanSD{0.11}{0.04} & \PDEMeanSD{0.82}{1.02} & \PDEMeanSD{1.34}{1.35} & \cellcolor{PDEFocus}\PDEMeanSD{0.11}{0.03} & \PDEMeanSD{1.33}{1.39} & \PDEMeanSD{0.10}{0.03} & \PDEMeanSD{0.10}{0.03} \\
\bottomrule
\end{tabular*}
\end{table}
\begin{table}[H]
\centering
\caption{Heat / Transolver. All 81 cells: mean $\pm$ SD on one line ($n=200$); blue marks matched views. Epoch/cohort in row order: 257/INT, 500/O500, 124/INT, 130/M120, 200/M120, 500/O500, 500/O500, 157/M120, 133/M120.}
\label{tab:matrix-heat-transolver}
\PDETableFont
\begin{tabular*}{\linewidth}{@{\extracolsep{\fill}}lrrrrrrrrr@{}}
\toprule
 & \multicolumn{9}{c}{Test observation pattern} \\
\cmidrule(l){2-10}
Train & R50 & H & V & CL & BL & LI & BD & R65 & R80 \\
\midrule
R50 & \cellcolor{PDEFocus}\PDEMeanSD{0.05}{0.04} & \PDEMeanSD{0.19}{0.08} & \PDEMeanSD{0.22}{0.09} & \PDEMeanSD{0.74}{0.14} & \PDEMeanSD{0.82}{0.10} & \PDEMeanSD{0.17}{0.05} & \PDEMeanSD{0.82}{0.10} & \PDEMeanSD{0.06}{0.04} & \PDEMeanSD{0.07}{0.03} \\
H & \PDEMeanSD{0.26}{0.04} & \cellcolor{PDEFocus}\PDEMeanSD{0.07}{0.05} & \PDEMeanSD{0.48}{0.06} & \PDEMeanSD{0.50}{0.14} & \PDEMeanSD{0.53}{0.14} & \PDEMeanSD{0.16}{0.03} & \PDEMeanSD{0.52}{0.13} & \PDEMeanSD{0.19}{0.04} & \PDEMeanSD{0.14}{0.04} \\
V & \PDEMeanSD{0.22}{0.02} & \PDEMeanSD{0.43}{0.04} & \cellcolor{PDEFocus}\PDEMeanSD{0.07}{0.04} & \PDEMeanSD{0.74}{0.15} & \PDEMeanSD{0.82}{0.14} & \PDEMeanSD{0.21}{0.04} & \PDEMeanSD{0.82}{0.14} & \PDEMeanSD{0.15}{0.03} & \PDEMeanSD{0.10}{0.03} \\
CL & \PDEMeanSD{0.14}{0.02} & \PDEMeanSD{0.17}{0.03} & \PDEMeanSD{0.17}{0.03} & \cellcolor{PDEFocus}\PDEMeanSD{0.35}{0.11} & \PDEMeanSD{0.51}{0.10} & \PDEMeanSD{0.16}{0.02} & \PDEMeanSD{0.50}{0.11} & \PDEMeanSD{0.13}{0.02} & \PDEMeanSD{0.11}{0.02} \\
BL & \PDEMeanSD{0.49}{0.04} & \PDEMeanSD{0.35}{0.08} & \PDEMeanSD{0.34}{0.07} & \PDEMeanSD{0.45}{0.11} & \cellcolor{PDEFocus}\PDEMeanSD{0.08}{0.04} & \PDEMeanSD{0.35}{0.07} & \PDEMeanSD{0.08}{0.04} & \PDEMeanSD{0.49}{0.03} & \PDEMeanSD{0.40}{0.02} \\
LI & \PDEMeanSD{0.06}{0.04} & \PDEMeanSD{0.09}{0.04} & \PDEMeanSD{0.08}{0.04} & \PDEMeanSD{0.42}{0.11} & \PDEMeanSD{0.48}{0.15} & \cellcolor{PDEFocus}\PDEMeanSD{0.06}{0.05} & \PDEMeanSD{0.47}{0.16} & \PDEMeanSD{0.06}{0.05} & \PDEMeanSD{0.06}{0.05} \\
\bottomrule
\end{tabular*}
\end{table}
\begin{table}[H]
\centering
\caption{Heat / PINO. All 81 cells: mean $\pm$ SD on one line ($n=200$); blue marks matched views. Epoch/cohort in row order: 500/O500, 500/R500, 500/O500, 500/O500, 500/O500, 500/R500, 500/R500, 500/O500, 500/O500.}
\label{tab:matrix-heat-pino}
\PDETableFont
\begin{tabular*}{\linewidth}{@{\extracolsep{\fill}}lrrrrrrrrr@{}}
\toprule
 & \multicolumn{9}{c}{Test observation pattern} \\
\cmidrule(l){2-10}
Train & R50 & H & V & CL & BL & LI & BD & R65 & R80 \\
\midrule
R50 & \cellcolor{PDEFocus}\PDEMeanSD{0.06}{0.05} & \PDEMeanSD{0.07}{0.06} & \PDEMeanSD{0.07}{0.06} & \PDEMeanSD{0.55}{0.11} & \PDEMeanSD{0.66}{0.10} & \PDEMeanSD{0.07}{0.04} & \PDEMeanSD{0.66}{0.10} & \PDEMeanSD{0.06}{0.04} & \PDEMeanSD{0.10}{0.03} \\
H & \PDEMeanSD{0.06}{0.03} & \cellcolor{PDEFocus}\PDEMeanSD{0.05}{0.04} & \PDEMeanSD{0.12}{0.03} & \PDEMeanSD{0.44}{0.11} & \PDEMeanSD{0.55}{0.09} & \PDEMeanSD{0.09}{0.02} & \PDEMeanSD{0.55}{0.09} & \PDEMeanSD{0.06}{0.03} & \PDEMeanSD{0.07}{0.03} \\
V & \PDEMeanSD{0.07}{0.04} & \PDEMeanSD{0.15}{0.04} & \cellcolor{PDEFocus}\PDEMeanSD{0.06}{0.04} & \PDEMeanSD{0.54}{0.11} & \PDEMeanSD{0.63}{0.08} & \PDEMeanSD{0.11}{0.03} & \PDEMeanSD{0.63}{0.08} & \PDEMeanSD{0.06}{0.05} & \PDEMeanSD{0.07}{0.05} \\
CL & \PDEMeanSD{0.08}{0.03} & \PDEMeanSD{0.09}{0.02} & \PDEMeanSD{0.17}{0.07} & \cellcolor{PDEFocus}\PDEMeanSD{0.08}{0.03} & \PDEMeanSD{0.09}{0.03} & \PDEMeanSD{0.12}{0.05} & \PDEMeanSD{0.09}{0.03} & \PDEMeanSD{0.08}{0.03} & \PDEMeanSD{0.09}{0.04} \\
BL & \PDEMeanSD{0.08}{0.04} & \PDEMeanSD{0.18}{0.05} & \PDEMeanSD{0.16}{0.04} & \PDEMeanSD{0.29}{0.09} & \cellcolor{PDEFocus}\PDEMeanSD{0.07}{0.04} & \PDEMeanSD{0.12}{0.04} & \PDEMeanSD{0.07}{0.05} & \PDEMeanSD{0.09}{0.05} & \PDEMeanSD{0.10}{0.05} \\
LI & \PDEMeanSD{0.06}{0.05} & \PDEMeanSD{0.07}{0.04} & \PDEMeanSD{0.07}{0.04} & \PDEMeanSD{0.34}{0.11} & \PDEMeanSD{0.42}{0.10} & \cellcolor{PDEFocus}\PDEMeanSD{0.06}{0.04} & \PDEMeanSD{0.42}{0.10} & \PDEMeanSD{0.06}{0.04} & \PDEMeanSD{0.06}{0.05} \\
\bottomrule
\end{tabular*}
\end{table}
\begin{table}[H]
\centering
\caption{Reaction--diffusion / U-FNO. All 81 cells: mean $\pm$ SD on one line ($n=200$); blue marks matched views. Epoch/cohort in row order: 500/O500, 500/R500, 500/R500, 500/R500, 500/R500, 500/R500, 500/R500, 500/O500, 500/O500.}
\label{tab:matrix-reaction_diffusion-ufno-2d}
\PDETableFont
%
\end{table}
\begin{table}[H]
\centering
\caption{Reaction--diffusion / FNO. All 81 cells: mean $\pm$ SD on one line ($n=200$); blue marks matched views. Epoch/cohort in row order: 500/O500, 500/O500, 500/O500, 469/INT, 500/O500, 500/O500, 500/O500, 500/O500, 500/O500.}
\label{tab:matrix-reaction_diffusion-fno}
\PDETableFont
%
\end{table}
\begin{table}[H]
\centering
\caption{Reaction--diffusion / CNO. All 81 cells: mean $\pm$ SD on one line ($n=200$); blue marks matched views. All training views: 500/O500.}
\label{tab:matrix-reaction_diffusion-cno}
\PDETableFont
%
\end{table}
\begin{table}[H]
\centering
\caption{Reaction--diffusion / DeepONet. All 81 cells: mean $\pm$ SD on one line ($n=200$); blue marks matched views. Epoch/cohort in row order: 500/O500, 500/O500, 500/O500, 500/O500, 500/O500, 500/O500, 500/O500, 251/INT, 500/O500.}
\label{tab:matrix-reaction_diffusion-deeponet}
\PDETableFont
%
\end{table}
\begin{table}[H]
\centering
\caption{Reaction--diffusion / GNOT. All 81 cells: mean $\pm$ SD on one line ($n=200$); blue marks matched views. Epoch/cohort in row order: 200/P10, 200/F200, 200/F200, 200/F200, 200/F200, 200/F200, 200/F200, 200/F200, 200/F200.}
\label{tab:matrix-reaction_diffusion-gnot}
\PDETableFont
%
\end{table}
\begin{table}[H]
\centering
\caption{Reaction--diffusion / Transolver. All 81 cells: mean $\pm$ SD on one line ($n=200$); blue marks matched views. Epoch/cohort in row order: 200/P10, 200/M120, 200/F200, 200/M120, 35/INT, 47/INT, 200/F200, 200/F200, 200/F200.}
\label{tab:matrix-reaction_diffusion-transolver}
\PDETableFont
%
\end{table}
\begin{table}[H]
\centering
\caption{Reaction--diffusion / PINO. All 81 cells: mean $\pm$ SD on one line ($n=200$); blue marks matched views. Epoch/cohort in row order: 500/O500, 500/O500, 500/R500, 500/R500, 500/R500, 500/R500, 500/O500, 392/INT, 391/INT.}
\label{tab:matrix-reaction_diffusion-pino}
\PDETableFont
%
\end{table}
\begin{table}[H]
\centering
\caption{Burgers / U-FNO. All 81 cells: mean $\pm$ SD on one line ($n=200$); blue marks matched views. All training views: 500/R500.}
\label{tab:matrix-burgers-ufno-2d}
\PDETableFont
%
\end{table}
\begin{table}[H]
\centering
\caption{Burgers / FNO. All 81 cells: mean $\pm$ SD on one line ($n=200$); blue marks matched views. Epoch/cohort in row order: 500/R500, 500/O500, 500/R500, 500/R500, 500/R500, 500/R500, 500/R500, 500/R500, 500/R500.}
\label{tab:matrix-burgers-fno}
\PDETableFont
%
\end{table}
\begin{table}[H]
\centering
\caption{Burgers / CNO. All 81 cells: mean $\pm$ SD on one line ($n=200$); blue marks matched views. All training views: 500/O500.}
\label{tab:matrix-burgers-cno}
\PDETableFont
%
\end{table}
\begin{table}[H]
\centering
\caption{Burgers / DeepONet. All 81 cells: mean $\pm$ SD on one line ($n=200$); blue marks matched views. Epoch/cohort in row order: 500/O500, 500/R500, 339/INT, 500/R500, 500/O500, 443/INT, 500/O500, 491/INT, 500/O500.}
\label{tab:matrix-burgers-deeponet}
\PDETableFont
%
\end{table}
\begin{table}[H]
\centering
\caption{Burgers / GNOT. All 81 cells: mean $\pm$ SD on one line ($n=200$); blue marks matched views. Epoch/cohort in row order: 200/F200, 500/O500, 200/F200, 200/F200, 500/O500, 200/P10, 200/F200, 200/F200, 200/F200.}
\label{tab:matrix-burgers-gnot}
\PDETableFont
%
\end{table}
\begin{table}[H]
\centering
\caption{Burgers / Transolver. All 81 cells: mean $\pm$ SD on one line ($n=200$); blue marks matched views. Epoch/cohort in row order: 200/F200, 200/M120, 200/M120, 200/M120, 200/M120, 200/M120, 73/P10, 200/F200, 200/F200.}
\label{tab:matrix-burgers-transolver}
\PDETableFont
%
\end{table}
\begin{table}[H]
\centering
\caption{Burgers / PINO. All 81 cells: mean $\pm$ SD on one line ($n=200$); blue marks matched views. Epoch/cohort in row order: 500/R500, 500/O500, 500/R500, 500/R500, 500/O500, 500/R500, 500/R500, 500/O500, 500/O500.}
\label{tab:matrix-burgers-pino}
\PDETableFont
%
\end{table}
\begin{table}[H]
\centering
\caption{Navier--Stokes / U-FNO. All 81 cells: mean $\pm$ SD on one line ($n=200$); blue marks matched views. Epoch/cohort in row order: 500/R500, 500/R500, 500/R500, 500/R500, 500/R500, 500/R500, 500/O500, 500/R500, 500/R500.}
\label{tab:matrix-navier_stokes-ufno-2d}
\PDETableFont
%
\end{table}
\begin{table}[H]
\centering
\caption{Navier--Stokes / FNO. All 81 cells: mean $\pm$ SD on one line ($n=200$); blue marks matched views. Epoch/cohort in row order: 500/O500, 500/O500, 36/INT, 500/O500, 500/O500, 500/O500, 500/O500, 500/O500, 500/O500.}
\label{tab:matrix-navier_stokes-fno}
\PDETableFont
%
\end{table}
\begin{table}[H]
\centering
\caption{Navier--Stokes / CNO. All 81 cells: mean $\pm$ SD on one line ($n=200$); blue marks matched views. All training views: 500/O500.}
\label{tab:matrix-navier_stokes-cno}
\PDETableFont
%
\end{table}
\begin{table}[H]
\centering
\caption{Navier--Stokes / DeepONet. All 81 cells: mean $\pm$ SD on one line ($n=200$); blue marks matched views. Epoch/cohort in row order: 500/O500, 500/O500, 500/O500, 500/O500, 500/O500, 128/INT, 500/O500, 500/O500, 500/O500.}
\label{tab:matrix-navier_stokes-deeponet}
\PDETableFont
%
\end{table}
\begin{table}[H]
\centering
\caption{Navier--Stokes / GNOT. All 81 cells: mean $\pm$ SD on one line ($n=200$); blue marks matched views. Epoch/cohort in row order: 200/F200, 200/F200, 200/F200, 200/F200, 200/F200, 200/F200, 200/F200, 500/O500, 200/F200.}
\label{tab:matrix-navier_stokes-gnot}
\PDETableFont
%
\end{table}
\begin{table}[H]
\centering
\caption{Navier--Stokes / Transolver. All 81 cells: mean $\pm$ SD on one line ($n=200$); blue marks matched views. Epoch/cohort in row order: 200/M120, 31/P10, 200/M120, 200/M120, 200/M120, 200/M120, 200/M120, 200/M120, 200/M120.}
\label{tab:matrix-navier_stokes-transolver}
\PDETableFont
%
\end{table}
\begin{table}[H]
\centering
\caption{Navier--Stokes / PINO. All 81 cells: mean $\pm$ SD on one line ($n=200$); blue marks matched views. All training views: 500/R500.}
\label{tab:matrix-navier_stokes-pino}
\PDETableFont
%
\end{table}
\endgroup

\clearpage
\section{\revnew{Mixed-Pattern Training}}
\label{app:mixed-control}
\begin{newrevision}
This appendix records five completed mixed-pattern models reported in Figure~\ref{fig:mixed-control}: Helmholtz/FNO, Helmholtz/DeepONet, Heat/DeepONet, Navier--Stokes/U-FNO, and Poisson/U-FNO. One additional model per pair is trained on a deterministic mixture of the nine frozen observation views and scored on those views with the same 200 held-out identities as the single-pattern models. These study rows supplement the frozen 441-model grid; no grid value is replaced.

\paragraph{Training mixture.}
The training mask is requested as one specification, the nine frozen views joined by \code{+} with their frozen arguments:
\par\begingroup\raggedright
\code{random\_3pct(ratio=0.5) + random\_3pct(ratio=0.65) + random\_3pct(ratio=0.8) + block\_missing(missing\_fraction=0.49) + line\_sensors(num\_lines=76, orientation=both) + line\_sensors(num\_lines=64, orientation=horizontal) + line\_sensors(num\_lines=64, orientation=vertical) + boundary\_sensors(width=19) + clustered\_sensors(ratio=0.5)}.
\par\endgroup
Each physical identity is assigned exactly one component: the nine-view cycle is rotated by a seed derived from the campaign mask seed, the canonical specification and the identity's regime, and indexed by its generation-time \code{regime\_sample\_index}. Selecting the training identities after this assignment yields 190--208 records per view among the 1,800 training records. The training mixture is therefore approximately balanced, not rebalanced within each training regime. Assignments remain fixed across epochs and methods. Each component derives the per-identity mask seed from its resolved protocol name in the same way as a frozen view, so a given record exposes the same sensors to the mixed model and the corresponding single-pattern model. Components were checked against the frozen views bit for bit before training.

\paragraph{Recorded training budgets and comparison scopes.}
The MIX models retain the corresponding recorded batch, optimizer, learning-rate schedule, loss and stopping settings, changing the observation assignment. Actual completion receipts record 500 epochs and 112,500 updates for Helmholtz/FNO; 500 epochs and 225,000 updates each for Helmholtz/DeepONet, Heat/DeepONet and Navier--Stokes/U-FNO; and 200 epochs and 90,000 updates for Poisson/U-FNO. Equal stopping rules alone would not establish equal completed budgets. The four 500-epoch pairs have 500-epoch single-pattern references. Poisson uses the settings-v3 one-cycle recipe (learning rate $5\times10^{-4}$, maximum 200 epochs, train-loss patience after epoch 120); only its BL, BD and CL references use that recipe and complete 200 epochs with 90,000 updates. Its other six references use a step schedule and stop at 130--163 epochs. All-nine Poisson comparisons are consequently descriptive: schedule, completed budget and exposure differ, and these results do not identify which factor dominates.

\paragraph{Scoring.}
Every MIX result uses strict float64 rescoring of retained prediction arrays with the main-grid scorer, not a legacy scalar. The 45 reference-view and 15 extra-view blocks each contain 200 held-out identities; contracts bind identities, masks, target frames, checkpoints and configurations. All destination-trained denominators and transferred-model comparisons use the main grid's \code{20260925-v1} prediction scores. The three extra protocols are uniformly random points at 20\,\% and 35\,\% density, and a deterministic mixture assigning either line sensors at 30\,\% or random points at 20\,\% to a record, not their union. No specialist was trained on these extra protocols, so no specialist ratio is reported for them. Full-precision comparisons and score bindings are recorded in \nolinkurl{results/revision_v2/mixed_comparisons.csv}, \nolinkurl{mixed_extra_views.csv} and \nolinkurl{mixed_summary.json}.

\begin{table}[htbp]
\centering
\caption{\revnew{\textbf{Mixed-pattern comparisons with explicit populations.} All columns in a row use the same destination set $W$. MIX and specialist errors are mean $\pm$ sample SD across the $|W|$ cell means, each based on 200 records, multiplied by 100 (error \%, not observation density). The geometric mean of $R_{\mathrm{MIX}\to w}$ is dimensionless. Poisson's all-nine and recipe-matched populations are separate; the full-grid $C$ remains in \appref{app:distributions} and is not reused for the restricted set.}}
\label{tab:mixed-control}
\small
\setlength{\tabcolsep}{4pt}
\begin{tabular}{@{}llrrr@{}}
\toprule
& & \multicolumn{2}{c}{Error (\%): mean $\pm$ view SD} & \\
\cmidrule(lr){3-4}
Pair & Destinations $W$ & MIX & Specialist & geo.\ $R$ \\
\midrule
Helmholtz / FNO & All nine & $5.31\pm5.81$ & $2.12\pm1.74$ & 2.07 \\
Helmholtz / DeepONet & All nine & $12.14\pm10.68$ & $5.01\pm4.82$ & 2.56 \\
Heat / DeepONet & All nine & $13.29\pm6.28$ & $9.83\pm3.78$ & 1.32 \\
Navier--Stokes / U-FNO & All nine & $3.49\pm3.98$ & $2.55\pm2.87$ & 1.19 \\
\midrule
Poisson / U-FNO & All nine (descriptive) & $7.12\pm8.01$ & $11.65\pm9.77$ & 0.46 \\
Poisson / U-FNO & BL, BD, CL (matched) & $17.10\pm5.69$ & $5.50\pm1.19$ & 3.05 \\
\bottomrule
\end{tabular}
\end{table}

\begin{table}[htbp]
\centering
\caption{\revnew{\textbf{MIX on protocols absent from training.} Error (\%) is $100\times$ per-record relative-$L^2$, shown as mean $\pm$ sample SD over 200 records ($\mathrm{ddof}=1$). Column percentages instead specify observation density. ``Line 30 / random 20'' assigns one of the two patterns per record. There is no matched specialist for these protocols.}}
\label{tab:mixed-extra}
\small
\begin{tabular}{@{}lrrr@{}}
\toprule
& \multicolumn{3}{c}{Test observation protocol} \\
\cmidrule(lr){2-4}
Pair & Random 20\% & Random 35\% & Line 30 / random 20 \\
\midrule
Helmholtz / FNO & $14.22\pm3.98$ & $4.34\pm1.44$ & $11.50\pm4.56$ \\
Helmholtz / DeepONet & $18.31\pm4.18$ & $6.73\pm1.97$ & $15.03\pm5.82$ \\
Heat / DeepONet & $23.21\pm9.93$ & $13.22\pm5.92$ & $18.92\pm8.73$ \\
Navier--Stokes / U-FNO & $8.39\pm2.23$ & $1.74\pm0.56$ & $6.27\pm2.80$ \\
Poisson / U-FNO & $15.30\pm1.80$ & $4.47\pm0.94$ & $12.49\pm7.14$ \\
\bottomrule
\end{tabular}
\end{table}

\paragraph{Reading.}
MIX is below the transferred-model median in 45/45 comparisons and below the best transfer in 34/45; transfer ratios reach 1,040.09. The four 500-epoch pairs give specialist-relative ratios of 0.63--4.07, including Navier--Stokes gains on R50, LI and V (0.63, 0.96 and 0.98). Poisson's recipe-matched range is 2.51--3.49. Extra densities (Figure~\ref{fig:mixed-density}, Table~\ref{tab:mixed-extra}) are separately sampled discrete tests, not a continuous or nested-mask density law. These completed reconstruction and forecasting cases are not an architecture ranking or a complete attempt inventory: an additional attempt collapsed at epoch 90 and was not scored. Longer-training and capacity controls remain untested.

\begin{figure}[h]
\centering
\includegraphics[width=0.6\linewidth]{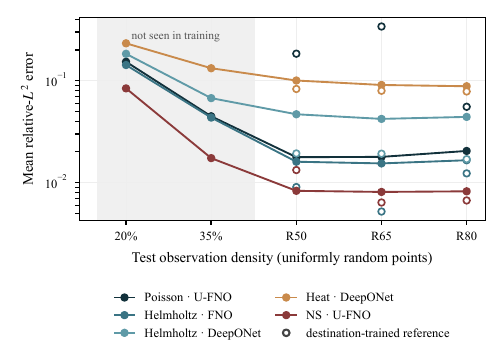}
\caption{\revnew{\textbf{Density response of five completed MIX models.} Mean relative-$L^2$ error at random observation densities of 20\%, 35\%, 50\% (R50), 65\% (R65) and 80\% (R80); shaded densities were absent from training. Open markers show destination-trained references where available. Masks are separately sampled, not nested; connecting lines do not establish a continuous density law.}}
\label{fig:mixed-density}
\end{figure}
\end{newrevision}

\stopcontents[sections]

\end{document}